\documentclass{article}

\PassOptionsToPackage{numbers}{natbib}

\usepackage[main, final]{neurips_2026}

\usepackage[utf8]{inputenc} 
\usepackage[T1]{fontenc}    
\usepackage{hyperref}       
\usepackage{url}            
\usepackage{booktabs}       
\usepackage{amsfonts}       
\usepackage{nicefrac}       
\usepackage{microtype}      
\usepackage{xcolor}         

\usepackage{graphicx}
\usepackage{subcaption}

\usepackage{amsmath}
\usepackage{amssymb}
\usepackage{mathtools}
\usepackage{amsthm}
\usepackage{tcolorbox} 
\usepackage{bm}
\usepackage{arydshln}
\usepackage{makecell}
\usepackage{adjustbox}
\usepackage{wrapfig}
\usepackage{multirow}
\usepackage{algorithm}
\usepackage{algorithmic}

\usepackage[table,dvipsnames]{xcolor}

\usepackage{hyperref}

\makeatletter
\newcommand{\REPEATN}[1]{\ALC@it\algorithmicrepeat%
\ #1 \textbf{times}\begin{ALC@rpt}}%
\newcommand{\ENDREPEAT}{\end{ALC@rpt}\ALC@it\algorithmicend}%
\makeatother

\newcommand{\highlightcyan}[1]{%
 \par\noindent
 \colorbox{cyan!7}{%
 \parbox{\dimexpr\linewidth-2\fboxsep\relax}{%
 #1
 }%
 }}
\newcommand{\highlightgray}[1]{%
\par\noindent
\colorbox{gray!7}{%
\parbox{\dimexpr\linewidth-2\fboxsep\relax}{%
#1
}%
}}

\newcolumntype{C}[1]{>{\centering\arraybackslash}p{#1}}
\newcommand{\smallsec}[1]{\noindent {\bf #1.}}

\title{D2D: Detector-to-Differentiable Critic for Improved Numeracy in Text-to-Image Generation}

\author{%
  Nobline Yoo\thanks{Work mainly completed while at Princeton University and the summer prior to doctoral program. Correspondence to \texttt{nobline@engineering.upenn.edu}.} \\
  University of Pennsylvania \\
  \And
  Olga Russakovsky \\
  Princeton University \\
  \And
  Ye Zhu \\
  LIX, CNRS, École Polytechnique, \\
  IP Paris \\
}

\begin{document}

\maketitle

\begin{abstract}
  Text-to-image (T2I) diffusion models have achieved strong performance in semantic alignment, yet they still struggle with generating the correct number of objects specified in prompts. Existing approaches typically incorporate auxiliary counting networks as external critics to enhance numeracy. However, since these critics must provide gradient guidance during generation, they are restricted to regression-based models that are inherently \emph{differentiable}, thus excluding detector-based models with superior counting ability, whose count-via-enumeration nature is \emph{non-differentiable}. To overcome this, we propose \textbf{Detector-to-Differentiable} (\emph{D2D}), a novel framework that transforms non-differentiable detection models into differentiable critics, leveraging their superior counting ability to guide numeracy generation. Specifically, we design custom activation functions to convert detector logits into soft binary indicators, which are then used to optimize the noise prior at inference time with pre-trained T2I models.
  Our extensive experiments on SDXL-Turbo, SD-Turbo, and Pixart-DMD across four benchmarks of varying complexity (low-density, high-density, and multi-object scenarios) demonstrate consistent and substantial improvements in object counting accuracy (e.g., boosting up to 13.7\% on D2D-Small, a 400-prompt, low-density benchmark), with minimal degradation in overall image quality and computational overhead. 
  Project page: \textcolor{magenta}{\texttt{https://nobliney.github.io/d2d/}}.
\end{abstract}

\section{Introduction}
\begin{figure}[h]
    \centering
    \includegraphics[width=\textwidth]{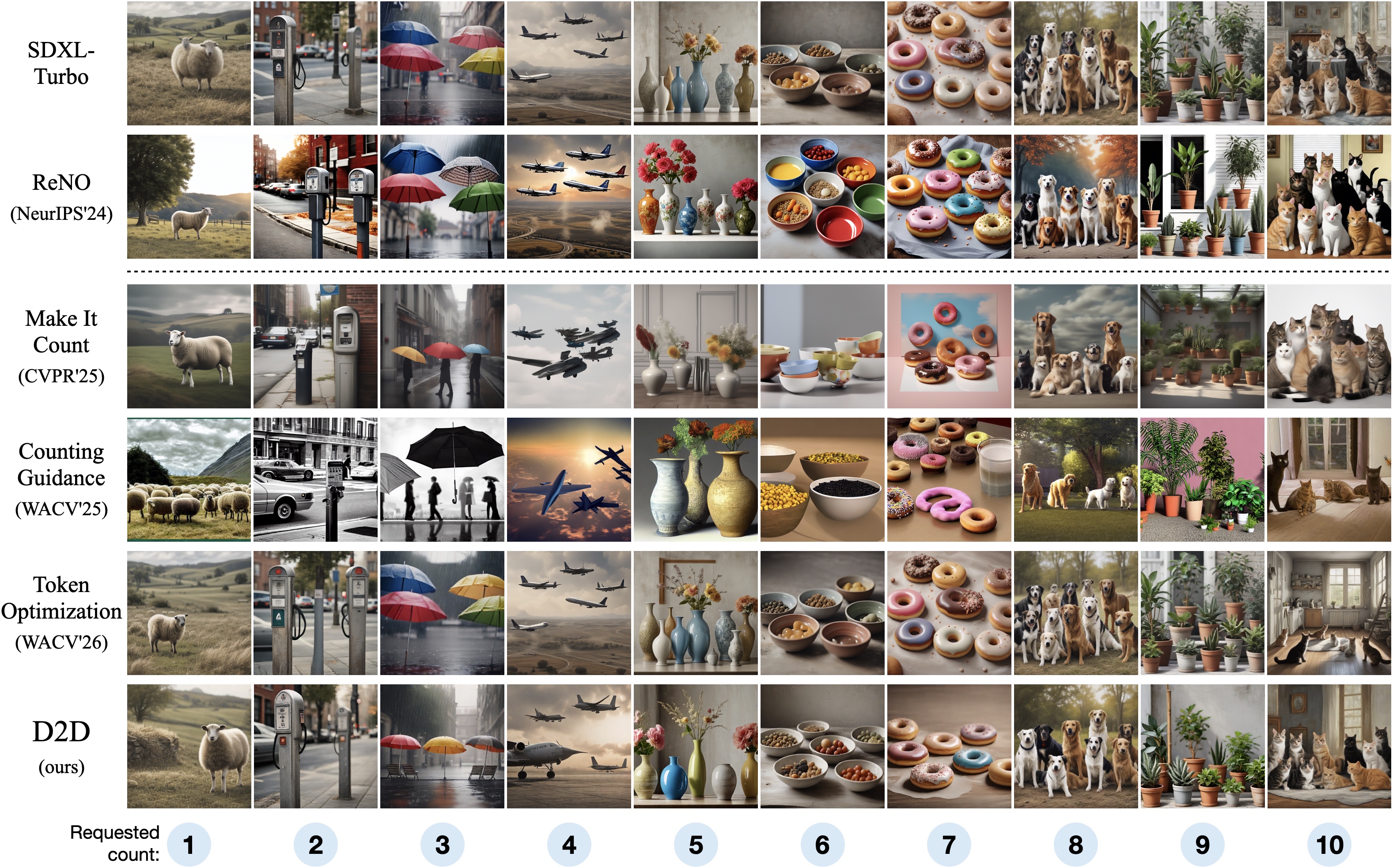}
    \caption{\textbf{Qualitative examples illustrating the count-correction ability of our detector-based critic on a variety of objects, counts 1-10.} SDXL-Turbo~\cite{sauer2025turbo} is a base model with no post-enhancement. ReNO~\cite{eyring2024reno} is a generic semantic alignment method that exhibits limited performance in this setting. More recent methods (e.g., Make It Count~\cite{binyamin2025mic}, Counting Guidance~\cite{kang2025cg}, TokenOpt~\cite{zafar2026tokenopt}), explicitly address count-correction. Our method proposes a new, effective way to leverage detectors for this challenging task. Prompt template: ``A realistic photo of a scene with [count] [object class].''}
    \label{fig:header}
    \vspace{-0.2cm}
\end{figure}

Open-sourced diffusion-based text-to-image generation models~\cite{podell2024sdxl, rombach2022sd, sauer2025turbo, chen2024pixartalpha, chen2025pixartdmd} have achieved promising performance in text-image semantic alignment, particularly with recent post-enhancement techniques such as fine-tuning~\cite{clark2024directly, chen2025enhancing, yang2024d3po, wallace2024diffdpo, black2024ddpo, xu2023imagereward, fan2023dpok} or sampling-based, inference strategies~\cite{wallace2023doodl, eyring2024reno, chung2024cfg++, chefer2023attend}. 
But even with such advanced alignment techniques (e.g. ReNO~\cite{eyring2024reno} optimizes image alignment via human preference rewards), they continue to struggle at generating exact counts of objects, as illustrated in Fig.~\ref{fig:header}. Importantly, this challenge persists even in newer, larger models: FLUX-schnell~\cite{flux2024} and SD3.5 Large Turbo~\cite{esser2024scaling} still score below 50\% on the CoCoCount~\cite{binyamin2025mic} benchmark (200 prompts with counts $\leq$ 10). Motivated by this, we tackle the challenge of accurate numeracy in generation with novel count-aware guidance.

\begin{figure*}[t]
     \begin{subfigure}[b]{0.48\textwidth}
         \centering
         \includegraphics[width=\textwidth]{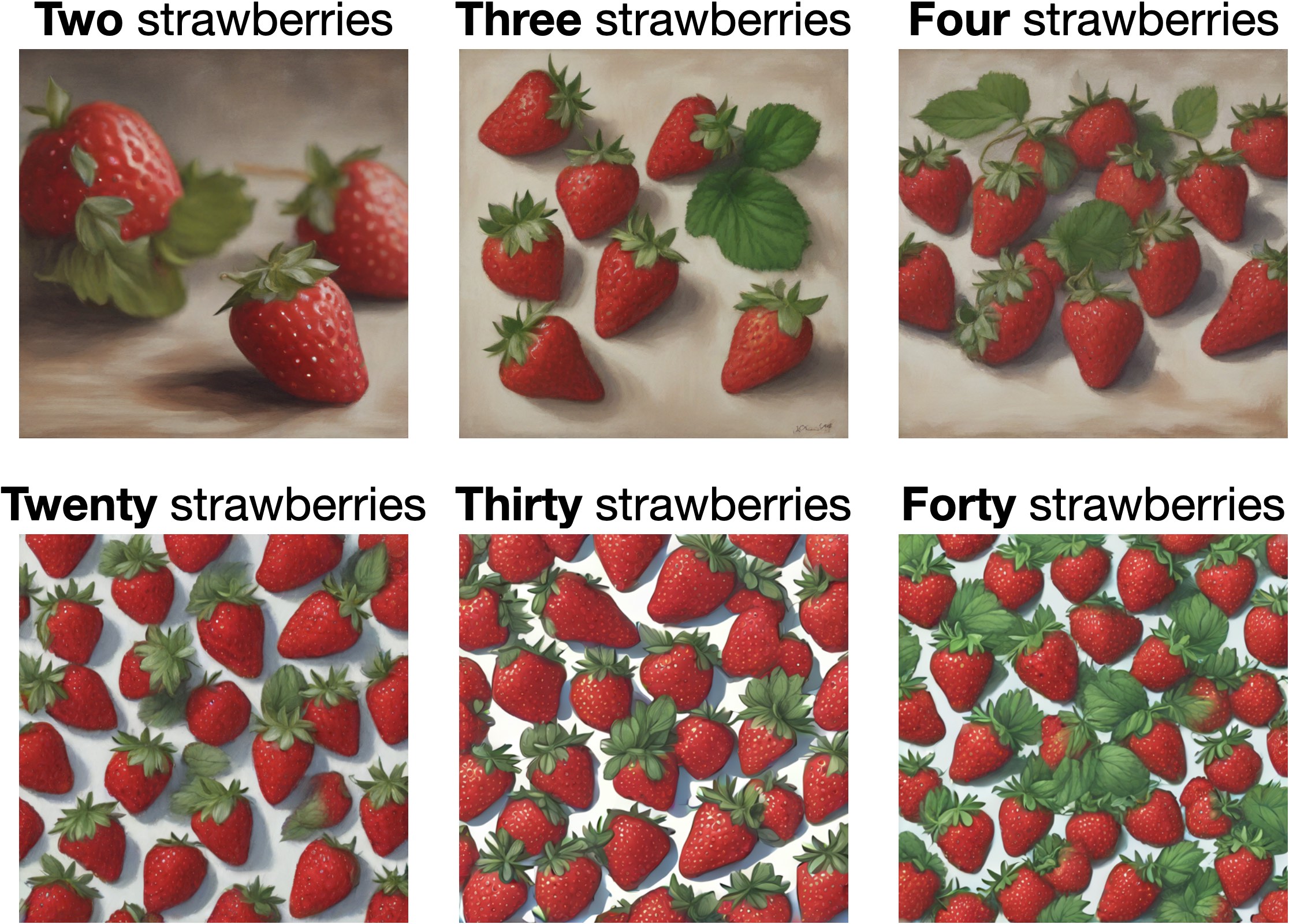}
         \caption{Low and high-density examples with incorrect numeracy, generated by SDXL-Turbo.}
         \label{fig:motiv_low}
     \end{subfigure}
     \hfill
     \begin{subfigure}[b]{0.48\textwidth}
         \centering
         \includegraphics[width=\textwidth]{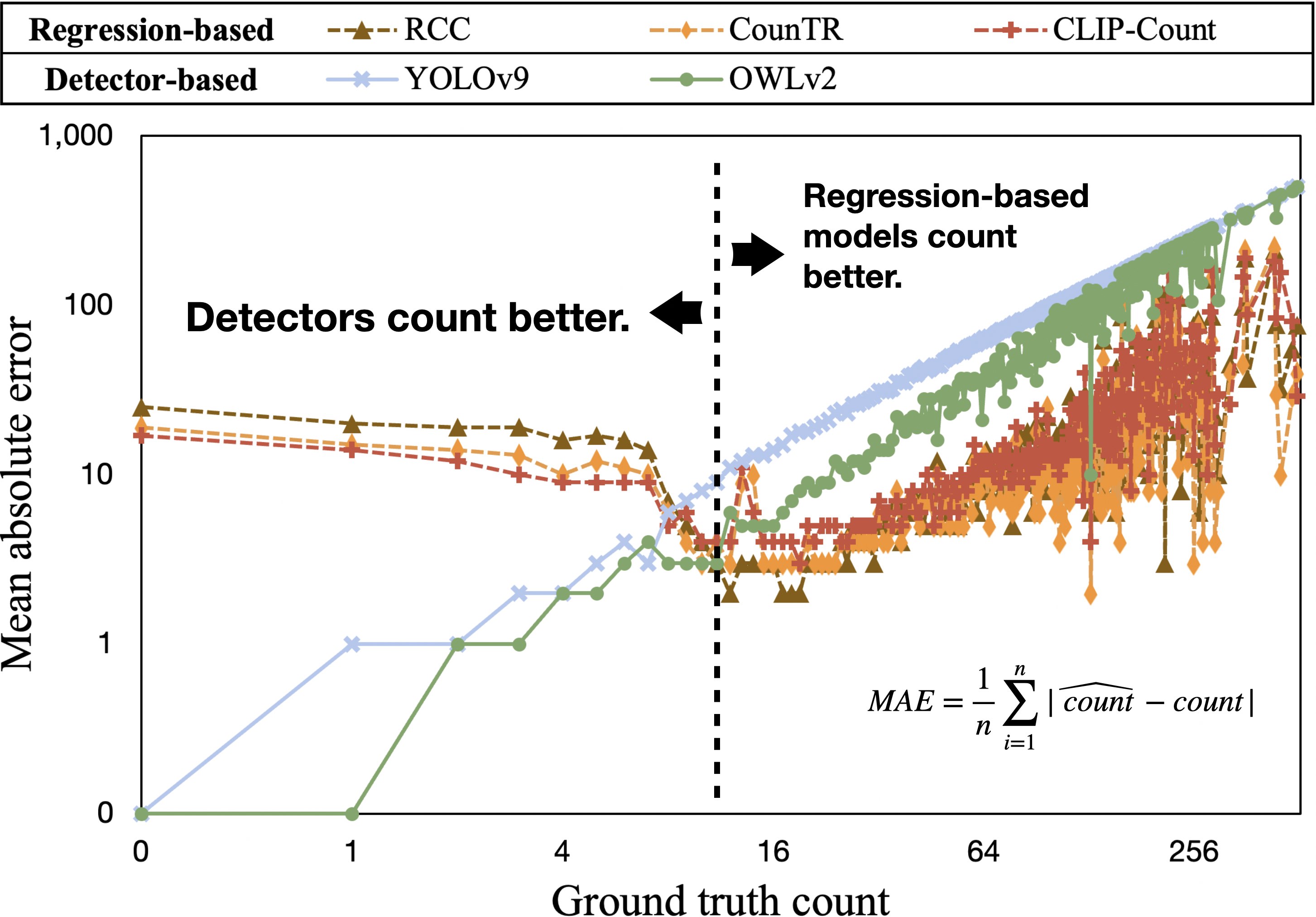}
         \caption{Error by ground truth count on TallyQA~\cite{acharya2019tally} and FSC147~\cite{ranjan2021fsc}.}
         \label{fig:det_better}
     \end{subfigure}
     \caption{\textbf{The low-density setting is where incorrect numeracy is most noticeable and also where detectors count better than regression-based methods.} \emph{But detectors are not differentiable, which precludes them from being used as critics for count correction.}}
    \label{fig:motiv}
    \vspace{-0.15in}
\end{figure*}

\smallsec{Ingredients of numeracy optimization}
Since vanilla T2I models are not explicitly trained to count, existing methods~\cite{kang2025cg, zafar2026tokenopt} introduce auxiliary counting critics to provide additional inference-time supervision via gradients propagated through the generative backbone. To effectively guide generation, these critics must provide both an accurate estimation of the current object count and strong gradients for adding/removing objects as needed, which restricts current approaches to differentiable, regression-based models such as RCC~\cite{hobley2022rcc} and CLIP-Count~\cite{jiang2023clipcount}. As a result, this design excludes the possibility of using stronger detector-based count critics, as they count via bounding box enumeration. As illustrated in Fig.~\ref{fig:det_better}, despite being non-differentiable, detectors outperform regression-based counterparts in low-density object scenarios important to the generative setting, due to their more advanced object localization ability.
To compensate, prior works like Token Optimization (TokenOpt)~\cite{zafar2026tokenopt} rely on a dual-model pipeline, using a regression-based model for the gradient signal and a detector for accurate counting, at the cost of processing the image through both models. Others~\cite{binyamin2025mic} forego the use of counters, instead relying on layout-based methodologies at the cost of image quality. We propose resolving this bottleneck by directly converting existing object detectors into differentiable critics, allowing T2I diffusion models to benefit from stronger counting models for improved numeracy.

\smallsec{The Detector-to-Differentiable (\emph{D2D}) recipe}
The \emph{D2D} framework builds on two key insights that set it apart from prior count correction methods~\cite{kang2025cg, zafar2026tokenopt, binyamin2025mic}. 
First, instead of relying on regression-based counters, we design a high-curvature activation function that relaxes detector-produced discrete object counts into a smooth surrogate. 
Crucially, the shape of this curve is \textit{carefully designed} to extract convergence-friendly gradients, as naive soft relaxations provide insufficient optimization signals. Second, to leverage our new gradient, unlike prior approaches that intervene at intermediate states or denoised predictions along the sampling trajectory, we instead optimize the initial noise using a lightweight test-time tunable module, the Latent Modifier Network (LMN), which learns a perturbation of the Gaussian latent while mixing it with the original noise to maintain proximity to the diffusion prior, thereby providing an implicit regularization effect. This backbone-agnostic design enables broader generalization of our method across diverse diffusion-based T2I architectures, including U-Net~\cite{ronneberger2015unet} and DiT~\cite{peebles2023dit}.

We demonstrate the effectiveness of \emph{D2D} via comprehensive experiments on various generative backbones (i.e., SDXL-Turbo~\cite{sauer2025turbo}, SD-Turbo~\cite{sauer2025turbo}, Pixart-DMD~\cite{chen2025pixartdmd}) and multiple prompt benchmarks (i.e., CoCoCount~\cite{binyamin2025mic}, D2D-Small, D2D-Multi, D2D-Large), covering diverse numeracy scenarios, including single and multiple objects. \emph{D2D} yields the highest numeracy across all multi-step and one-step baselines and benchmarks. On base model SDXL-Turbo, \emph{D2D} effectively corrects 42\% of under-generations (i.e., when the initial generation contains fewer than requested objects) and 40\% of over-generations, nearly or more than 2$\times$ the correction rate of ReNO~\cite{eyring2024reno} and TokenOpt~\cite{zafar2026tokenopt}. In summary, our contributions are as follows:
\begin{itemize}
    \item We highlight the importance of accurate numeracy in T2I generation and propose a framework to convert robust object detectors into differentiable critics for count-correction with a newly designed activation function, overcoming the bottleneck of having to rely on existing regression-based methods with limited accuracy.
    \item We reposition the count-correction problem within the initial noise optimization framework, leveraging structural priors in the Gaussian latent with mixing-based regularization that preserves proximity to the diffusion prior across models.
    \item Our method \emph{D2D} outperforms previous one-step and multi-step count-correction methods by up to \textbf{13.7\%} points (from 30\% with Make It Count to 43.7\% with \emph{D2D} on D2D-Small), with minimal degradation in image quality (Fig.~\ref{fig:header}). On single-object prompts with counts less than or equal to 10, our method introduces less than or comparable computational overhead than baselines.
     \item When applied to a smaller model (SDXL-Turbo~\cite{sauer2025turbo}), \emph{D2D} outperforms larger models FLUX-schnell~\cite{flux2024} and SD3.5 Large Turbo~\cite{esser2024scaling} by up to 9.5\% (from FLUX-schnell's 46.0\% to 55.5\% using \emph{D2D} with SDXL-Turbo on the CoCoCount benchmark).
\end{itemize}

\section{Related Work}
\subsection{Generic alignment-enhancement methods}
As noted in the literature~\cite{black2024ddpo, chen2025enhancing}, the base log-likelihood objective of diffusion models is insufficient on its own to achieve state-of-the-art alignment. To address this, prior works optimize human preference scores via post-enhancement strategies ranging from fine-tuning the U-Net or text encoder~\cite{clark2024directly, xu2023imagereward, yang2024d3po, wallace2024diffdpo, black2024ddpo, fan2023dpok, chen2025enhancing} to inference-time, training-free sampling strategies that update the intermediate latents~\cite{chung2024cfg++, chefer2023attend}. A promising recent line of work~\cite{wallace2023doodl, eyring2024reno} proposes inference-time alignment via initial noise selection, motivated by the presence of semantic/structural priors in the initial noise~\cite{wang2024silent} that influence the semantics/structure of the generated output consistently across diffusion models even with different backbones. But regardless of the specific optimization strategy, the problem remains that generic alignment objectives like human-preference scores are insufficient to steer the model towards count-correct distributions; there persists a significant gap relative to state-of-the-art count-correction methods like \cite{binyamin2025mic}. In our work, we specifically address the challenge of improving numeracy with a new formulation for the objective, as well as adopt initial noise optimization as the method of learning, for the ease with which it can be applied across different backbones and the ability to leverage optimized seeds to complement existing methods, as we demonstrate in experiments.

\subsection{Numeracy correction methods}
Existing count-correction methods leverage two main mechanisms at inference-time to correct count: (1) apply the gradient of external counting models to correct a tunable portion of the generation process, like Counting Guidance~\cite{kang2025cg} and TokenOpt~\cite{zafar2026tokenopt}, or (2) use attention to control the layout of generated instances, like Make It Count~\cite{binyamin2025mic}. Counting Guidance uses the RCC counting model~\cite{hobley2022rcc} to optimize the predicted noises, and TokenOpt uses CLIP-Count~\cite{jiang2023clipcount} to optimize the embedding of a count token injected into the prompt as well as a detector to correct for CLIP-Count's overestimates, increasing the computational overhead (about 2-6$\times$ longer than \emph{D2D} on average). Make It Count~\cite{binyamin2025mic} is an SDXL-specific~\cite{podell2024sdxl} method that uses self-attention features of the U-Net to extract masks of generated instances and cross-attention to enforce a corrected set of masks. These works are either limited by the need to rely on regression-based counters or manner in which they enforce structure at the cost of image quality, a phenomenon documented in \cite{dinh2023rethink, zafar2026tokenopt, patel2025enhancing} and noted in our experiments (Appendix~\ref{app:extended_qual}). Instead, \emph{D2D} leverages a more robust \emph{detector-based} critic that enables more effective correction in the low-density setting.

\subsection{Regression vs. detector-based counting models}
Regression-based counting methods take an input image and estimate count on a continuous scale. Different variations allow for (1) exemplar-based (i.e., count the instances that look similar to the user-provided example), (2) zero-shot (i.e., count the most salient object), and (3) text-prompted counting (i.e., count the text-specified object). Designed to help count high-density images, where continuous-scale predictions are appropriate, they exhibit limited performance in low-density images~\cite{zhang2025improving}, as illustrated in Fig.~\ref{fig:det_better}. On the other hand, our \emph{D2D} critic is derived from detectors which show robust performance given low-density images, which is critical to the generative setting (Fig.~\ref{fig:motiv}). Furthermore, \emph{D2D} can be used to generate objects in the open set by leveraging \emph{open-vocabulary} detectors, like OWLv2~\cite{mindererowlv2}, with minimal modification to detector architecture. In our work, we compare our critic against three regression-based counting methods: RCC~\cite{hobley2022rcc} (zero-shot), CLIP-Count~\cite{jiang2023clipcount} (text-specified), and CounTR~\cite{liu2022countr} (zero-shot).

\begin{figure*}[t]
\centering
    \includegraphics[width=\textwidth]{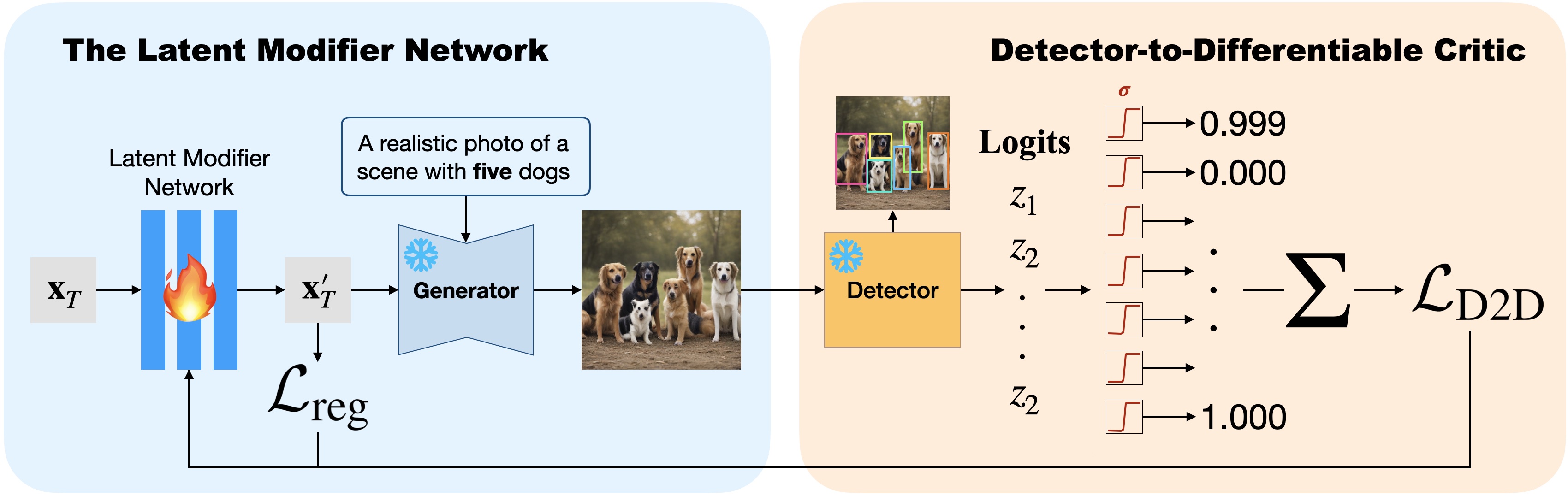}
    \caption{
    \textbf{The \emph{D2D} pipeline for improving T2I numeracy.} \emph{D2D} consists of two main components that work together to improve numeracy: our detector-based count critic guides the Latent Modifier Network (LMN) on how to transform the original initial noise $\mathbf{x}_T$ into a more optimal $\mathbf{x}_T'$. Our count critic uses sigmoid-derived activation functions to convert logits from the frozen detector into gradient signals, which are backpropagated through the frozen generator to update the LMN weights.}
    \label{fig:pipeline}
    \vspace{-0.1in}
\end{figure*}

\section{The \emph{D2D} Framework}

\smallsec{Problem statement} 
Given a pre-trained, one-step T2I model $G_\theta$ and prompt $p$ requesting $N$ counts of an object of class $C$, our goal is to generate an image with exactly $N$ counts of $C$.

\smallsec{Summary of approach}  We propose a detector-based count critic that provides a more effective gradient signal. We then design a mixing method to use that signal to influence the generation process, by leveraging the structural priors in the initial latent which we modify to align with the gradient.

\subsection{Detector-to-Differentiable critic}

Detector $\mathcal{D}$ takes as inputs an object class $C$ and image $I$ and outputs a set of $n$ bboxes $\{B_i |1\leq i \leq n\}$ and logits $\mathbf{z}=\{z_i |1\leq i \leq n\}$. A sigmoid $\sigma(z_i) = \frac{1}{1+e^{-z_i}}$ converts the logits into confidence scores between 0 and 1, filtering the salient bboxes with threshold $\tau$, as follows: $\mathbf{B} = \{B_i | \sigma(z_i) \geq \tau \} = \{B_i | z_i \geq \tau_{z} \}$, where $\tau_z = \sigma^{-1}(\tau)$. The final count is $|\mathbf{B}|$. Our goal is to derive a gradient from $\mathcal{D}$ that can effectively increase/decrease $|\mathbf{B}|$ as needed. Our approach is to first, define a differentiable function $f:\mathbf{z}\in \mathbb{R}^n \mapsto \mathbb{N}$ that extracts the count from the logits $\mathbf{z}$, and second, transform $f$ so its gradient is more amenable to convergence (a step we find crucial), arriving at critic $\mathcal{L}_{\text{D2D}}$.

\smallsec{Extract the count via $\bm{f}$} Counting is discrete in nature, but we convert it into a continuous, differentiable one by borrowing techniques from logistic regression for binary classification, which optimizes the steepness and transition threshold of a sigmoid-curve for discrete 0/1 prediction. We convert each logit into a soft binary indicator of whether to ``count'' the corresponding bbox, by applying to each logit a steep sigmoid curve with transition threshold $\tau_z$ and steepness coefficient $\beta$, with the final differentiable count formulated as a sum of sigmoids: $f_{\beta, \tau_z}(\mathbf{z}) = \sum_{i=1}^n \sigma(\beta \cdot (z_i - \tau_{z}))$. The next challenge is to make this gradient-friendly.

\smallsec{Transform $\bm{f}$ to effectively handle over/under-generation}
An effective critic provides a strong gradient signal above/below $\tau_z$ (our domain of interest) to push logits below or beyond the threshold as needed to erase/add objects in response to over/under-generation.
To illustrate, in over-generated cases, the goal is to prioritize decreasing logits above $\tau$ and allowing some flexibility for those already below $\tau$.
However, by nature of its sigmoidal shape, $f$ has significant plateauing (i.e., weak gradient signals) above and below $\tau$. To improve the gradient steepness in our domain of interest, we scale each sigmoid output by the corresponding logit (Eq.~\ref{eq:diffcritic}), arriving at $\mathcal{L}_{\text{D2D}}$.  Appendix~\ref{app:d2d_scalingform} compares other scaling formulations. At inference, we use $\nabla\mathcal{L}_{\text{D2D}}$ to optimize the generated image.\footnote{Unless otherwise noted, we use $f$ to perform early-stopping once the requested count is met.} 
Extending this to prompts with $m > 1$ object classes, we update Eq.~\ref{eq:diffcritic} to correct each bbox's largest logit, while minimizing all others. Details in Appendix~\ref{app:multi_object}.
\begin{equation}\label{eq:diffcritic}
    \mathcal{L}_{\text{D2D}} = 
    \begin{cases}
    \sum_{i=1}^n \sigma(\beta \cdot (z_i - \tau_{z})) \cdot (z_i - \tau_z), & \text{if }f_{\beta,\tau_z} > N \\
    \sum_{i=1}^n \sigma(\beta \cdot (\tau_{z} - z_i)) \cdot (\tau_z - z_i), & \text{if } f_{\beta,\tau_z} < N 
    \end{cases}
\end{equation}

\subsection{The Latent Modifier Network (LMN)}
Given our count critic, we now turn to the learning method used to optimize this objective. Motivated by the presence of meaningful priors in the initial noise, previous works~\cite{eyring2024reno, wang2024silent} have used generic alignment metrics to tune the initial noise directly. Building on this motivation, we propose the Latent Modifier Network (LMN), a test-time tunable module whose output is mixed with the original noise to determine the optimal initial noise and whose weights are steered using $\mathcal{L}_{\text{D2D}}$.

Given initial noise $\mathbf{x}_T \sim \mathcal{N}(0, \bm{I})$, $\mathbf{x}_T\in \mathbb{R}^d$ and prompt $p$ that requests $N$ counts of an object of class $C$, one-step T2I model $G_\theta$ generates image $I$. Our goal is to find an optimal $\mathbf{x}_T^*$ that produces an image $I^*$ with exactly $N$ of the specified object. To achieve this, we introduce a tunable Latent Modifier Network (LMN) $M_\phi$: a small, 3-layer perceptron, between the initial random latent and T2I model (Fig.~\ref{fig:pipeline}), with input/output dimensions equal to that of the initial latent and whose output dictates how to update $\mathbf{x}_T$. As shown in Eq.~\ref{eq:mod_latent}, the new latent is a weighted sum of $\mathbf{x}_T$ and $M_\phi(\mathbf{x}_T)$, with mixing weight $w = 0.2$. Compared to tuning the initial latent directly, the LMN composes a relatively larger parameter space and introduces an additional form of regularization by preserving a consistent Gaussian component sourced from the original latent throughout all iterations. At inference-time, we tune $\phi$ using $\nabla \mathcal{L}_{\text{D2D}}$ with the goal of correcting the initial noise, and thereby the numeracy, as described in the following section.
\begin{equation}\label{eq:mod_latent}
    \mathbf{x}_T^\prime = w \cdot \mathbf{x}_T + (1-w) \cdot M_\phi(\mathbf{x}_T).
\end{equation}
\subsection{Optimization}
The goal is to find the optimal set of parameters $\phi$ that minimizes the error between the generated and requested count, as follows: $\phi^* = \arg\min_{\phi} |\mathcal{D}(G_\theta(\mathbf{x}_T^\prime)) - N|.$

Since detector $\mathcal{D}$ is non-differentiable, we leverage $\mathcal{L}_{\text{D2D}}$ to optimize $\phi$ iteratively, rendering our final update rule Eq.~\ref{eq:final_update} with regularization term $\mathcal{L}_{\text{reg}}$, learning rate $\eta$, and weights $\alpha$ and $\lambda$. 
\begin{equation}\label{eq:final_update}
    \phi \Leftarrow \phi - \eta \nabla(\alpha\mathcal{L}_{\text{D2D}} + 
    \lambda \mathcal{L}_{\text{reg}}).
\end{equation}
We adaptively rescale the loss to address exploding gradients that may be encountered due to the large number of tunable parameters. During numeracy optimization, we apply a variant of the regularization term used in ReNO~\cite{eyring2024reno}, using the negative log-likelihood of the norm of $\mathbf{x}_T'$ as follows: $\mathcal{L}_{\text{reg}}' = ||\mathbf{x}_T'||^2 / 2 - (d - 1) \cdot \log(||\mathbf{x}_T'||)$. We use $\mathcal{L}_{\text{reg}} = ( a \mathcal{L}_{\text{reg}}' + c)^{10}$, with scaling coefficient $a$ and shift constant $c$. Full algorithm in Appendix~\ref{app:algo}.

\smallsec{$\bm{\phi}$ initialization and calibration} To give $M_\phi$ a good starting point (i.e., initialize the network's initial output distribution to Gaussian), we propose a short, pre-inference alignment stage to be performed one time per base model using only the regularization term. Specifically, we train $M_\phi$ on 100 different randomly sampled latents ($\mathbf{x}_T$) for 200 epochs each (Algorithm~\ref{alg:mphi_init} in the appendix). At inference-time, given a new, randomly sampled $\mathbf{x}_T$ the network has never seen, we introduce a \textasciitilde0.2-second calibration phase to allow the network to adapt to the new input, using only the regularization term. Afterward, we leverage both \emph{D2D} and regularization terms. 

\section{Experiments and Analysis}
\newlength{\oldaboverulesep}
\newlength{\oldbelowrulesep}
\newlength{\oldextrarowheight}
\newlength{\oldtabcolsep}
\setlength{\oldaboverulesep}{\aboverulesep}
\setlength{\oldbelowrulesep}{\belowrulesep}
\setlength{\oldextrarowheight}{\extrarowheight}
\setlength{\oldtabcolsep}{\tabcolsep}

\subsection{Experimental setup}
\smallsec{Benchmarks} Our main experimental setting of single-object, low-density prompts leverages two benchmarks, CoCoCount \cite{binyamin2025mic} and D2D-Small. D2D-Small is a set of 400 prompts created using 40 countable objects from COCO \cite{lin2014coco} with counts ranging from 1-10 and a prompt template adapted from \cite{lian2024lmd}: ``A realistic photo of a scene with [count] [object].'' CoCoCount consists of 200 prompts from 20 COCO classes and requested counts roughly equally split among 2, 3, 4, 5, 7, and 10. Experiments on multi-object or high-density prompts were performed on D2D-Multi (400 prompts with two objects sampled from 40 countable COCO classes, with $N_1, N_2  \leq 10$, and following the template: ``A realistic photo of a scene with [count] [object] and [count] [object]'') and D2D-Large (similarly constructed with counts 11-20).

\smallsec{Base models} We apply \emph{D2D} to three one-step models: SDXL-Turbo~\cite{sauer2025turbo}, SD-Turbo~\cite{sauer2025turbo}, and Pixart-DMD~\cite{chen2025pixartdmd}. SDXL-Turbo and SD-Turbo, respectively distilled from SDXL~\cite{podell2024sdxl} and SD2.1~\cite{rombach2022sd}, have U-Net backbones~\cite{ronneberger2015unet}. Pixart-DMD, distilled from Pixart-$\alpha$~\cite{chen2024pixartalpha}, has a DiT backbone~\cite{peebles2023dit}. We also benchmark \emph{D2D} against larger base models FLUX-schnell~\cite{flux2024} and SD3.5 Large Turbo~\cite{esser2024scaling}.

\smallsec{Comparison of numeracy enhancement methods}
We compare \emph{D2D} against count-correction baselines \textbf{(1) Make It Count}~\cite{binyamin2025mic} (multi-step), which uses attention-based mechanisms to identify/correct object layout via updates to the intermediate latents, \textbf{(2) Counting Guidance}~\cite{kang2025cg} (multi-step), which uses the auxiliary counting model RCC \cite{hobley2022rcc} to correct the predicted noises, and \textbf{(3) TokenOpt}~\cite{zafar2026tokenopt}, a one-step method which injects a count token into the prompt and tunes it using CLIP-Count~\cite{jiang2023clipcount}. We run each baseline following its original experiment setup. Importantly, Make It Count is an SDXL-based method; Counting Guidance (original experiments, codebase) is primarily centered on SD1.4; and TokenOpt is built on SDXL-Turbo, so we evaluate Make It Count on SDXL, Counting Guidance on SD1.4, and TokenOpt on SDXL-Turbo. Furthermore, Make It Count addresses the low-density, single-object setting and TokenOpt addresses the single-object setting, so we only evaluate Make It Count on CoCoCount and D2D-Small and TokenOpt on CoCoCount and D2D-Small/Large.

\smallsec{Measuring the impact of the \emph{D2D} critic}
We demonstrate \emph{D2D} on detectors OWLv2~\cite{mindererowlv2} (open-vocabulary, robust) and YOLOv9~\cite{wang2024yolov9} (high-throughput, trained on COCO~\cite{lin2014coco} objects). We expect a small accuracy-cost tradeoff (i.e., OWLv2 enables superior numeracy with greater computational overhead; YOLOv9 yields slightly lower numeracy but faster inference). We further compare our detector-based critic against regression-based ones (i.e., RCC~\cite{hobley2022rcc}, CLIP-Count~\cite{jiang2023clipcount}, CounTR~\cite{liu2022countr}).

\smallsec{Measuring the impact of the Latent Modifier Network}
The most relevant prior initial noise optimization work is ReNO~\cite{eyring2024reno}, a framework for one-step T2I models that uses the combined gradient of multiple image quality and prompt-image alignment metrics (ImageReward~\cite{xu2023imagereward}, PickScore~\cite{kirstain2023pickscore}, HPSv2~\cite{wu2023hpsv2}, CLIPScore~\cite{hessel2021clipscore}) to optimize semantic alignment and image quality. Instead of tuning an LMN, ReNO directly modifies the initial latent over 20-50 iterations, with a regularization loss term to keep the noise within the initial distribution and gradient clipping to prevent gradient explosion. Though its use of human-preference reward models does improve numeracy relative to the base model, there remains a gap between using such generic objectives and our count-tailored critic (Tab.~\ref{tab:main_one}). A key difference between our method and ReNO's is our introduction of the LMN, which provides an additional opportunity for regularization via mixing and expands the tunable parameter space. To assess the impact of introducing the LMN, we compare our initial noise optimization method with ReNO's, controlling for the loss by swapping out their human-preference models for our \emph{D2D} critic.

\smallsec{Evaluation} Following similar evaluation protocols~\cite{binyamin2025mic, kang2025cg, zafar2026tokenopt}, we use CountGD~\cite{amini2024countgd}, a SOTA counting model built on detector GroundingDINO~\cite{liu2025gdino}, to count the generated objects and compute the proportion of correct images (see Appendix~\ref{app:countgd} for CountGD's counting accuracy compared to other regression/detector-based models). Like \cite{eyring2024reno}, we analyze image-quality/prompt alignment with human-preference-trained models (ImageReward~\cite{xu2023imagereward}, PickScore~\cite{kirstain2023pickscore}, HPSv2~\cite{wu2023hpsv2}), and CLIPScore~\cite{hessel2021clipscore}.

\smallsec{Implementation details} Our main experiments were completed on an L40. For Make It Count~\cite{binyamin2025mic} which requires $>$ 50 GB, we used an A100. Although $\mathcal{L}_{\text{D2D}}$ is piecewise, optimization is generally stable due to several implementation choices. We use momentum, gradient clipping, and reduce the learning rate as needed when the soft count approaches the target (N) to help prevent oscillation near the decision boundary. If the count reaches N, we early-stop without further backpropagation. Average inference time is between 11 and 21 seconds. Additional details in Appendix~\ref{app:hyper}.

\setlength{\aboverulesep}{1pt}
\setlength{\belowrulesep}{0pt}
\setlength{\extrarowheight}{.4ex}
\setlength{\tabcolsep}{6pt}
\begin{table*}[t]
  \caption{\textbf{Quantitative results.} \emph{D2D} outperforms all baselines across all four benchmarks (CoCoCount~\cite{binyamin2025mic} and D2D-Small/Multi/Large), even generalizing across detector variants OWLv2~\cite{mindererowlv2} and YOLOv9~\cite{wang2024yolov9}. \emph{D2D} with YOLOv9 on base model SDXL-Turbo is in \textit{\textbf{bold italics}} to show that while it outperforms all baselines, it is second to using OWLv2. The higher-performing OWLv2 detector is used in all subsequent experiments on SD-Turbo and Pixart-DMD. Standard deviations indicate the significance of our findings. Base models with no post-enhancement highlighted in gray. Avg. over four seeds.}
  \label{tab:main_one}
  \centering
  \begin{small}
  \begin{tabular}{p{17em} l l l l}
    \toprule
    \multicolumn{1}{c}{Method} & CoCoCount & D2D-Small & D2D-Multi & D2D-Large \\ \midrule
    \rowcolor{gray!22} SDXL \small{\cite{podell2024sdxl}} & 24.88 \small{$\pm$1.70} & 16.06 \small{$\pm$1.86} & 2.44 \small{$\pm$0.59} & 1.44 \small{$\pm$0.38} \\
    \:\:\: + Make It Count \small{\cite{binyamin2025mic}} & 46.75 \small{$\pm$2.10} & 30.00 \small{$\pm$1.93} & \multicolumn{1}{c}{-----} & \multicolumn{1}{c}{-----} \\
    \rowcolor{gray!22} SDXL-Turbo \small{\cite{sauer2025turbo}} & 27.38 \small{$\pm$2.69} & 20.31 \small{$\pm$1.95} & 2.12 \small{$\pm$0.83} & 2.56 \small{$\pm$0.55} \\ 
    \:\:\: + ReNO \small{\cite{eyring2024reno}} & 41.88 \small{$\pm$1.03} & 27.50 \small{$\pm$0.68} & 5.31 \small{$\pm$0.38} & 4.69 \small{$\pm$1.25}  \\
    \:\:\: + TokenOpt \small{\cite{zafar2026tokenopt}} & 35.12 \small{$\pm$0.75} & 23.31 \small{$\pm$1.66} & \multicolumn{1}{c}{-----} & 3.94 \small{$\pm$0.72}  \\
    \:\:\: + \emph{D2D} w/ YOLOv9 (Ours) & \textbf{\textit{52.75}} \small{$\pm$1.55} & \textbf{\textit{36.69}} \small{$\pm$2.40} & \textbf{\textit{6.25}} \small{$\pm$1.77} & \textbf{\textit{7.50}} \small{$\pm$1.06}  \\ 
    \:\:\: + \emph{D2D} w/ OWLv2 (Ours) & \textbf{55.62} \small{$\pm$2.72} & \textbf{43.69} \small{$\pm$2.36} & \textbf{9.81} \small{$\pm$0.97} & \textbf{9.94} \small{$\pm$1.57}  \\ 
    \midrule
    \rowcolor{gray!22} SD2.1 \small{\cite{rombach2022sd}} & 32.75 \small{$\pm$1.32} & 24.75 \small{$\pm$2.85} & 4.81 \small{$\pm$1.23} & 2.94 \small{$\pm$0.75}  \\
    \rowcolor{gray!22} SD1.4 \small{\cite{rombach2022sd}} & 27.62 \small{$\pm$4.11} & 16.69 \small{$\pm$2.59} & 2.81 \small{$\pm$0.31} & 2.12 \small{$\pm$0.32}  \\
    \:\:\: + Counting Guidance \small{\cite{kang2025cg}} & 28.38 \small{$\pm$1.11} & 17.12 \small{$\pm$1.69} & 3.38 \small{$\pm$1.16} & 1.88 \small{$\pm$0.60} \\
    \rowcolor{gray!22} SD-Turbo \small{\cite{sauer2025turbo}} & 20.88 \small{$\pm$3.07} & 15.31 \small{$\pm$0.87} & 2.56 \small{$\pm$0.83} &  3.19 \small{$\pm$1.18} \\
    \:\:\: + ReNO \small{\cite{eyring2024reno}} & 43.38 \small{$\pm$3.47} & 32.06 \small{$\pm$0.99} & 8.94 \small{$\pm$1.76} &  4.25 \small{$\pm$1.14} \\
    \:\:\: + \emph{D2D} w/ OWLv2 (Ours) & \textbf{48.38} \small{$\pm$3.09} & \textbf{39.44} \small{$\pm$2.37} & \textbf{10.75} \small{$\pm$1.06} & \textbf{11.44} \small{$\pm$1.98} \\
    \midrule
    \rowcolor{gray!22} Pixart-$\alpha$ \small{\cite{chen2024pixartalpha}} & 19.62 \small{$\pm$1.03} & 14.00 \small{$\pm$1.08} & 1.31 \small{$\pm$0.75} & 1.81 \small{$\pm$0.66}  \\
    \rowcolor{gray!22} Pixart-DMD \small{\cite{chen2025pixartdmd}} & 38.12 \small{$\pm$2.32} & 27.88 \small{$\pm$1.51} & 6.25 \small{$\pm$0.46} & 3.19 \small{$\pm$0.62} \\
    \:\:\: + ReNO \small{\cite{eyring2024reno}} & 44.75 \small{$\pm$1.44} & 37.25 \small{$\pm$1.70} & 9.44 \small{$\pm$0.75} & 4.75 \small{$\pm$0.74} \\
    \:\:\: + \emph{D2D} w/ OWLv2 (Ours) & \textbf{53.25} \small{$\pm$2.40} & \textbf{41.25} \small{$\pm$2.81} & \textbf{13.31} \small{$\pm$1.36} & \textbf{7.62} \small{$\pm$1.18} \\
    \bottomrule
  \end{tabular}
 \end{small}
 \vspace{-0.05in}
\end{table*}
\setlength{\aboverulesep}{\oldaboverulesep}
\setlength{\belowrulesep}{\oldbelowrulesep}
\setlength{\extrarowheight}{\oldextrarowheight}
\setlength{\tabcolsep}{\oldtabcolsep}

\subsection{Numeracy improvements}
\begin{figure}[t]
  \begin{minipage}{0.485\textwidth}
    \centering
    \includegraphics[width=\textwidth]{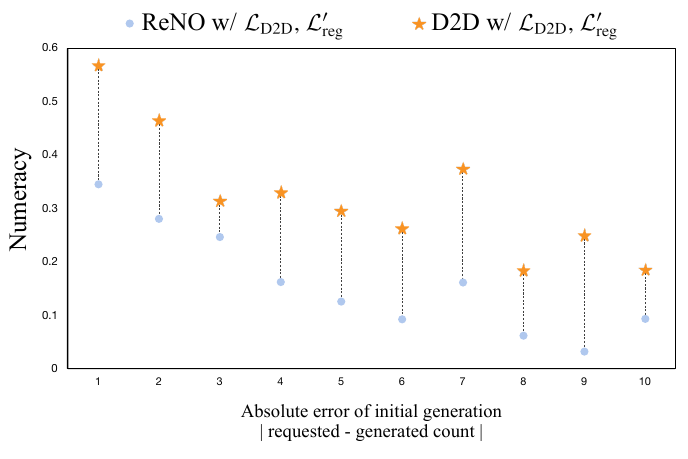}
    \caption{\textbf{The LMN consistently boosts numeracy.} Breakdown of numeracy by absolute error in the initial generation. To compare optimizing via LMN vs. direct tuning of the initial latent, we benchmark \emph{D2D} against ReNO~\cite{eyring2024reno} using $\mathcal{L}_{\text{D2D}}$ and $\mathcal{L}_{\text{reg}}'$ for both, controlling for the number of iterations tuned. On SDXL-Turbo, across CoCoCount~\cite{binyamin2025mic} and D2D-Small. Avg. over four seeds.}
     \label{fig:lmn_by_init_error}
  \end{minipage}
  \hfill
  \begin{minipage}{0.485\textwidth}
    \centering
    \includegraphics[width=0.92\textwidth]{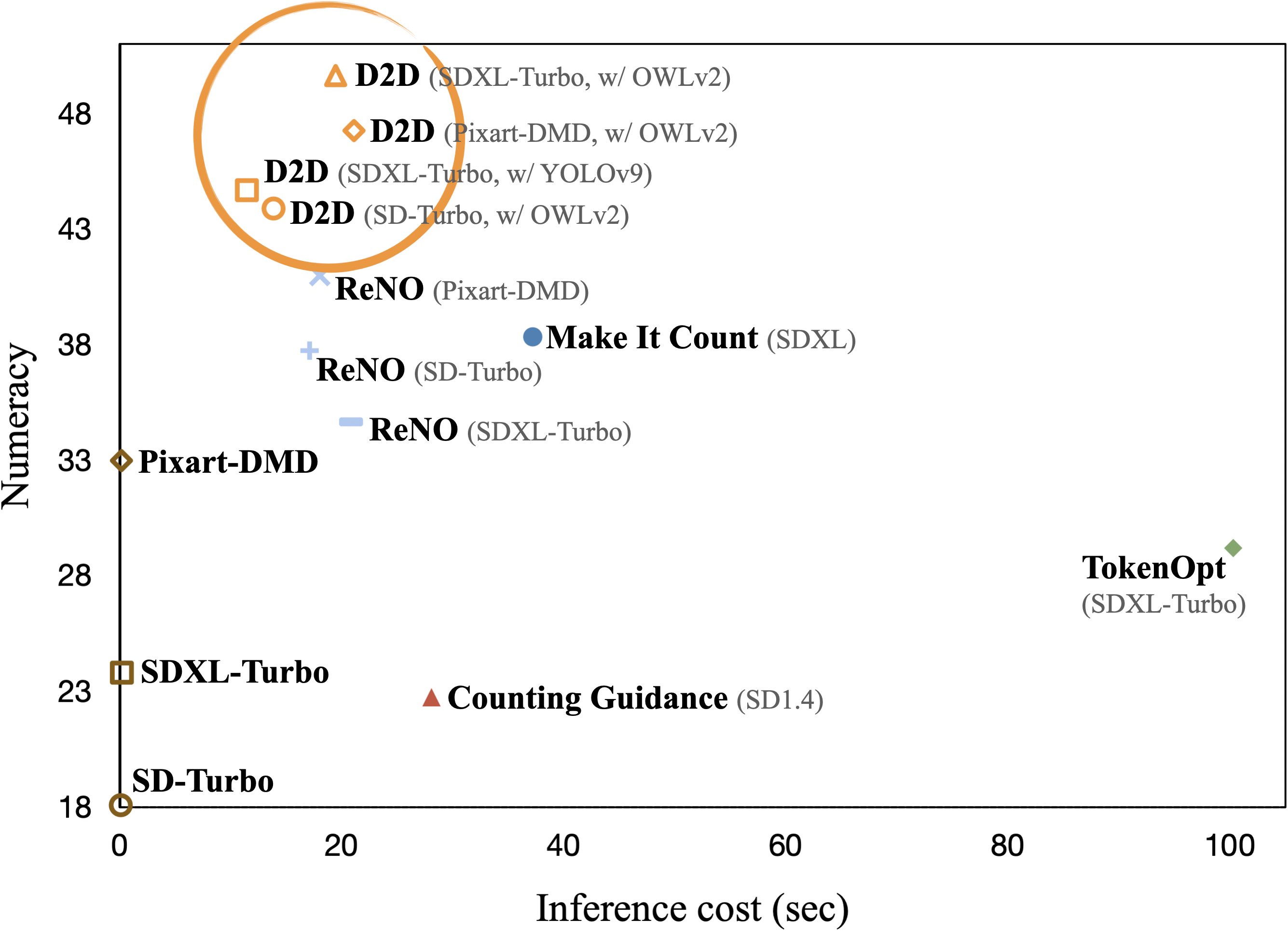}
    \caption{\textbf{Numeracy vs. inference cost by method.} Across base models (SDXL-Turbo, SD-Turbo, Pixart-DMD) and detectors (OWLv2~\cite{mindererowlv2}, YOLOv9~\cite{wang2024yolov9}), \emph{D2D} scores in the top left (i.e. both high-numeracy and low-cost). \emph{D2D} w/ YOLOv9 is even more compute-efficient than w/ OWLv2. Base model/detector in gray. Avg. on CoCoCount~\cite{binyamin2025mic}, D2D-Small, over four seeds.}
         \label{fig:main_inf}
  \end{minipage}
  \vspace{-0.1in}
\end{figure}

Tab.~\ref{tab:main_one} shows our main \emph{D2D}-to-baseline comparisons. Baseline evaluations illustrate that though the prompt setting is relatively simple, generating accurate counts remains challenging. On numeracy, \emph{D2D} consistently outperforms baselines across low-density, single-object, multi-object, and high-density prompts, across base models with U-Net and DiT backbones. On SDXL-Turbo, we demonstrate that performance boosts from \emph{D2D} generalize across OWLv2 and YOLOv9 detector backbones (i.e., the detector used to compute $\mathcal{L}_{\text{D2D}}$), with a small accuracy-cost tradeoff as expected (Fig.~\ref{fig:main_inf}). The robust OWLv2 detector yields higher numeracy with slightly more overhead, while the real-time YOLOv9 detector yields slightly lower (but still high) numeracy with faster inference (in all other experiments, we use the higher-performing OWLv2 backbone unless otherwise noted). Additionally, \emph{D2D} effectively complements baselines, boosting numeracy across all four benchmarks when used in combination with TokenOpt or ReNO (Tab.~\ref{tab:d2d_complement} in appendix). For example, applying \emph{D2D}-optimized seeds to TokenOpt improves numeracy by 13.63\% points, relative to TokenOpt's baseline performance (from 35.12\% to 48.75\%) on CoCoCount.

\smallsec{Improved numeracy on multi-object/high-density prompts}
\emph{D2D} maintains relative improvement even in the more challenging multi-object/high-density settings. Nevertheless, the accuracy drop from low-density benchmarks to D2D-Large illustrates the remaining challenge of correctly generating large counts (e.g., base SDXL-Turbo: from 43.69\% on D2D-Small to 9.94\% on D2D-Large).

\begin{figure}[t]
  \begin{minipage}{0.485\textwidth}
    \centering
    \includegraphics[width=\textwidth]{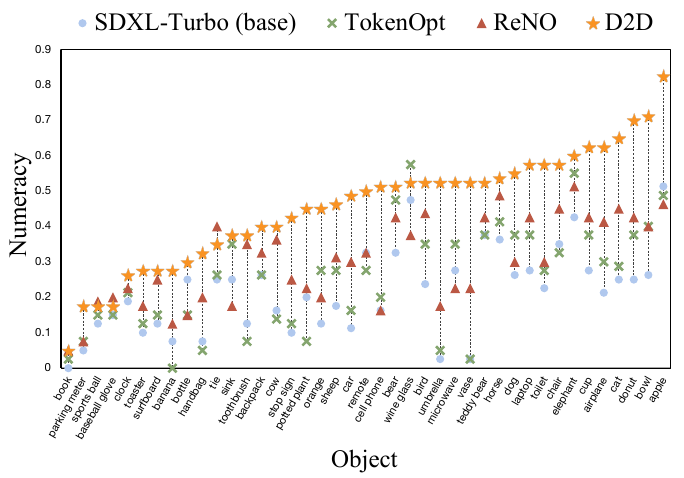}
    \caption{\textbf{\emph{D2D} improves numeracy on the majority of the 41 objects in CoCoCount and D2D-Small.} Evaluated against ReNO~\cite{eyring2024reno} and TokenOpt~\cite{zafar2026tokenopt} on base SDXL-Turbo. Avg. over four seeds.}
    \label{fig:by_class}
  \end{minipage}
  \hfill
  \begin{minipage}{0.485\textwidth}
    \centering
    \includegraphics[width=\textwidth]{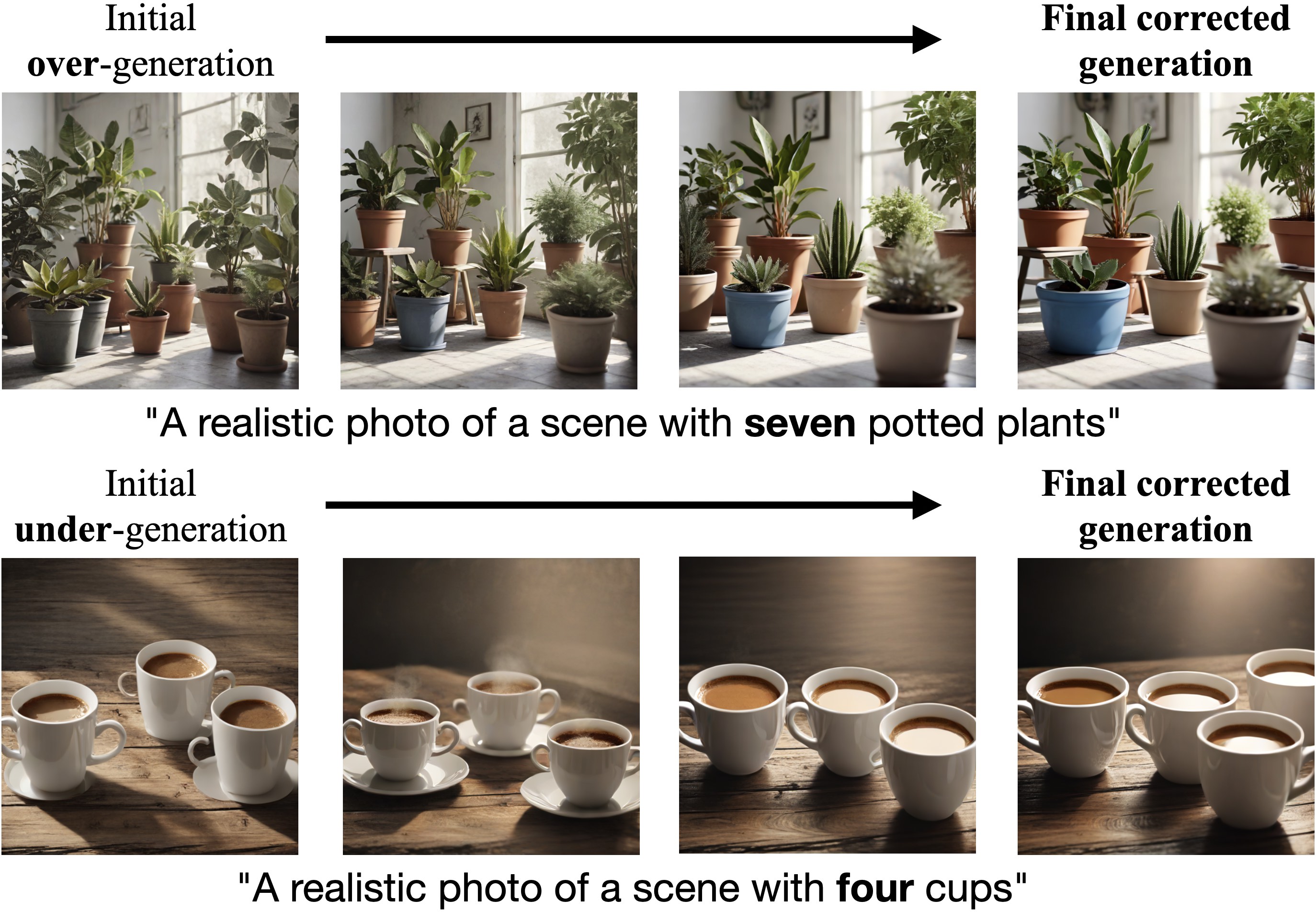}
    \caption{\textbf{\emph{D2D} effectively corrects over and under-generation}. The initial image contains at least two more potted plants/one fewer cup than requested. \emph{D2D} iteratively corrects them, arriving at seven potted plants/four cups, as requested.}
    \label{fig:over_under}
  \end{minipage}
  \vspace{-0.01in}
\end{figure}

\setlength{\tabcolsep}{5pt}
\begin{table}[t]
    \centering
    \begin{minipage}{.485\textwidth}
        \centering
        \captionsetup{skip=7pt}
          \resizebox{\textwidth}{!}{
          \begin{tabular}{p{7em} C{3em} C{3em} C{3em}}
            \toprule
            Initial generation & Over & Under & Correct  \\ \midrule
            TokenOpt~\cite{zafar2026tokenopt} & 13.28 & 25.24 & 69.92  \\
            ReNO~\cite{eyring2024reno} & 23.32 & 25.11 & 62.19 \\
            \emph{D2D} w/ OWLv2 & \textbf{40.13} & \textbf{41.83} & \textbf{72.57}  \\ 
            \bottomrule
          \end{tabular}
          }
        \caption{\textbf{Given the same initial conditions, \emph{D2D} effectively corrects both over and under-generation.} We report the correction rate of initial over/under-generations, as well as the proportion of correct generations that were maintained. On SDXL-Turbo, across CoCoCount~\cite{binyamin2025mic} and D2D-Small. Avg. over four seeds.}
          \label{tab:over_and_under}
    \end{minipage}
    \hfill 
    \begin{minipage}{.485\textwidth}
          \centering
          \captionsetup{skip=7pt}
           \resizebox{\textwidth}{!}{
        \begin{tabular}{p{14.7em} C{8em}}
        \toprule
        \multicolumn{1}{c}{Method} & Numeracy \\ \midrule
        FLUX-schnell~\cite{flux2024} (1-step) &  46.0  \\ 
        FLUX-schnell~\cite{flux2024} (4-step) & 47.5  \\ 
        SD3.5 Large Turbo~\cite{esser2024scaling} (4-step) &  48.0   \\ \midrule
        SDXL-Turbo~\cite{sauer2025turbo} & 30.0   \\ 
        SDXL-Turbo~\cite{sauer2025turbo} + \emph{D2D}            &  \textbf{55.5}   \\ 
        \bottomrule
        \end{tabular}
        }
        \caption{\textbf{\emph{D2D} on a weaker base model outperforms stronger base models.} Integrating \emph{D2D} with a smaller model like SDXL-Turbo outperforms stronger base models such as SD3.5 Large Turbo and FLUX-schnell. On CoCoCount~\cite{binyamin2025mic}, seed=0.
        }
          \label{tab:newer}
    \end{minipage}
    \vspace{-0.15in}
\end{table}
\setlength{\tabcolsep}{\oldtabcolsep}

\smallsec{$\bm{\mathcal{L}_{\textbf{D2D}}}$ effectively boosts numeracy across all classes} \emph{D2D} improves numeracy over the base model across all 41 object categories in CoCoCount and D2D-Small (Fig.~\ref{fig:by_class}), spanning a large variety (e.g., apples, elephants, cars, etc.). Upon applying \textit{D2D} to SDXL-Turbo, umbrella and vase are the two classes that see the most improvement, each jumping from 2.50\% (SDXL-Turbo base) to 52.50\% (\emph{D2D}) accuracy. Wine glass and bottle, both of which are (semi)transparent objects, are among the classes that see the least improvement (47.50\% to 52.50\% and 25.00\% to 30.00\% accuracy, respectively), which may suggest a future direction where detectors are fine-tuned on more difficult classes, or similar, with the purpose of generating highly-tailored scenes of objects.

\smallsec{\emph{D2D} best handles over and under-generation} Tab.~\ref{tab:over_and_under} breaks down results by the numeracy of the initial generation $I$, illustrating how well different methods are able to \textit{correct} over/under-generation while \textit{maintaining} the numeracy of already-correct images. We compare TokenOpt, ReNO, and \emph{D2D} on base model SDXL-Turbo, across benchmarks CoCoCount and D2D-Small. \emph{D2D} has the highest correction rate, correcting 40.13\% of over-generations and 41.83\% of under-generations, which is at least 16\% points over the baselines, while maintaining 72.57\% of already-correct generations, which is also more than both TokenOpt and ReNO. Fig.~\ref{fig:over_under} illustrates \emph{D2D}'s iterative correction process on two sample prompts, going from at least nine potted plants to the requested seven and from three cups to the requested four. Additional qualitative examples in Appendix~\ref{app:add_qual}.

\smallsec{\emph{D2D} vs. larger models} Tab.~\ref{tab:newer} compares the numeracy of newer, larger models like FLUX-schnell~\cite{flux2024} and SD3.5 Large Turbo~\cite{esser2024scaling} against \emph{D2D} on the smaller base model SDXL-Turbo~\cite{sauer2025turbo}. Even on a weaker base model (30.0\% accuracy), \emph{D2D} (55.5\%) yields a \textbf{+9.5\%}, \textbf{+8\%}, \textbf{+7.5\% point improvement} over FLUX-schnell (1-step: 46.0\%; 4-step: 47.5\%) and SD3.5 Large Turbo (48.0\%).

\subsection{Analysis of \emph{D2D} critic}\label{sec:main_analysis}

\begin{table}[t]
    \centering
    \begin{minipage}{.485\textwidth}
        \captionsetup{skip=7pt}
            \setlength{\tabcolsep}{6pt}
          \resizebox{0.99\textwidth}{!}{
          \begin{tabular}{c @{} c c c c }
            \toprule
            $\tau$ & \hspace{1em}0.1\hspace{1em} & 0.2\hspace{1em} & 0.5\hspace{1em} & 0.8 \\
            \midrule
            CountGD~\cite{amini2024countgd} & \hspace{1em}51.50\hspace{1em} & \textbf{55.50}\hspace{1em} & 43.50\hspace{1em} & 32.50 \\
            \bottomrule
          \end{tabular}}
          
          \vspace{0.9em}
          
          \resizebox{\textwidth}{!}{
          \begin{tabular}{c c c c c c}
            \toprule
            $\beta$ & 1 & 10 & 100 & 300 & 400 \\
            \midrule
            CountGD~\cite{amini2024countgd} & 43.00 & 40.00 & 52.00 & \textbf{55.50} & 52.50 \\
            \bottomrule
          \end{tabular}
          }
        \caption{\textbf{Ablation study on key hyperparameters $\bm\tau$ and $\bm\beta$.} Detector threshold $\tau=0.2$ is optimal. A lower $\tau$ (which counts low-confidence bboxes) and higher $\tau$ (which potentially discards actually-legitimate bboxes) result in drops in numeracy. Steepness coefficient $\beta=300$ is optimal. Tested on CoCoCount~\cite{binyamin2025mic}, seed=0.}
        \label{tab:abl_hyperparam}
    \end{minipage}
    \hfill 
    \begin{minipage}{.485\textwidth}
          \centering
          \captionsetup{skip=7pt}
          \setlength{\aboverulesep}{2pt}
            \setlength{\belowrulesep}{2pt}
            \setlength{\extrarowheight}{.4ex}
            \setlength{\tabcolsep}{1.2pt}
           \resizebox{\textwidth}{!}{
          \begin{tabular}{p{6.5em} c c c c}
            \toprule
            \multicolumn{1}{c}{Count Critic} & CoCoCount & D2D-Small & D2D-Multi & D2D-Large \\ \midrule
            RCC & 37.75 & 26.38 & ----- & 4.25  \\
            CounTR & 38.38 & 25.62 & ----- & 5.31  \\
            CLIP-Count & 40.00 & 25.88 & 5.19 & 6.38  \\
            $f$ (OWLv2) & 32.00 & 20.75 & 3.06 & 3.38  \\
            $\mathcal{L}_{\text{D2D}}$ (OWLv2) & \textbf{55.62} & \textbf{43.69} & \textbf{9.81} & \textbf{9.94}  \\
            \bottomrule
          \end{tabular}
        }
        \caption{\textbf{Comparison of count critics. Using $\bm{f}$ alone yields weak gradients and poor convergence, but $\bm{\mathcal{L}_{\text{D2D}}}$'s logit-scaled form enables effective optimization and higher numeracy.} $\mathcal{L}_{\text{D2D}}$ outperforms RCC~\cite{hobley2022rcc}, CounTR~\cite{liu2022countr}, and CLIP-Count~\cite{jiang2023clipcount}. On SDXL-Turbo. Avg. over four seeds.}
        \label{tab:our_critic_versions}
    \end{minipage}
\end{table}
\setlength{\aboverulesep}{\oldaboverulesep}
\setlength{\belowrulesep}{\oldbelowrulesep}
\setlength{\extrarowheight}{\oldextrarowheight}
\setlength{\tabcolsep}{\oldtabcolsep}

\smallsec{\emph{D2D} vs. regression-based counters}
Tab.~\ref{tab:our_critic_versions} compares our critic against existing regression-based ones and confirms the importance of $\mathcal{L}_{\text{D2D}}$'s convergence-friendly form. Across all four benchmarks, our detector-based critic outperforms regression-based methods RCC, CLIP-Count, and CounTR on numeracy (e.g., ours reaches 55.62\% when the max score reached by any regression-based one is 40\% on CoCoCount). Notably, $\mathcal{L}_{\text{D2D}}$ outperforms the others even on the high-density benchmark D2D-Large, though regression-based methods outperform detectors in the non-generative, counting setting (Fig.~\ref{fig:det_better}). Notably, $f_{\beta,\tau_z}$ yields the lowest numeracy, which indicates that though it composes the mathematical backbone of $\mathcal{L}_{\text{D2D}}$, its sigmoid saturation leads to weak gradients and poor convergence when used directly as the critic (Tab.~\ref{tab:our_critic_versions}). In contrast, $\mathcal{L}_{\text{D2D}}$'s logit-scaled formulation preserves gradient magnitude and improves convergence, making it crucial to \emph{D2D}'s performance.

\vspace{0.1in}
\smallsec{Impact of key hyperparameters $\bm{\tau}$ and $\bm{\beta}$}
We report our key hyperparameter studies on values for $\tau$ (detector threshold) and $\beta$ (steepness coefficient), conducted using base model SDXL-Turbo on benchmark CoCoCount (seed=0) on an A6000/L40. Results (Tab.~\ref{tab:abl_hyperparam}) show that $\tau=0.2$, $\beta=300$ are optimal, which we use in all our main experiments.

\subsection{Analysis of the Latent Modifier Network}

\smallsec{Impact of the LMN on numeracy}
Next, we assess the impact of introducing the LMN, a module whose output is mixed with the original noise to arrive at the optimal noise, by comparing our method with ReNO's, controlling for the optimization objectives used ($\mathcal{L}_{\text{D2D}}$, $\mathcal{L}_{\text{reg}}'$) and number of iterations tuned. Fig.~\ref{fig:lmn_by_init_error} shows the LMN provides robust numeracy gains even as the magnitude of the initial error increases; numeracy jumps 10\% points on CoCoCount and D2D-Small from 43.25\% to 53.88\% and from 32.00\% to 42.44\%, respectively (Tab.~\ref{tab:mphi} in appendix).

\smallsec{Optimal balance between count correction and regularization} $w$ is the mixing weight that determines the extent to which the original noise $\mathbf{x}_T$ should be preserved in the corrected noise $\mathbf{x}_T'$. When $w$ is small (e.g., $w=0$) the final noise consists largely of the count correction signal from $M_\phi$, at the risk of steering away from the initial noise distribution. When $w$ is large (e.g., $w=1$), the initial noise is largely preserved due to heavy regularization (i.e., minimal count correction). The key is to find a $w$ that is high enough to preserve image quality and low enough to improve numeracy. As shown in Tab.~\ref{tab:abl_w}, $w=0$ is numeracy-wise the best, followed by $w=0.2$, but results in patchy visual artifacts, relative to $w=0.2$ (Fig.~\ref{fig:w_abl_qual}). Hence, we use $w=0.2$ in our main experiments.

\smallsec{\emph{D2D} is both high-numeracy and low-cost} Our method averages between 11 and 21 seconds in inference cost, compared to counting baselines, which average upwards of 28 to 100 seconds (Fig.~\ref{fig:main_inf}).
ImageReward, PickScore, HPSv2, and CLIPScore metrics in Fig.~\ref{fig:main_qual} show \emph{D2D}'s image quality and overall prompt alignment is comparable to counting baselines and even surpasses multi-step baselines in many cases, including the layout control-based Make It Count (MIC). For example, SDXL-Turbo + \emph{D2D} (OWLv2) yields ImageReward 0.51 (MIC: 0.30), PickScore 21.98 (MIC: 21.48), and HPSv2 0.28 (MIC: 0.26) on D2D-Small.

\section{Conclusion and Discussion}

In this work, we address the challenge of correcting numeracy in generation. We identify a central limitation of previous methods, specifically their reliance on differentiable, regression-based counting models as critics. We propose \emph{D2D}, a novel way to convert more robust detectors into differentiable count critics, which we then use to optimize the initial noise at inference-time to improve numeracy. Our method yields the highest numeracy across prompt scenarios, including low-density, single-object, multi-object, high-density settings, effectively correcting both over and under-generation, with minimal additions to temporal overhead (i.e., the additional inference time introduced beyond the base model's generation time) and minimal degradation in image quality.

\smallsec{Limitation and future directions} While our method exhibits significant improvements in numeracy, high-density scenarios remain challenging. Given regression-based methods are more appropriate in this setting, a future direction may explore how to adapt them into the generative setting. \emph{D2D} is not a layout-control approach and so is limited in more fine-grained control (e.g., object placement). But future directions may explore using \emph{D2D} to perform more complex tasks, such as object positioning and attribute binding, by leveraging detectors that can robustly work with prompts specifying objects and associated attributes.

\section*{Acknowledgements}

This research was partially supported by an Amazon Research Award.
Any opinions, findings, and conclusions or recommendations expressed in this material are those of the author(s) and do not necessarily reflect the views of Amazon. YZ also acknowledges the support of the French National Research Agency (ANR) via the ``MiMoGC'' JCJC grant (ANR-26-CE23-7469). We are also grateful to Princeton Research Computing for compute resources.

\newpage

{
\small
\bibliographystyle{unsrtnat}
\bibliography{neurips_2026}

@article{wei2026enhancing,
  title={Enhancing {MMDiT}-based text-to-image models for similar subject generation},
  author={Wei, Tianyi and Chen, Dongdong and Zhou, Yifan and Pan, Xingang},
  journal={IEEE Transactions on Pattern Analysis and Machine Intelligence},
  year={2026},
  publisher={IEEE}
}

@inproceedings{xiaofeng2026intricate,
  title={The Intricate Dance of Prompt Complexity, Quality, Diversity and Consistency in {T2I} Models},
  author={Xiaofeng, Zhang and Courville, Aaron and Drozdzal, Michal and Romero-Soriano, Adriana},
  booktitle={International Conference on Learning Representations},
  volume={2026},
  pages={50001--50038},
  year={2026}
}

@inbook{krippendorff2019reliability,
  author    = {Krippendorff, Klaus},
  title     = {Content Analysis: An Introduction to Its Methodology},
  chapter   = {Reliability},
  edition   = {Fourth},
  pages     = {277--360},
  publisher = {SAGE Publications, Inc.},
  address   = {Thousand Oaks, California},
  year      = {2019},
  doi       = {10.4135/9781071878781.n13}
}

@InProceedings{zeng2025yolocount,
    author    = {Zeng, Guanning and Zhang, Xiang and Wang, Zirui and Xu, Haiyang and Chen, Zeyuan and Li, Bingnan and Tu, Zhuowen},
    title     = {{YOLO-Count}: Differentiable Object Counting for Text-to-Image Generation},
    booktitle = {Proceedings of the IEEE/CVF International Conference on Computer Vision (ICCV)},
    month     = {October},
    year      = {2025},
    pages     = {16765-16775}
}

@inproceedings{
esser2024scaling,
title={Scaling Rectified Flow Transformers for High-Resolution Image Synthesis},
author={Patrick Esser and Sumith Kulal and Andreas Blattmann and Rahim Entezari and Jonas M{\"u}ller and Harry Saini and Yam Levi and Dominik Lorenz and Axel Sauer and Frederic Boesel and Dustin Podell and Tim Dockhorn and Zion English and Robin Rombach},
booktitle={Forty-first International Conference on Machine Learning},
year={2024},
url={https://openreview.net/forum?id=FPnUhsQJ5B}
}

@misc{flux2024,
    author={Black Forest Labs},
    title={{FLUX}},
    year={2024},
    howpublished={\url{https://github.com/black-forest-labs/flux}},
}

@article{chen2025sana,
  title={{SANA-Sprint}: One-step diffusion with continuous-time consistency distillation},
  author={Chen, Junsong and Xue, Shuchen and Zhao, Yuyang and Yu, Jincheng and Paul, Sayak and Chen, Junyu and Cai, Han and Han, Song and Xie, Enze},
  journal={arXiv preprint arXiv:2503.09641},
  year={2025}
}

@article{
friedman2023vendi,
title={The {Vendi} Score: A Diversity Evaluation Metric for Machine Learning},
author={Dan Friedman and Adji Bousso Dieng},
journal={Transactions on Machine Learning Research},
issn={2835-8856},
year={2023},
url={https://openreview.net/forum?id=g97OHbQyk1},
note={}
}

@inproceedings{
tang2025dno,
title={Inference-Time Alignment of Diffusion Models with Direct Noise Optimization},
author={Zhiwei Tang and Jiangweizhi Peng and Jiasheng Tang and Mingyi Hong and Fan Wang and Tsung-Hui Chang},
booktitle={Forty-second International Conference on Machine Learning},
year={2025},
url={https://openreview.net/forum?id=JpbqiD7n9r}
}

@InProceedings{wang2024silent,
    author    = {Wang, Ruoyu and Huang, Huayang and Zhu, Ye and Russakovsky, Olga and Wu, Yu},
    title     = {The Silent Assistant: {NoiseQuery} as Implicit Guidance for Goal-Driven Image Generation},
    booktitle = {Proceedings of the IEEE/CVF International Conference on Computer Vision (ICCV)},
    month     = {October},
    year      = {2025},
    pages     = {17618-17628}
}

@InProceedings{ronneberger2015unet,
author="Ronneberger, Olaf
and Fischer, Philipp
and Brox, Thomas",
editor="Navab, Nassir
and Hornegger, Joachim
and Wells, William M.
and Frangi, Alejandro F.",
title="{U-Net}: Convolutional Networks for Biomedical Image Segmentation",
booktitle="Medical Image Computing and Computer-Assisted Intervention -- MICCAI 2015",
year="2015",
publisher="Springer International Publishing",
address="Cham",
pages="234--241",
isbn="978-3-319-24574-4"
}

@InProceedings{peebles2023dit,
    author    = {Peebles, William and Xie, Saining},
    title     = {Scalable Diffusion Models with Transformers},
    booktitle = {Proceedings of the IEEE/CVF International Conference on Computer Vision (ICCV)},
    month     = {October},
    year      = {2023},
    pages     = {4195-4205}
}

@InProceedings{liu2025gdino,
author="Liu, Shilong
and Zeng, Zhaoyang
and Ren, Tianhe
and Li, Feng
and Zhang, Hao
and Yang, Jie
and Jiang, Qing
and Li, Chunyuan
and Yang, Jianwei
and Su, Hang
and Zhu, Jun
and Zhang, Lei",
title="{Grounding DINO}: Marrying {DINO} with Grounded Pre-training for Open-Set Object Detection",
booktitle="Computer Vision -- ECCV 2024",
year="2025",
pages="38--55"
}

@article{acharya2019tally, 
title={{TallyQA}: Answering Complex Counting Questions}, 
volume={33}, 
DOI={10.1609/aaai.v33i01.33018076}, 
number={01}, 
journal={Proceedings of the AAAI Conference on Artificial Intelligence}, 
author={Acharya, Manoj and Kafle, Kushal and Kanan, Christopher}, 
year={2019}, 
month={Jul.}, 
pages={8076-8084} }

@InProceedings{liu2018decide,
author = {Liu, Jiang and Gao, Chenqiang and Meng, Deyu and Hauptmann, Alexander G.},
title = {{DecideNet}: Counting Varying Density Crowds Through Attention Guided Detection and Density Estimation},
booktitle = {Proceedings of the IEEE Conference on Computer Vision and Pattern Recognition (CVPR)},
month = {June},
year = {2018}
}

@inproceedings{amini2024countgd,
 author = {Amini-Naieni, Niki and Han, Tengda and Zisserman, Andrew},
 booktitle = {Advances in Neural Information Processing Systems},
 editor = {A. Globerson and L. Mackey and D. Belgrave and A. Fan and U. Paquet and J. Tomczak and C. Zhang},
 pages = {48810--48837},
 publisher = {Curran Associates, Inc.},
 title = {{CountGD}: Multi-Modal Open-World Counting},
 volume = {37},
 year = {2024}
}

@InProceedings{ranjan2021fsc,
    author    = {Ranjan, Viresh and Sharma, Udbhav and Nguyen, Thu and Hoai, Minh},
    title     = {Learning To Count Everything},
    booktitle = {Proceedings of the IEEE/CVF Conference on Computer Vision and Pattern Recognition (CVPR)},
    month     = {June},
    year      = {2021},
    pages     = {3394-3403}
}

@InProceedings{gupta2019lvis,
author = {Gupta, Agrim and Dollar, Piotr and Girshick, Ross},
title = {{LVIS}: A Dataset for Large Vocabulary Instance Segmentation},
booktitle = {Proceedings of the IEEE/CVF Conference on Computer Vision and Pattern Recognition (CVPR)},
month = {June},
year = {2019}
}

@inproceedings{
chen2024pixartalpha,
title={{PixArt}-{$\alpha$}: Fast Training of Diffusion Transformer for Photorealistic Text-to-Image Synthesis},
author={Junsong Chen and Jincheng YU and Chongjian GE and Lewei Yao and Enze Xie and Zhongdao Wang and James Kwok and Ping Luo and Huchuan Lu and Zhenguo Li},
booktitle={The Twelfth International Conference on Learning Representations},
year={2024}
}

@InProceedings{rombach2022sd,
    author    = {Rombach, Robin and Blattmann, Andreas and Lorenz, Dominik and Esser, Patrick and Ommer, Bj\"orn},
    title     = {High-Resolution Image Synthesis With Latent Diffusion Models},
    booktitle = {Proceedings of the IEEE/CVF Conference on Computer Vision and Pattern Recognition (CVPR)},
    month     = {June},
    year      = {2022},
    pages     = {10684-10695}
}

@inproceedings{
podell2024sdxl,
title={{SDXL}: Improving Latent Diffusion Models for High-Resolution Image Synthesis},
author={Dustin Podell and Zion English and Kyle Lacey and Andreas Blattmann and Tim Dockhorn and Jonas M{\"u}ller and Joe Penna and Robin Rombach},
booktitle={The Twelfth International Conference on Learning Representations},
year={2024}
}

@InProceedings{chen2025pixartdmd,
author="Chen, Junsong
and Ge, Chongjian
and Xie, Enze
and Wu, Yue
and Yao, Lewei
and Ren, Xiaozhe
and Wang, Zhongdao
and Luo, Ping
and Lu, Huchuan
and Li, Zhenguo",
editor="Leonardis, Ale{\v{s}}
and Ricci, Elisa
and Roth, Stefan
and Russakovsky, Olga
and Sattler, Torsten
and Varol, G{\"u}l",
title="PIXART-{$\Sigma$}: Weak-to-Strong Training of Diffusion Transformer for {4K} Text-to-Image Generation",
booktitle="Computer Vision -- ECCV 2024",
year="2025",
publisher="Springer Nature Switzerland",
address="Cham",
pages="74--91",
isbn="978-3-031-73411-3"
}

@InProceedings{sauer2025turbo,
author="Sauer, Axel
and Lorenz, Dominik
and Blattmann, Andreas
and Rombach, Robin",
editor="Leonardis, Ale{\v{s}}
and Ricci, Elisa
and Roth, Stefan
and Russakovsky, Olga
and Sattler, Torsten
and Varol, G{\"u}l",
title="Adversarial Diffusion Distillation",
booktitle="Computer Vision -- ECCV 2024",
year="2025",
publisher="Springer Nature Switzerland",
address="Cham",
pages="87--103",
isbn="978-3-031-73016-0"
}

@InProceedings{lin2014coco,
author="Lin, Tsung-Yi
and Maire, Michael
and Belongie, Serge
and Hays, James
and Perona, Pietro
and Ramanan, Deva
and Doll{\'a}r, Piotr
and Zitnick, C. Lawrence",
editor="Fleet, David
and Pajdla, Tomas
and Schiele, Bernt
and Tuytelaars, Tinne",
title="{Microsoft COCO}: Common Objects in Context",
booktitle="Computer Vision -- ECCV 2014",
year="2014",
publisher="Springer International Publishing",
address="Cham",
pages="740--755",
isbn="978-3-319-10602-1"
}

@inproceedings{liu2022countr,
  author = {Liu, Chang and Zhong, Yujie and Zisserman, Andrew and Xie, Weidi},
  title = {{CounTR}: Transformer-based Generalised Visual Counting},
  booktitle={British Machine Vision Conference (BMVC)},
  year = {2022}
}

@inproceedings{jiang2023clipcount,
author = {Jiang, Ruixiang and Liu, Lingbo and Chen, Changwen},
title = "{CLIP-C}ount: Towards Text-Guided Zero-Shot Object Counting",
year = {2023},
isbn = {9798400701085},
publisher = {Association for Computing Machinery},
address = {New York, NY, USA},
doi = {10.1145/3581783.3611789},
booktitle = {Proceedings of the 31st ACM International Conference on Multimedia},
pages = {4535--4545},
numpages = {11},
location = {Ottawa ON, Canada},
series = {MM '23}
}

@article{hobley2022rcc,
  title={Learning to count anything: Reference-less class-agnostic counting with weak supervision},
  author={Hobley, Michael and Prisacariu, Victor},
  journal={arXiv preprint arXiv:2205.10203},
  year={2022}
}

@inproceedings{hessel2021clipscore,
    title = "{CLIPS}core: A Reference-free Evaluation Metric for Image Captioning",
    author = "Hessel, Jack  and
      Holtzman, Ari  and
      Forbes, Maxwell  and
      Le Bras, Ronan  and
      Choi, Yejin",
    editor = "Moens, Marie-Francine  and
      Huang, Xuanjing  and
      Specia, Lucia  and
      Yih, Scott Wen-tau",
    booktitle = "Proceedings of the 2021 Conference on Empirical Methods in Natural Language Processing",
    month = nov,
    year = "2021",
    address = "Online and Punta Cana, Dominican Republic",
    publisher = "Association for Computational Linguistics",
    doi = "10.18653/v1/2021.emnlp-main.595",
    pages = "7514--7528"
}

@inproceedings{kirstain2023pickscore,
 author = {Kirstain, Yuval and Polyak, Adam and Singer, Uriel and Matiana, Shahbuland and Penna, Joe and Levy, Omer},
 booktitle = {Advances in Neural Information Processing Systems},
 editor = {A. Oh and T. Naumann and A. Globerson and K. Saenko and M. Hardt and S. Levine},
 pages = {36652--36663},
 publisher = {Curran Associates, Inc.},
 title = {{Pick-a-Pic}: An Open Dataset of User Preferences for Text-to-Image Generation},
 volume = {36},
 year = {2023}
}

@article{wu2023hpsv2,
  title={Human preference score v2: A solid benchmark for evaluating human preferences of text-to-image synthesis},
  author={Wu, Xiaoshi and Hao, Yiming and Sun, Keqiang and Chen, Yixiong and Zhu, Feng and Zhao, Rui and Li, Hongsheng},
  journal={arXiv preprint arXiv:2306.09341},
  year={2023}
}

@inproceedings{wang2024yolov9,
author = {Wang, Chien-Yao and Yeh, I-Hau and Mark Liao, Hong-Yuan},
title = {{YOLOv9}: Learning What You Want to Learn Using Programmable Gradient Information},
year = {2024},
isbn = {978-3-031-72750-4},
publisher = {Springer-Verlag},
address = {Berlin, Heidelberg},
doi = {10.1007/978-3-031-72751-1_1},
booktitle = {Computer Vision – ECCV 2024: 18th European Conference, Milan, Italy, September 29–October 4, 2024, Proceedings,  Part XXXI},
pages = {1--21},
numpages = {21},
location = {Milan, Italy}
}

@inproceedings{mindererowlv2,
 author = {Minderer, Matthias and Gritsenko, Alexey and Houlsby, Neil},
 booktitle = {Advances in Neural Information Processing Systems},
 editor = {A. Oh and T. Naumann and A. Globerson and K. Saenko and M. Hardt and S. Levine},
 pages = {72983--73007},
 publisher = {Curran Associates, Inc.},
 title = {Scaling Open-Vocabulary Object Detection},
 volume = {36},
 year = {2023}
}

@article{
lian2024lmd,
title={{LLM}-grounded Diffusion: Enhancing Prompt Understanding of Text-to-Image Diffusion Models with Large Language Models},
author={Long Lian and Boyi Li and Adam Yala and Trevor Darrell},
journal={Transactions on Machine Learning Research},
issn={2835-8856},
year={2024},
note={Featured Certification}
}

@INPROCEEDINGS{patel2025enhancing,
  author={Patel, Zakaria and Serkh, Kirill},
  booktitle={2025 IEEE/CVF Winter Conference on Applications of Computer Vision (WACV)}, 
  title={Enhancing Image Layout Control with Loss-Guided Diffusion Models}, 
  year={2025},
  volume={},
  number={},
  pages={3916-3924},
  doi={10.1109/WACV61041.2025.00385}}

@inproceedings{dinh2023rethink,
 author = {Dinh, Anh-Dung and Liu, Daochang and Xu, Chang},
 booktitle = {Advances in Neural Information Processing Systems},
 editor = {A. Oh and T. Naumann and A. Globerson and K. Saenko and M. Hardt and S. Levine},
 pages = {42285--42297},
 publisher = {Curran Associates, Inc.},
 title = {Rethinking Conditional Diffusion Sampling with Progressive Guidance},
 volume = {36},
 year = {2023}
}

@article{
zhang2025improving,
title={Improving {CLIP} Counting Accuracy via Parameter-Efficient Fine-Tuning},
author={Ruisu Zhang and Yicong Chen and Kangwook Lee},
journal={Transactions on Machine Learning Research},
issn={2835-8856},
year={2025}
}

@inproceedings{
clark2024directly,
title={Directly Fine-Tuning Diffusion Models on Differentiable Rewards},
author={Kevin Clark and Paul Vicol and Kevin Swersky and David J. Fleet},
booktitle={The Twelfth International Conference on Learning Representations},
year={2024}
}

@InProceedings{chen2025enhancing,
author="Chen, Chaofeng
and Wang, Annan
and Wu, Haoning
and Liao, Liang
and Sun, Wenxiu
and Yan, Qiong
and Lin, Weisi",
editor="Leonardis, Ale{\v{s}}
and Ricci, Elisa
and Roth, Stefan
and Russakovsky, Olga
and Sattler, Torsten
and Varol, G{\"u}l",
title="Enhancing Diffusion Models with Text-Encoder Reinforcement Learning",
booktitle="Computer Vision -- ECCV 2024",
year="2025",
publisher="Springer Nature Switzerland",
address="Cham",
pages="182--198",
isbn="978-3-031-72698-9"
}

@InProceedings{yang2024d3po,
    author    = {Yang, Kai and Tao, Jian and Lyu, Jiafei and Ge, Chunjiang and Chen, Jiaxin and Shen, Weihan and Zhu, Xiaolong and Li, Xiu},
    title     = {Using Human Feedback to Fine-tune Diffusion Models without Any Reward Model},
    booktitle = {Proceedings of the IEEE/CVF Conference on Computer Vision and Pattern Recognition (CVPR)},
    month     = {June},
    year      = {2024},
    pages     = {8941-8951}
}

@inproceedings{
black2024ddpo,
title={Training Diffusion Models with Reinforcement Learning},
author={Kevin Black and Michael Janner and Yilun Du and Ilya Kostrikov and Sergey Levine},
booktitle={The Twelfth International Conference on Learning Representations},
year={2024}
}

@InProceedings{wallace2024diffdpo,
    author    = {Wallace, Bram and Dang, Meihua and Rafailov, Rafael and Zhou, Linqi and Lou, Aaron and Purushwalkam, Senthil and Ermon, Stefano and Xiong, Caiming and Joty, Shafiq and Naik, Nikhil},
    title     = {Diffusion Model Alignment Using Direct Preference Optimization},
    booktitle = {Proceedings of the IEEE/CVF Conference on Computer Vision and Pattern Recognition (CVPR)},
    month     = {June},
    year      = {2024},
    pages     = {8228-8238}
}

@inproceedings{xu2023imagereward,
 author = {Xu, Jiazheng and Liu, Xiao and Wu, Yuchen and Tong, Yuxuan and Li, Qinkai and Ding, Ming and Tang, Jie and Dong, Yuxiao},
 booktitle = {Advances in Neural Information Processing Systems},
 editor = {A. Oh and T. Naumann and A. Globerson and K. Saenko and M. Hardt and S. Levine},
 pages = {15903--15935},
 publisher = {Curran Associates, Inc.},
 title = {{ImageReward}: Learning and Evaluating Human Preferences for Text-to-Image Generation},
 volume = {36},
 year = {2023}
}

@InProceedings{wallace2023doodl,
    author    = {Wallace, Bram and Gokul, Akash and Ermon, Stefano and Naik, Nikhil},
    title     = {End-to-End Diffusion Latent Optimization Improves Classifier Guidance},
    booktitle = {Proceedings of the IEEE/CVF International Conference on Computer Vision (ICCV)},
    month     = {October},
    year      = {2023},
    pages     = {7280-7290}
}

@inproceedings{fan2023dpok,
 author = {Fan, Ying and Watkins, Olivia and Du, Yuqing and Liu, Hao and Ryu, Moonkyung and Boutilier, Craig and Abbeel, Pieter and Ghavamzadeh, Mohammad and Lee, Kangwook and Lee, Kimin},
 booktitle = {Advances in Neural Information Processing Systems},
 editor = {A. Oh and T. Naumann and A. Globerson and K. Saenko and M. Hardt and S. Levine},
 pages = {79858--79885},
 publisher = {Curran Associates, Inc.},
 title = {{DPOK}: Reinforcement Learning for Fine-tuning Text-to-Image Diffusion Models},
 volume = {36},
 year = {2023}
}

@inproceedings{eyring2024reno,
 author = {Eyring, Luca and Karthik, Shyamgopal and Roth, Karsten and Dosovitskiy, Alexey and Akata, Zeynep},
 booktitle = {Advances in Neural Information Processing Systems},
 editor = {A. Globerson and L. Mackey and D. Belgrave and A. Fan and U. Paquet and J. Tomczak and C. Zhang},
 pages = {125487--125519},
 publisher = {Curran Associates, Inc.},
 title = {{ReNO}: Enhancing One-step Text-to-Image Models through Reward-based Noise Optimization},
 volume = {37},
 year = {2024}
}

@article{chung2024cfg++,
  publtype={informal},
  author={Hyungjin Chung and Jeongsol Kim and Geon Yeong Park and Hyelin Nam and Jong Chul Ye},
  title={{CFG}++: Manifold-constrained Classifier Free Guidance for Diffusion Models},
  year={2024},
  cdate={1704067200000},
  journal={CoRR},
  volume={abs/2406.08070}
}

@article{chefer2023attend,
author = {Chefer, Hila and Alaluf, Yuval and Vinker, Yael and Wolf, Lior and Cohen-Or, Daniel},
title = {Attend-and-Excite: Attention-Based Semantic Guidance for Text-to-Image Diffusion Models},
year = {2023},
issue_date = {August 2023},
publisher = {Association for Computing Machinery},
address = {New York, NY, USA},
volume = {42},
number = {4},
issn = {0730-0301},
doi = {10.1145/3592116},
journal = {ACM Trans. Graph.},
month = jul,
articleno = {148},
numpages = {10}
}

@InProceedings{binyamin2025mic,
    author    = {Binyamin, Lital and Tewel, Yoad and Segev, Hilit and Hirsch, Eran and Rassin, Royi and Chechik, Gal},
    title     = {Make It Count: Text-to-Image Generation with an Accurate Number of Objects},
    booktitle = {Proceedings of the IEEE/CVF Conference on Computer Vision and Pattern Recognition (CVPR)},
    month     = {June},
    year      = {2025},
    pages     = {13242-13251}
}

@INPROCEEDINGS{kang2025cg,
  author={Kang, Wonjun and Galim, Kevin and Il Koo, Hyung and Cho, Nam Ik},
  booktitle={2025 IEEE/CVF Winter Conference on Applications of Computer Vision (WACV)}, 
  title={Counting Guidance for High Fidelity Text-to-Image Synthesis}, 
  year={2025},
  volume={},
  number={},
  pages={899-908},
  doi={10.1109/WACV61041.2025.00097}}

@InProceedings{zafar2026tokenopt,
    author    = {Zafar, Oz and Cohen, Yuval and Wolf, Lior and Schwartz, Idan},
    title     = {Detection-Driven Object Count Optimization for Text-to-Image Diffusion Models},
    booktitle = {Proceedings of the IEEE/CVF Winter Conference on Applications of Computer Vision (WACV)},
    month     = {March},
    year      = {2026},
    pages     = {1885-1894}
}
}

\newpage
\appendix
\onecolumn

\section*{Appendix}

Appendix~\ref{app:why_detector} gives a high-level summary comparing detection vs. regression-based numeracy critics in the generative setting. 
Appendix~\ref{app:pilot} validates our method via human pilot study.
Appendix~\ref{app:vs_bestofn} compares \emph{D2D} to a roughly compute-matched, best-of-N baseline.
Appendix~\ref{app:overhead} gives additional details and discussion on the numeracy-speed tradeoff.
Appendix~\ref{app:algo} details the algorithms for pre-inference alignment of the Latent Modifier Network (LMN) and inference-time, numeracy optimization.
Appendix~\ref{app:hyper} describes additional implementation details, including hyperparameters used for the LMN architecture (Appendix~\ref{app:hyp_lmn}), core \emph{D2D} formulation (Appendix~\ref{app:hyp_core}), and optimization (Appendix~\ref{app:hyp_opt}). Appendix~\ref{app:multi_object} discusses the implementation details of extending \emph{D2D} to multi-object scenarios. 
Appendix~\ref{app:sana_flux} applies \emph{D2D} to FLUX-schnell~\cite{flux2024}, a larger base model, and SANA-Sprint~\cite{chen2025sana}, a recent one-step T2I model.
Appendix~\ref{app:lmn_breakdown} shows quantitatively how the LMN improves numeracy. 
Appendix~\ref{app:d2d_quality} shows \emph{D2D} has image quality and overall prompt alignment comparable to counting baselines, even surpassing multi-step baselines in many cases (Appendix~\ref{app:extended_qual}). Also included are per-object fidelity results (Appendix~\ref{app:per_object}).
Appendix~\ref{app:d2d_complement} shows the performance improvements that result from using \emph{D2D} to complement existing baselines.
Appendix~\ref{app:multi_lowhigh} shows a breakdown of numeracy by total density on D2D-Multi. 
Appendix~\ref{app:high_density_analysis} includes further deep-dive analysis on the impact of object size, detector confidence, and requested count on numeracy.  
Appendix~\ref{app:overlap_analysis} shows how \emph{D2D} handles overlapping detections.
Appendix~\ref{app:yolocount} shows \emph{D2D}'s numeracy improvements over the more recent YOLO-COUNT~\cite{zeng2025yolocount} work.
Appendix~\ref{app:enmmdit} compares \emph{D2D} with method EnMMDiT~\cite{wei2026enhancing}.
Appendix~\ref{app:countgd} shows CountGD's baseline counting performance, relative to other regression/detector-based counters.
Additional qualitative results are illustrated in Appendix~\ref{app:add_qual}, including on more free-form prompts (Appendix~\ref{app:freeform}).
Appendix~\ref{app:add_abl} includes results from additional ablation experiments, including studies on the value of the mixing weight $w$ used to mix the LMN output with the original noise (Appendix~\ref{app:w_abl}); variants of the LMN architecture (Appendix~\ref{app:lmn_arc}); the necessity of pre-inference alignment (Appendix~\ref{app:preinf_align}) and inference-time calibration (Appendix~\ref{app:lmd_infalign}) of $M_\phi$; and regularization term variants (Appendix~\ref{app:reg}).
Appendix~\ref{app:d2d_scalingform} compares alternative formulations of the \emph{D2D} loss.
Appendix~\ref{app:opt_traj} shows a sample numeracy optimization trajectory for a single prompt, comparing the raw values of the \emph{D2D} loss $\mathcal{L}_{\text{D2D}}$ and $f$ over iterations.
Appendix~\ref{app:selective_d2d} details methods to preprocess prompts for selective activation of the \emph{D2D} method.
Appendix~\ref{app:multistepdm} demonstrates an extension of the \emph{D2D} critic for multi-step diffusion models, using an existing multi-step-oriented initial noise optimization framework.
Appendix~\ref{app:div_gen} presents analysis on the impact of \emph{D2D} on the diversity of generated images.
Appendix~\ref{app:broader} describes broader impacts.
Appendix~\ref{app:llm} describes our LLM usage.

\section{Detectors vs. regression-based numeracy critics}\label{app:why_detector}
Based on our analysis of the baseline counting accuracies of detectors compared to regression-based models (Fig.~\ref{fig:det_better}), we observe that detectors are more robust, as also observed in previous works~\cite{zhang2025improving, liu2018decide}. Furthermore, this performance boost generalizes to the generative setting, because accurate count understanding is one of the key components of numeracy optimization. As for the question of why detectors are more robust in the first place, we refer to existing work~\cite{liu2018decide}, which indicates this may be due to detectors' more advanced localization ability. In the low-density setting, objects are more likely to be sparsely separated, so it becomes essential to localize each instance in order to obtain a count; detectors count explicitly by localizing instances first, whereas regression-based methods capture the density of the crowd of instances, generally without specific localization objectives. In the high-density setting, when object instances are too small and potentially occluded, localization may become more difficult, thereby reducing the accuracy of detector counts.

\newpage
\section{Human pilot study}\label{app:pilot}

To validate \emph{D2D} performance, we conducted additional human evaluation and multi-evaluator agreement studies, which confirm \emph{D2D}’s performance boosts. During model development, we employed different detectors for \emph{D2D}’s critic (OWLv2) and evaluator (CountGD) to avoid potential critic-evaluator leak. Human evaluation and automated YOLOv9/CountGD-based metrics agree that \emph{D2D} outperforms counting baselines (Tab.~\ref{tab:valid_study}).

\begin{table}[H]
  \caption{\textbf{\emph{D2D} outperforms counting baselines,} as validated by human evaluators and YOLOv9/CountGD. On 50 randomly selected prompts from D2D-Small.}
  \label{tab:valid_study}
  \centering
  \begin{tabular}{l c c c}
    \toprule
     \multicolumn{1}{c}{Method} & Human & YOLOv9~\cite{wang2024yolov9} & CountGD~\cite{amini2024countgd} \\ \midrule
    SDXL~\cite{podell2024sdxl} & 0.10 & 0.10& 0.10 \\
    SDXL-Turbo~\cite{sauer2025turbo} & 0.24& 0.20 & 0.22 \\
    TokenOpt~\cite{zafar2026tokenopt} (SDXL-Turbo) & 0.26 & 0.24 & 0.22 \\
    Counting Guidance~\cite{kang2025cg} (SD1.4) & 0.12 & 0.16 & 0.24 \\
    Make It Count~\cite{binyamin2025mic} (SDXL) & 0.32 & 0.38 & 0.34 \\
    \emph{D2D} w/ OWLv2 (SDXL-Turbo)& \textbf{0.40} & \textbf{0.40} & \textbf{0.42} \\
    \bottomrule
  \end{tabular}
\end{table}

We conducted the pilot study with 22 volunteers from the author's network using 50 randomly selected prompts from D2D-Small across the six models in the table above (300 total images), with each image evaluated by at least two different participants and the rounded average of their responses used as the final score. Each volunteer was given a questionnaire of 30 randomly shuffled questions, each with an image and a short, free-response ``How many [object class] are there in the image?'' Importantly, we did not disclose which baseline generated each image. The collected data consists of images and corresponding counts; the images carry minimal risk. Below is the prompt we used:

\textit{Thank you so much for agreeing to participate in this short questionnaire. I really appreciate you! Below are 30 questions. Each question presents an image and a short prompt: ``How many <object> are there in the image?'' Based on the image, answer the question. Below is an example, to illustrate. How many backpacks are there in the image? [image] Example user response: 6. Continue to the next section to begin answering the questions. All answers must be integers, in numerical format. So 15 is a valid answer, but fifteen is not.
0 is also a possible answer.
Emphasis: a single integer response for each question - no additional text.
When the answer seems unclear, please make your best effort. Thank you!}

Our method yields the lowest error between generated and requested counts, as computed from human scores (Tab.~\ref{tab:valid_study_error}).

\begin{table}[H]
  \caption{\textbf{\emph{D2D} yields the lowest mean absolute error.} On 50 randomly selected prompts from D2D-Small.}
  \label{tab:valid_study_error}
  \centering
  \begin{tabular}{l c}
    \toprule
     \multicolumn{1}{c}{Method} & (Human) mean $|$generated - requested count$|$ \\ \midrule
    SDXL~\cite{podell2024sdxl} & 6.00 \\
    SDXL-Turbo~\cite{sauer2025turbo} & 3.42 \\
    TokenOpt~\cite{zafar2026tokenopt} (SDXL-Turbo)&3.22\\
    Counting Guidance~\cite{kang2025cg} (SD1.4)&3.34 \\
    Make It Count~\cite{binyamin2025mic} (SDXL) & 3.88 \\
    \emph{D2D} w/ OWLv2 (SDXL-Turbo) & \textbf{1.76}  \\
    \bottomrule
  \end{tabular}
\end{table}

Using Krippendorff's $\alpha$~\cite{krippendorff2019reliability} with the interval metric to measure inter-rater agreement (maximum value of 1), we compute a score of 0.924, indicating strong agreement.

\newpage
\section{\emph{D2D} vs. compute-matched best-of-N baseline}\label{app:vs_bestofn}
We additionally run experiments on an A100 (80 GB), comparing \emph{D2D} against a roughly compute-matched, best-of-N baseline on CoCoCount and D2D-Small, using the OWLv2 detector to select the best-of-N image. Tab.~\ref{tab:best_of_n_coco} shows \emph{D2D} outperforms best-of-N, with N=45 for CoCoCount and N=70 for D2D-Small.

\begin{table}[H]
  \caption{\textbf{\emph{D2D} beats best-of-N baseline.} Numeracy followed by inference cost in seconds and compute-matched value of N for the baseline in parentheses. On SDXL-Turbo, seed=0, using A100 GPU.}
  \label{tab:best_of_n_coco}
  \centering
  \begin{tabular}{l c c}
    \toprule
     \multicolumn{1}{c}{Method} & CoCoCount~\cite{binyamin2025mic} & D2D-Small \\ \midrule
     Best-of-N & 54.0 (22 sec, N=45) & 33.5 (35 sec, N=70) \\
    \emph{D2D} w/ OWLv2 & \textbf{59.0} (22 sec) & \textbf{41.0} (34 sec) \\
    \bottomrule
  \end{tabular}
\end{table}

\section{\emph{D2D} introduces minimal inference time overhead}\label{app:overhead}
Improving generation capabilities beyond the base model introduces tradeoffs, as seen in other T2I settings, such as quality–speed and quality/alignment-diversity~\cite{xiaofeng2026intricate}. \emph{D2D} achieves this improvement with minimal overhead compared to existing count-correction methods, as shown in Tab.~\ref{tab:overhead} below.

\begin{table}[h]
\centering
\small
\caption{\textbf{\emph{D2D} introduces substantially less overhead than existing count-correction methods.} Base model inference time and additional overhead in seconds. On D2D-Small. Avg. over four seeds.}
\label{tab:overhead}
\begin{tabular}{lcc}
\toprule
 \multicolumn{1}{c}{Method} & \multicolumn{1}{c}{Base inference time (s)} & \multicolumn{1}{c}{Additional overhead (s)} \\
\midrule
TokenOpt~\cite{zafar2026tokenopt}          & 0.16 (SDXL-Turbo) & 102.94 \\
Make It Count~\cite{binyamin2025mic} & 7.68 (SDXL)       & 34.60 \\
Counting Guidance~\cite{kang2025cg} & 1.52 (SD1.4)      & 26.87 \\
\emph{D2D}               & 0.16 (SDXL-Turbo) & \textbf{21.55} \\
\bottomrule
\end{tabular}
\end{table}

Moreover, for applications where count correctness matters, this additional cost can be a worthwhile tradeoff: unlike subjective image quality metrics, numeracy is often a correctness criterion. Finally, \emph{D2D} is backbone-flexible and does not require retraining, allowing it to directly benefit from future improvements in diffusion inference efficiency and faster generation pipelines.

\newpage
\section{Algorithms}\label{app:algo}

\begin{algorithm}[h]
 \caption{Inference}\label{alg:inference}
\begin{algorithmic}
    \small
    
    \STATE {\bfseries Input:} Prompt $p$ specifying $N$ of object of class $C$; pre-trained Latent Modifier Network $M_\phi$; latent dimension $d$; weight $w$, diffusion model $G_\theta$; minimum \# calibration iterations $t_{\text{min}}$; threshold value specifying ``good enough'' regularization $\tau_{\text{reg}}$; counter $f_{\beta,\tau_z}$ and critic $\mathcal{L}_{\text{D2D}}$; Stage 1 (Calibration) learning rate $\eta_{\text{calib}}$ and loss weight $\lambda_{\text{calib}}$; Stage 2 numeracy optimization learning rate $\eta$ and loss weights $\alpha$ and $\lambda$; \# tuning steps $K$.
    \STATE {\bfseries Output:} Optimal noise $\mathbf{x}_T^*$.

    \highlightcyan{
    \STATE \textbf{Stage 1:} Calibrate $M_\phi$ to newly sampled $\mathbf{x}_T$.
    \STATE resample $\gets $\texttt{True}
    \WHILE{resample}
        \STATE Sample $\mathbf{x}_T \in \mathbb{R}^d \sim \mathcal{N}(0, \bm{I}) $
        \FOR{$1 \leq t \leq K$}
            \STATE $\mathbf{x}_T' = w \cdot \mathbf{x}_T + (1-w) \cdot M_\phi(\mathbf{x}_T)$
            \STATE Compute $\mathcal{L} = \lambda_{\text{calib}} \mathcal{L}_{\text{reg}}'$

            \IF{$ t \geq t_{\text{min}} $ and $\mathcal{L} \leq \tau_{\text{reg}} $ }  
                \STATE resample $ \gets $ \texttt{False}
                \STATE \textbf{break}
            \ELSE 
                \STATE $\phi \Leftarrow \phi - \eta_{\text{calib}}\nabla \mathcal{L}$
            \ENDIF
        \ENDFOR
    \ENDWHILE
    }

    \highlightgray{
        \STATE \textbf{Stage 2:} Optimize numeracy.
        \FOR{$t \leq \text{epoch} \leq K$} 
            \STATE Compute $\mathcal{L}_{\text{reg}}$
            \STATE $I = G_\theta(\mathbf{x}_T', p)$
            \STATE Compute $f_{\beta,\tau_z}$ and $\mathcal{L}_{\text{D2D}}$
            \STATE  {\bfseries return} if $f_{\beta,\tau_z} = N$ \COMMENT{if $I$ is optimal, stop early}
            \STATE $\phi \Leftarrow \phi - \eta \nabla(\alpha \mathcal{L}_{\text{D2D}} + 
    \lambda \mathcal{L}_{\text{reg}})$
             \STATE $ \mathbf{x}_T' = w \cdot \mathbf{x}_T + (1-w) \cdot M_\phi(\mathbf{x}_T)$
        \ENDFOR
    }
\end{algorithmic}
\end{algorithm}

\begin{algorithm}[H]
\caption{Pre-inference alignment stage (done once per model)}\label{alg:mphi_init}
\begin{algorithmic}
    \STATE {\bfseries Input:} Latent Modifier Network $M_\phi$, initialized with random weights; latent dimension $d$; weight $w$; learning rate $\eta$, loss weight $\lambda$.
    \STATE {\bfseries Output:} Pre-trained $M_\phi$ with output aligned to Gaussian distribution.

    \STATE Set seed = 1.
    \REPEATN{100} 
        \STATE Sample $\mathbf{x}_T \in \mathbb{R}^d \sim \mathcal{N}(0, \bm{I}) $
        \FOR{$1 \leq \text{epoch} \leq 200$}
            \STATE $\mathbf{x}_T^\prime = w \cdot \mathbf{x}_T + (1-w) \cdot M_\phi(\mathbf{x}_T)$
            \STATE Compute $\mathcal{L} = \lambda$$\mathcal{L}'_{\text{reg}}$
            
            \STATE $\phi \Leftarrow \phi - \eta\nabla \mathcal{L} $
        \ENDFOR
    \ENDREPEAT
\end{algorithmic}
\end{algorithm}

\newpage
\section{Additional implementation details}\label{app:hyper}

\subsection{Architecture of the Latent Modifier Network}\label{app:hyp_lmn}
The LMN is a three-layer perceptron, with input/output layers of size $d$ (dimension of initial latent, $\mathbf{x}_T \in \mathbb{R}^d$), and two hidden layers of size 100 each. On base models SDXL-Turbo, SD-Turbo, and Pixart-DMD, the LMN has 3,303,384 tunable parameters.

\subsection{\emph{D2D} formulation}\label{app:hyp_core}
The core \emph{D2D} hyperparameters are the detector threshold $\tau$ and steepness coefficient $\beta$, which determine the transition threshold and curvature of the sigmoid underlying $\mathcal{L}_{\text{D2D}}$, and the mixing weight $w$, which determines the ratio with which to combine the LMN output and original noise to obtain the optimal noise. We use $\tau=0.2$, $\beta=300$, and $w = 0.2$. Studies on hyperparameters $\tau$ and $\beta$ are reported in Tab.~\ref{tab:abl_hyperparam} in Sec.~\ref{sec:main_analysis} of the main text. Experiments on $w$ are reported in Tab.~\ref{tab:abl_w} in Appendix~\ref{app:w_abl}. Unlike the OWLv2 critic, for YOLOv9 we apply NMS filtering (threshold=0.7) prior to count summation.

\subsection{Optimization}\label{app:hyp_opt}
During pre-inference alignment and inference-time calibration, when we optimize only the regularization term, we use $\mathcal{L}_{\text{reg}}'$. At inference-time, during numeracy optimization, we use $\mathcal{L}_{\text{reg}}$ in conjunction with $\mathcal{L}_{\text{D2D}}$. Appendix~\ref{app:reg} reports ablation studies on using $\mathcal{L}_{\text{reg}}'$ vs. $\mathcal{L}_{\text{reg}}$ during numeracy optimization.

In the pre-inference alignment stage, we use learning rate $\eta=10^{-4}$ and loss weight $\lambda=0.01$. At inference-time (stage 1: calibration), we use minimum number of iterations $t_{\text{min}} = 70$; ``good enough'' threshold for determining when calibration is done $\tau_{\text{reg}}=-712.8$, learning rate $\eta_{\text{calib}} =10^{-3}$, and loss weight $\lambda_{\text{calib}}=0.01$. During stage 2 (numeracy optimization), we use learning rate $\eta=5\times10^{-4}$ and loss weights $\alpha=5$ and $\lambda=10^{-4}$. During this stage, we use count-based, adaptive learning rate scheduling of $\eta$ to ease steady convergence when the generated count approximates the requested count; we also adaptively rescale (i.e., increase) $\lambda$ to counteract larger deviations from Gaussian (i.e., divergence).

For \emph{D2D} experimental results with Pixart-DMD on benchmark D2D-Large, we use the base detector OWLv2, instead of $f_{\beta,\tau_z}$, to perform early-stopping. For single-object scenarios, we set $K=200$, except for \emph{D2D} with Pixart-DMD on D2D-Large, for which we use $K=400$. For the multi-object scenario, we use $K=400$.


\section{Multi-object \boldmath$\mathcal{L}_{\text{D2D}}$}\label{sec:multi_critic}\label{app:multi_object}
In extending \emph{D2D} to prompts with $m > 1$ object classes $\{C_{j}, 1\leq j \leq m\}$, we have to note that every predicted bbox $B_i$ now comes with $m$ corresponding scores $z_i^{(1)},z_i^{(2)},...,z_i^{(m)}$, the max of which, $z_i^{\text{max}}$, indicates its corresponding label. Only the bboxes where $z_i^{\text{max}}>\tau_z$ are counted. To correct over/under-generation, our approach then is to focus on appropriately adjusting the largest score $z_i^{\text{max}}$ of each bbox, while minimizing all other (i.e., non-max) scores. This means minimizing the $z_i^{\text{max}}$ of bboxes that correspond to over-generated classes and maximizing the $z_i^{\text{max}}$ of those that correspond to under/correctly generated classes (to increase/maintain the count). We use roughly the same formulation as Eq.~\ref{eq:diffcritic}, with $z_i$ replaced by the specific per-class, per-bbox logits $z_i^j$ that need to be minimized or maximized (Eq.~\ref{eq:multi_critic}).

If $\mathbb{C}_{\geq}$ is the set of classes that are under-generated or correctly generated, then \[\mathbb{S}_{\text{bbox}}=\{(i, j) | 1 \leq i \leq n \land j = (\arg\max_{j'} z_i^{j'}) \in \mathbb{C}_{\geq} \}\] refers to the set of the bboxes corresponding to the under/correctly generated classes, along with their max logits. We compute the loss $\mathcal{L}_{\text{D2D}}$ like so:
\begin{equation}\label{eq:multi_critic}
    \mathcal{L}_{\text{D2D}}^{\text{multi}} = \sum_{(i,j)\in \mathbb{S}_{\text{bbox}}} \sigma(\beta \cdot (\tau_{z} - z_i^j)) \cdot (\tau_z - z_i^j) + \sum_{(i,j)\notin \mathbb{S}_{\text{bbox}}} \sigma(\beta \cdot (z_i^j - \tau_{z})) \cdot (z_i^j - \tau_z)
\end{equation}

\newpage
\section{\emph{D2D} extensions on FLUX-schnell~\cite{flux2024} and SANA-Sprint~\cite{chen2025sana}}\label{app:sana_flux}
\vspace{-0.1in}

\begin{table}[H]
  \caption{\textbf{\emph{D2D} also yields numeracy gains on FLUX-schnell, a larger base model.} On CoCoCount~\cite{binyamin2025mic}, seed=0.}
  \label{tab:flux}
  \centering
  
        \begin{tabular}{lcc}
        \toprule
        \multicolumn{1}{c}{Method} & Numeracy \\ \midrule
        FLUX-schnell~\cite{flux2024} (1-step) &  46.0  \\ 
        FLUX-schnell~\cite{flux2024} (4-step) & 47.5  \\ \midrule
        FLUX-schnell~\cite{flux2024} (1-step) + \emph{D2D} & \textbf{57.0} \\
        \bottomrule
      \end{tabular}
\end{table}

\vspace{-0.1in}

\begin{table}[H]
  \caption{\textbf{\emph{D2D} also yields numeracy gains on SANA-Sprint, a recent advancement in one-step T2I.} Evaluated on 50 randomly sampled prompts from CoCoCount~\cite{binyamin2025mic}, seed=0.}
  \label{tab:sana}
  \centering
        \begin{tabular}{lcc}
        \toprule
        \multicolumn{1}{c}{Method} & Numeracy \\ \midrule
        SANA-Sprint~\cite{chen2025sana} & 42.0 \\ \midrule
        SANA-Sprint~\cite{chen2025sana} + \emph{D2D} & \textbf{62.0} \\
        \bottomrule
      \end{tabular}
\end{table}

\begin{figure}[H]
     \centering
     \includegraphics[width=0.6\textwidth]{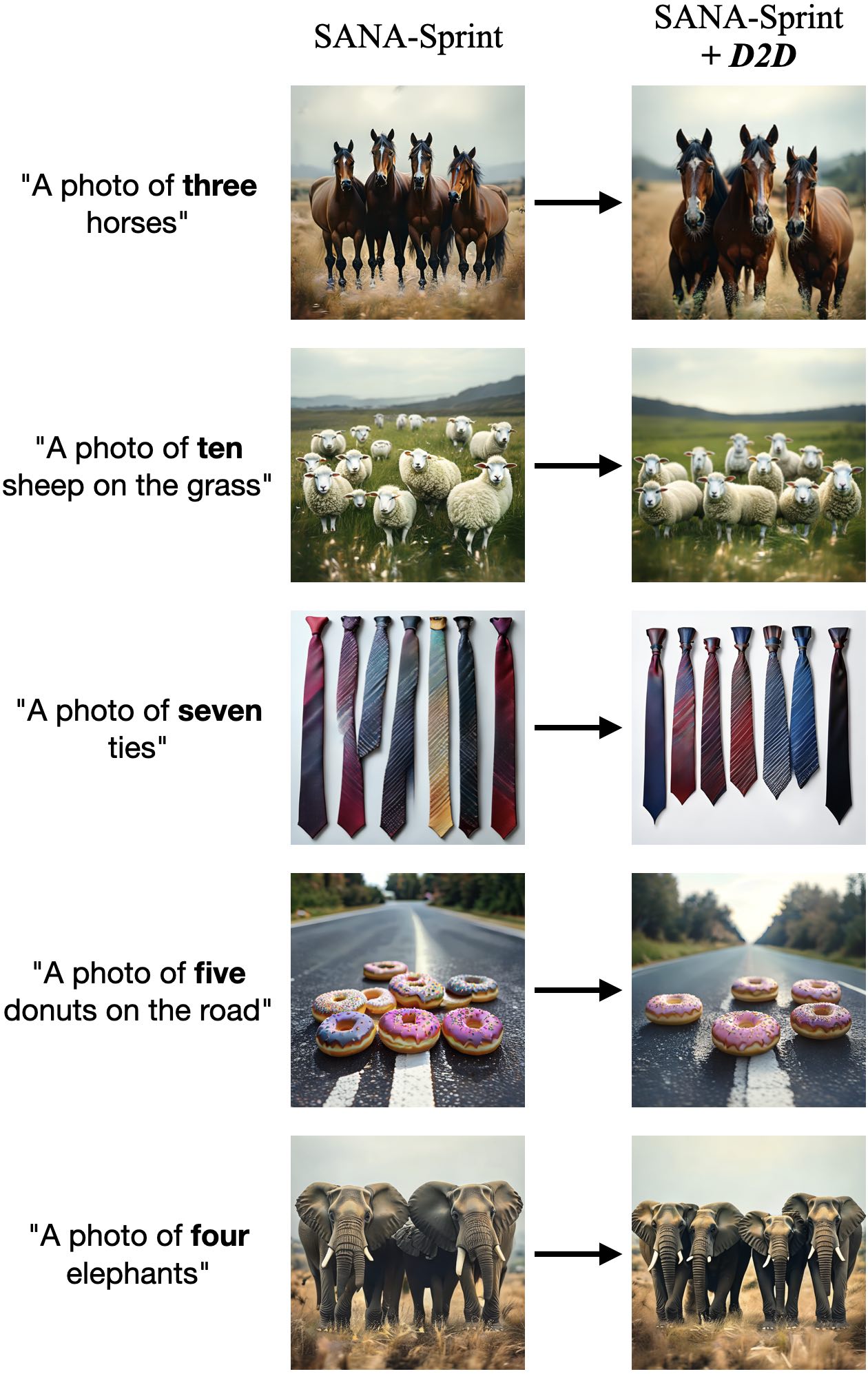}
     \caption{Qualitative results of \emph{D2D} on SANA-Sprint~\cite{chen2025sana}.}
     \label{fig:sana}
\end{figure}

\newpage
\section{The Latent Modifier Network effectively corrects numeracy}\label{app:lmn_breakdown}

\setlength{\aboverulesep}{2pt}
\setlength{\belowrulesep}{0pt}
\setlength{\extrarowheight}{.6ex}
\setlength{\tabcolsep}{2pt}
\begin{table*}[h]
\captionsetup{skip=2pt}
  \caption{\textbf{The LMN boosts numeracy.} We compare  \emph{D2D} against ReNO~\cite{eyring2024reno} using $\mathcal{L}_{\text{D2D}}$ and $\mathcal{L}_{\text{reg}}'$ for both, controlling for the number of iterations tuned. We note boosts in numeracy, with comparable image quality. On CoCoCount~\cite{binyamin2025mic} and D2D-Small. Base model SDXL-Turbo. Avg. over four seeds.}
  \label{tab:mphi}
  \centering
   \resizebox{\textwidth}{!}{
  \begin{tabular}{p{10em}cc@{}p{0.5em}@{}cc@{}p{0.5em}@{}cc@{}p{0.5em}@{}cc@{}p{0.5em}@{}cc@{}p{0.5em}@{}}
    \toprule
    \multicolumn{1}{c}{\multirow{2}{*}{Method}} & \multicolumn{2}{c}{CountGD~\cite{amini2024countgd} $\uparrow$} && \multicolumn{2}{c}{ImageReward~\cite{xu2023imagereward} $\uparrow$} && \multicolumn{2}{c}{PickScore~\cite{kirstain2023pickscore} $\uparrow$} && \multicolumn{2}{c}{HPSv2~\cite{wu2023hpsv2} $\uparrow$} && \multicolumn{2}{c}{CLIPScore~\cite{hessel2021clipscore} $\uparrow$} \\ 
    \cmidrule{2-3} \cmidrule{5-6} \cmidrule{8-9} \cmidrule{11-12} \cmidrule{14-15}
    & CoCoCount & D2D-Small && CoCoCount & D2D-Small && CoCoCount & D2D-Small && CoCoCount & D2D-Small && CoCoCount & D2D-Small \\ \midrule
    ReNO~\cite{eyring2024reno} w/ $\mathcal{L}_{\text{D2D}}$, $\mathcal{L}_{\text{reg}}'$ & 43.25 & 32.00 && 1.04 & 0.45 && 23.25 & 21.98 && 0.296 & 0.281 && 32.81 & 31.79 \\
    \emph{D2D} w/ $\mathcal{L}_{\text{D2D}}$, $\mathcal{L}_{\text{reg}}'$ & \textbf{53.88} & \textbf{42.44} && 1.08 & 0.52 && 23.28 & 21.99 && 0.299 & 0.282 && 32.77 & 31.71 \\
    \bottomrule
  \end{tabular}
  }
  \vspace{-0.1in}
\end{table*}
\setlength{\aboverulesep}{\oldaboverulesep}
\setlength{\belowrulesep}{\oldbelowrulesep}
\setlength{\extrarowheight}{\oldextrarowheight}
\setlength{\tabcolsep}{\oldtabcolsep}

\section{\emph{D2D} on image quality and overall prompt alignment}\label{app:d2d_quality}

\begin{figure}[h]
         \centering
         \includegraphics[width=\textwidth]{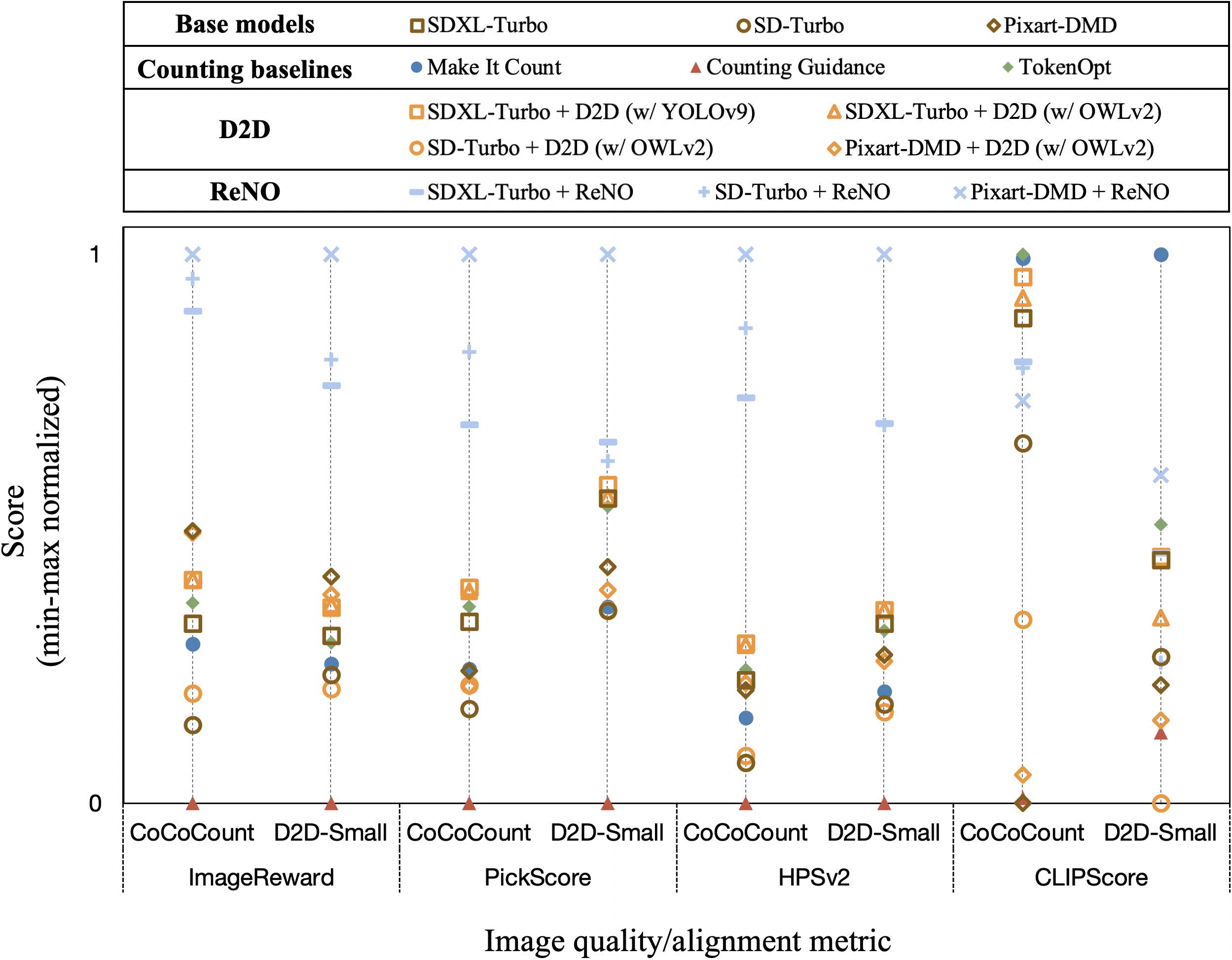}
         \caption{\textbf{Image quality/alignment scores} (ImageReward~\cite{xu2023imagereward}, PickScore~\cite{kirstain2023pickscore}, HPSv2~\cite{wu2023hpsv2}, CLIPScore~\cite{hessel2021clipscore}) by method. Aside from ReNO~\cite{eyring2024reno}, which often scores highest (it specifically optimizes those metrics), \emph{D2D} is comparable to counting baselines. Min-max normalized.}
         \label{fig:main_qual}
\end{figure}

\subsection{\emph{D2D} exhibits minimal degradation in image quality, compared to layout-control method Make It Count}\label{app:extended_qual}

Tab.~\ref{tab:qual_base} reports the image quality of layout control-based, count-correction method Make It Count~\cite{binyamin2025mic}, its base model SDXL, and \emph{D2D}, along with base model SDXL-Turbo. Make It Count exhibits some tradeoffs between numeracy and image quality, yielding slightly lower PickScore and HPSv2, compared to its base model SDXL (e.g., on D2D-Small: PickScore: \textbf{21.70} (SDXL), 21.48 (MIC); HPSv2: \textbf{0.272} (SDXL), 0.264 (MIC)). 

\emph{D2D} offers minimal image quality degradation (and sometimes better image quality) than base model SDXL-Turbo (e.g., on D2D-Small: ImageReward: 0.40 (SDXL-Turbo), \textbf{0.51} (\emph{D2D} w/ OWLv2); HPSv2: 0.279 (SDXL-Turbo), \textbf{0.282} (\emph{D2D} w/ OWLv2)).

\setlength{\aboverulesep}{2pt}
\setlength{\belowrulesep}{0pt}
\setlength{\extrarowheight}{.6ex}
\setlength{\tabcolsep}{2pt}
\begin{table}[h]
  \caption{\textbf{D2D yields minimal degradation in image quality and alignment, with minimal computational overhead, compared to layout control-based, multi-step method Make It Count.} Base models with no post-enhancement highlighted in gray. On CoCoCount~\cite{binyamin2025mic} and D2D-Small. Avg. over four seeds.
  }
  \label{tab:qual_base}
  \centering
  \resizebox{\textwidth}{!}{
  \begin{tabular}{p{16em} p{0.5em} @{} c c p{0.5em} @{} c c p{0.5em} @{} c c p{0.5em} @{} c c p{0.5em} @{} r}
    \toprule
    \addlinespace[3pt]
    \multicolumn{1}{c}{\multirow{2}{*}{Method}}  && \multicolumn{2}{c}{ImageReward $\uparrow$} && \multicolumn{2}{c}{PickScore $\uparrow$} && \multicolumn{2}{c}{HPSv2 $\uparrow$} && \multicolumn{2}{c}{CLIPScore $\uparrow$} && \multirow{2}{*}{\adjustbox{valign=c}{\parbox{3em}{\centering \small{Inference Time (s)}}}}   \\ 
    \cmidrule{3-4} \cmidrule{6-7} \cmidrule{9-10} \cmidrule{12-13}
    && \small{CoCoCount} & \small{D2D-Small} && \small{CoCoCount} & \small{D2D-Small} && \small{CoCoCount} & \small{D2D-Small} && \small{CoCoCount} & \small{D2D-Small} && \\ \midrule
    \rowcolor{gray!22} SDXL \small{\cite{podell2024sdxl}} && 0.87 & \textbf{0.31} && \textbf{22.99} & \textbf{21.70} && \textbf{0.290} & \textbf{0.272} && 32.91 & 32.58 && 7.69 \\
    \:\:\: + Make It Count \small{\cite{binyamin2025mic}} && \textbf{0.91} & 0.30 && 22.83 & 21.48 && 0.286 & 0.264 && \textbf{32.92} & \textbf{32.96} && 37.16 \\ \midrule
    \rowcolor{gray!22} SDXL-Turbo \small{\cite{sauer2025turbo}} && 0.96 & 0.40 && 23.12 & 21.98 && 0.293 & 0.279 && 32.80 & 31.89 && 0.16 \\
    \:\:\: + \emph{D2D} (w/ OWLv2) && \textbf{1.06} & \textbf{0.51} && 23.30 & 21.98 && \textbf{0.300} & \textbf{0.282} && 32.84 & 31.69 && 19.42 \\ 
    \:\:\: + \emph{D2D} (w/ YOLOv9) && \textbf{1.06} & 0.50 && \textbf{23.33} & \textbf{22.04} && \textbf{0.300} & \textbf{0.282} && \textbf{32.88} & \textbf{31.90} && 11.39 \\  
    \bottomrule
  \end{tabular}
 }
\end{table}
\setlength{\aboverulesep}{\oldaboverulesep}
\setlength{\belowrulesep}{\oldbelowrulesep}
\setlength{\extrarowheight}{\oldextrarowheight}
\setlength{\tabcolsep}{\oldtabcolsep}

\subsection{Per-object fidelity}\label{app:per_object}

We compute the PickScore (quality check) and CLIPScore (alignment score) per detected object instance to measure per-object quality in addition to overall image quality. \emph{D2D} object fidelity is comparable with baselines.

\setlength{\aboverulesep}{1pt}
\setlength{\belowrulesep}{0pt}
\setlength{\extrarowheight}{.4ex}
\setlength{\tabcolsep}{6pt}
\begin{table}[h]
\centering
\small
\caption{\textbf{\emph{D2D} object fidelity is comparable with baselines.} PickScore and CLIPScore computed per detected object instance. On D2D-Small, across four seeds.}
\label{tab:per_obj_quality}
\begin{tabular}{lcc}
\toprule
 \multicolumn{1}{c}{Method} & PickScore~\cite{kirstain2023pickscore} & CLIPScore~\cite{hessel2021clipscore} \\
\midrule
\rowcolor{gray!22} SD1.4~\cite{rombach2022sd}   & 19.29 & 28.26 \\
\:\:\: + Counting Guidance~\cite{kang2025cg} & 19.04 & 27.51 \\
\midrule
\rowcolor{gray!22} SDXL~\cite{podell2024sdxl}   & 19.51 & 28.80 \\
\:\:\: + Make It Count~\cite{binyamin2025mic}   & 19.80 & 28.55 \\
\midrule
\rowcolor{gray!22} SDXL-Turbo~\cite{sauer2025turbo}   & 19.62 & 28.41 \\
\:\:\: + TokenOpt~\cite{zafar2026tokenopt}        & 19.65 & 28.45 \\
\:\:\: + \emph{D2D}             & 19.72 & 28.61 \\
\bottomrule
\end{tabular}
\end{table}
\setlength{\aboverulesep}{\oldaboverulesep}
\setlength{\belowrulesep}{\oldbelowrulesep}
\setlength{\extrarowheight}{\oldextrarowheight}
\setlength{\tabcolsep}{\oldtabcolsep}

\section{\emph{D2D} yields boosts in numeracy, in complement with baselines} \label{app:d2d_complement}
\setlength{\aboverulesep}{1pt}
\setlength{\belowrulesep}{1pt}
\setlength{\extrarowheight}{.4ex}
\begin{table}[H]
  \caption{\textbf{\emph{D2D} additionally yields improvements, in combination with T2I counting/enhancement baselines.} Standard deviations indicate the significance of our findings. Avg. over four seeds.}
  \label{tab:d2d_complement}
  \centering
  \resizebox{\textwidth}{!}{
  \begin{tabular}{l p{9em} c c c c}
    \toprule
    \multicolumn{1}{c}{Base Model} & \multicolumn{1}{c}{Method} & CoCoCount~\cite{binyamin2025mic} & D2D-Small & D2D-Multi & D2D-Large \\ \midrule
    \multicolumn{1}{l}{\multirow{4}{*}{SDXL-Turbo~\cite{sauer2025turbo}}} & TokenOpt~\cite{zafar2026tokenopt} & 35.12$\pm$0.75 & 23.31$\pm$1.66 & ----- &  3.94$\pm$0.72 \\
    & TokenOpt~\cite{zafar2026tokenopt} + \emph{D2D} & \textbf{48.75}$\pm$2.40 & \textbf{34.00}$\pm$1.70 & ----- & \textbf{8.56}$\pm$1.28 \\ \cdashline{2-6}
    & ReNO \cite{eyring2024reno} & 41.88$\pm$1.03 & 27.50$\pm$0.68 & 5.31$\pm$0.38 & 4.69$\pm$1.25 \\
    & ReNO \cite{eyring2024reno} + \emph{D2D} & \textbf{54.38}$\pm$2.17 & \textbf{41.25}$\pm$2.45 & \textbf{9.69}$\pm$0.38 & \textbf{10.38}$\pm$0.72 \\ \midrule
    \multicolumn{1}{l}{\multirow{2}{*}{SD-Turbo~\cite{sauer2025turbo}}} & ReNO \cite{eyring2024reno} & 43.38$\pm$3.47 & 32.06$\pm$0.99 & 8.94$\pm$1.76 &  4.25$\pm$1.14 \\
    & ReNO \cite{eyring2024reno} + \emph{D2D} & \textbf{53.62}$\pm$2.02 & \textbf{42.44}$\pm$2.82 & \textbf{11.62}$\pm$1.30 & \textbf{10.19}$\pm$1.77 \\ \midrule
    \multicolumn{1}{l}{\multirow{2}{*}{Pixart-DMD~\cite{chen2025pixartdmd}}} & ReNO \cite{eyring2024reno} & 44.75$\pm$1.44 & 37.25$\pm$1.70 & 9.44$\pm$0.75 & 4.75$\pm$0.74 \\
    & ReNO \cite{eyring2024reno} + \emph{D2D} & \textbf{52.50}$\pm$2.27 & \textbf{40.62}$\pm$2.37 & \textbf{12.00}$\pm$0.68 & \textbf{8.12}$\pm$0.92 \\
    \bottomrule
  \end{tabular}
 }
\end{table}
\setlength{\aboverulesep}{\oldaboverulesep}
\setlength{\belowrulesep}{\oldbelowrulesep}
\setlength{\extrarowheight}{\oldextrarowheight}
\setlength{\tabcolsep}{\oldtabcolsep}

\newpage
\section{Density-based differential performance persists in D2D-Multi}\label{app:multi_lowhigh}
\setlength{\aboverulesep}{1pt}
\setlength{\belowrulesep}{0pt}
\setlength{\extrarowheight}{.5ex}
\setlength{\tabcolsep}{10pt}
\begin{table}[H]
  \caption{\textbf{Quantitative results.} \emph{D2D} yields higher performance on low total-density prompts than high total-density ones. Avg. over four seeds.}
  \label{tab:d2dmulti_density}
  \centering
  \resizebox{0.8\textwidth}{!}{
  \begin{tabular}{p{17em} c c}
    \toprule
    \addlinespace[2pt]
    \multicolumn{1}{c}{\multirow{2}{*}{\textbf{Method}}} & \multicolumn{2}{c}{D2D-Multi} \\
    \cmidrule{2-3}
    & Low Total Density & High Total Density \\ \midrule
    \rowcolor{gray!22} SDXL \small{\cite{podell2024sdxl}} & 3.08 & 0.50 \\
    \:\:\: Make It Count \small{\cite{binyamin2025mic}} & ----- & ----- \\
    \rowcolor{gray!22} SDXL-Turbo \small{\cite{sauer2025turbo}} & 2.75 & 0.25 \\ 
    \:\:\: ReNO \small{\cite{eyring2024reno}} & 6.67 & 1.25  \\
    \:\:\: TokenOpt \small{\cite{zafar2026tokenopt}} & ----- & -----  \\
    \:\:\: \emph{D2D} w/ OWLv2 (Ours) & \textbf{12.08} & \textbf{3.00} \\ 
    \:\:\: \emph{D2D} w/ YOLOv9 (Ours) & \textbf{\textit{7.83}} & \textbf{\textit{1.50}} \\ 
    \midrule
    \rowcolor{gray!22} SD2.1 \small{\cite{rombach2022sd}} & 6.17 & 0.75  \\
    \rowcolor{gray!22} SD1.4 \small{\cite{rombach2022sd}} & 3.75 &  0.00 \\
    \:\:\: Counting Guidance \small{\cite{kang2025cg}} & 4.42 & 0.25 \\
    \rowcolor{gray!22} SD-Turbo \small{\cite{sauer2025turbo}} & 3.00 & 1.25 \\
    \:\:\: ReNO \small{\cite{eyring2024reno}} & 11.58 & 1.00  \\
    \:\:\: \emph{D2D} w/ OWLv2 (Ours) & \textbf{13.42} &  \textbf{2.75} \\
    \midrule
    \rowcolor{gray!22} Pixart-$\alpha$ \small{\cite{chen2024pixartalpha}} & 1.67 & 0.25 \\
    \rowcolor{gray!22} Pixart-DMD \small{\cite{chen2025pixartdmd}} & 8.00 & 1.00  \\
    \:\:\: ReNO \small{\cite{eyring2024reno}} & 12.25 &  1.00 \\
    \:\:\: \emph{D2D} w/ OWLv2 (Ours) & \textbf{17.17} & \textbf{1.75}  \\
    \bottomrule
  \end{tabular}
 }
\end{table}
\setlength{\aboverulesep}{\oldaboverulesep}
\setlength{\belowrulesep}{\oldbelowrulesep}
\setlength{\extrarowheight}{\oldextrarowheight}
\setlength{\tabcolsep}{\oldtabcolsep}

\section{Analysis of high-density performance}\label{app:high_density_analysis}

We further analyze \emph{D2D}'s performance in the high-density case by reporting numeracy, binned by object size, detector confidence, total count, and object class on D2D-Large/Multi. We use the mean absolute error $|$generated $-$ requested count$|$ as the evaluation metric. Unless otherwise noted, the bin ranges are followed by the error in parentheses, as follows: [bin start, bin end] (error $\downarrow$).

Tab.~\ref{tab:error_vs_size},~\ref{tab:error_vs_confidence}, and ~\ref{tab:error_vs_count} show the performance breakdown by quartiles. As expected, error decreases with larger objects, higher confidence, and lower total count. For higher counts, small object size remains a bottleneck, in particular for object classes that naturally appear in “herds” such as cows and sheep. Potential mitigation strategies could include adopting a bbox loss to explicitly increase object sizes at test-time and improving training sample diversity for the generative model (e.g. individual sheep, flocks).

\setlength{\aboverulesep}{1pt}
\setlength{\belowrulesep}{0pt}
\setlength{\extrarowheight}{.4ex}
\setlength{\tabcolsep}{6pt}
\begin{table}[h]
\centering
\small
\caption{\textbf{Error vs. object size. The error decreases as object size increases. \emph{D2D} tends to increase object size.} We bin the performance by object size (quartiles), calculated as $\frac{\text{instance size}}{\text{image size}} \times 100$. Error binned by object size quartile. Bin ranges in brackets, error ($\downarrow$) in parentheses. Base model SDXL-Turbo, across four seeds.}
\label{tab:error_vs_size}
\begin{tabular}{lllll}
\toprule
 \multicolumn{1}{c}{Method} & \multicolumn{1}{c}{Quartile 1} & \multicolumn{1}{c}{Quartile 2} &\multicolumn{1}{c}{Quartile 3} & \multicolumn{1}{c}{Quartile 4} \\
\midrule
\rowcolor{gray!22} \multicolumn{5}{l}{D2D-Large} \\
\:\:\:  Base   & $[0.3, 2.6]$ (\textbf{52.2}) & $(2.6, 4.5]$ (\textbf{35.2}) & $(4.5, 6.9]$ (\textbf{21.2}) & $(6.9, 92.4]$ (\textbf{13.3}) \\
\:\:\: + \emph{D2D}  & $[0.1, 3.5]$ (\textbf{21.1}) & $(3.5, 5.2]$ (\textbf{17.9}) & $(5.2, 7.7]$ (\textbf{11.7}) & $(7.7, 93.6]$ (\textbf{12.0}) \\
\midrule
\rowcolor{gray!22} \multicolumn{5}{l}{D2D-Multi} \\
\:\:\: Base   & $[0.4, 3.9]$ (\textbf{19.8}) & $(3.9, 6.6]$ (\textbf{8.2}) & $(6.6, 10.3]$ (\textbf{5.1}) & $(10.3, 54.3]$ (\textbf{3.5}) \\
\:\:\: + \emph{D2D}  & $[0.9, 5.3]$ (\textbf{9.2})  & $(5.3, 8.3]$ (\textbf{7.4}) & $(8.3, 12.5]$ (\textbf{4.5}) & $(12.5, 70.2]$ (\textbf{3.1}) \\
\bottomrule
\end{tabular}
\end{table}
\setlength{\aboverulesep}{\oldaboverulesep}
\setlength{\belowrulesep}{\oldbelowrulesep}
\setlength{\extrarowheight}{\oldextrarowheight}
\setlength{\tabcolsep}{\oldtabcolsep}

\newpage
\setlength{\aboverulesep}{1pt}
\setlength{\belowrulesep}{0pt}
\setlength{\extrarowheight}{.4ex}
\setlength{\tabcolsep}{6pt}
\begin{table}[h]
\centering
\small
\caption{\textbf{Error vs. detector confidence. The error decreases with higher detector confidence. \emph{D2D} tends to increase overall detector confidence.} Error binned by detector confidence quartile. Bin ranges in brackets, error ($\downarrow$) in parentheses. Base model SDXL-Turbo, across four seeds.}
\label{tab:error_vs_confidence}
\begin{tabular}{lllll}
\toprule
 \multicolumn{1}{c}{Method} & \multicolumn{1}{c}{Quartile 1} & \multicolumn{1}{c}{Quartile 2} &\multicolumn{1}{c}{Quartile 3} & \multicolumn{1}{c}{Quartile 4} \\
\midrule
\rowcolor{gray!22} \multicolumn{5}{l}{D2D-Large} \\
\:\:\: Base   & $[0.20, 0.38]$ (\textbf{70.3}) & $(0.38, 0.45]$ (\textbf{26.8}) & $(0.45, 0.51]$ (\textbf{14.2}) & $(0.51, 0.80]$ (\textbf{10.4}) \\
\:\:\: + \emph{D2D}  & $[0.22, 0.40]$ (\textbf{35.3}) & $(0.40, 0.48]$ (\textbf{18.8}) & $(0.48, 0.55]$ (\textbf{5.6})  & $(0.55, 0.78]$ (\textbf{3.1})  \\
\midrule
\rowcolor{gray!22} \multicolumn{5}{l}{D2D-Multi} \\
\:\:\: Base   & $[0.25, 0.42]$ (\textbf{18.7}) & $(0.42, 0.50]$ (\textbf{8.1})  & $(0.50, 0.57]$ (\textbf{5.8})  & $(0.57, 0.88]$ (\textbf{4.0})  \\
\:\:\: + \emph{D2D}  & $[0.24, 0.44]$ (\textbf{14.6}) & $(0.44, 0.52]$ (\textbf{4.3})  & $(0.52, 0.59]$ (\textbf{3.2})  & $(0.59, 0.85]$ (\textbf{2.1})  \\
\bottomrule
\end{tabular}
\end{table}
\setlength{\aboverulesep}{\oldaboverulesep}
\setlength{\belowrulesep}{\oldbelowrulesep}
\setlength{\extrarowheight}{\oldextrarowheight}
\setlength{\tabcolsep}{\oldtabcolsep}

\setlength{\aboverulesep}{1pt}
\setlength{\belowrulesep}{0pt}
\setlength{\extrarowheight}{.4ex}
\setlength{\tabcolsep}{6pt}
\begin{table}[h]
\centering
\small
\caption{\textbf{Error vs. total requested count. The error increases with higher total requested counts}, likely due to smaller object instances, which aligns with the observations above. Base model error and \emph{D2D} error ($\downarrow$) binned by total requested count. Base SDXL-Turbo, across four seeds.}
\label{tab:error_vs_count}
\begin{tabular}{lll}
\toprule
\rowcolor{gray!22} \multicolumn{3}{c}{D2D-Large} \\
 \multicolumn{1}{c}{Requested count} & \multicolumn{1}{c}{Base} & \multicolumn{1}{c}{+ \emph{D2D}} \\
\midrule
\:\:\: $[11, 13]$ & 16.8 & \textbf{8.2}  \\
\:\:\: $(13, 15]$ & 32.4 & \textbf{16.4} \\
\:\:\: $(15, 18]$ & 34.7 & \textbf{18.2} \\
\:\:\: $(18, 20]$ & 42.5 & \textbf{22.4} \\
\bottomrule
\end{tabular}
\hspace{25pt}
\begin{tabular}{lll}
\toprule
\rowcolor{gray!22} \multicolumn{3}{c}{D2D-Multi} \\
 \multicolumn{1}{c}{Requested count} & \multicolumn{1}{c}{Base} & \multicolumn{1}{c}{+ \emph{D2D}} \\
\midrule
\:\:\: $[2, 4]$   & 4.2  & \textbf{2.2}  \\
\:\:\: $(4, 5]$   & 6.0  & \textbf{3.3}  \\
\:\:\: $(5, 10]$  & 9.1  & \textbf{5.7}  \\
\:\:\: $(10, 19]$ & 17.6 & \textbf{13.0} \\
\bottomrule
\end{tabular}
\end{table}
\setlength{\aboverulesep}{\oldaboverulesep}
\setlength{\belowrulesep}{\oldbelowrulesep}
\setlength{\extrarowheight}{\oldextrarowheight}
\setlength{\tabcolsep}{\oldtabcolsep}

\smallsec{Object categories with highest error}
Sheep, book and cow are among the top four most difficult categories for \emph{D2D} on both D2D-Large and D2D-Multi, which may be explained by their tendency to occur in large groups (i.e., flock of sheep, shelf of books, etc).

\section{How \emph{D2D} handles overlapping detections}\label{app:overlap_analysis}

To analyze how \emph{D2D} handles overlapping detections, we compare bbox overlap before and after \emph{D2D} correction in the high-density setting, where overlap is more likely. We define an image’s overall “overlap score” as the global overlap-to-union ratio: $\frac{(\sum_i{\operatorname{Area}(B_i)}) - \operatorname{Area}(\bigcup_i B_i)}{\operatorname{Area}(\bigcup_i B_i)}$. Furthermore, the mean absolute error of an image is: $|$requested $-$ generated count$|$.

Using base model SDXL-Turbo and prompt set D2D-Large, we bin the initial generations into quartiles based on their overlap scores. For each quartile, we report the mean initial error, and after applying \emph{D2D}, the new mean overlap score and error. Tab.~\ref{tab:error_vs_overlap} details the results.

\setlength{\aboverulesep}{1pt}
\setlength{\belowrulesep}{0pt}
\setlength{\extrarowheight}{.4ex}
\setlength{\tabcolsep}{6pt}
\begin{table}[h]
\centering
\small
\caption{\textbf{\emph{D2D} reduces error across all initial overlap levels.} Error and overlap before/after \emph{D2D}, binned by initial overlap quartile. On D2D-Large, base model SDXL-Turbo, across four seeds.}
\label{tab:error_vs_overlap}
\begin{tabular}{cccc}
\toprule
 Initial overlap bin & Initial mean abs. error & Post-\emph{D2D} mean overlap & Post-\emph{D2D} mean abs. error \\
\midrule
$[0.00, 0.10]$ & 12.19 & 0.07 & \textbf{7.47}  \\
$(0.10, 0.22]$ & 18.17 & 0.16 & \textbf{12.88} \\
$(0.22, 0.83]$ & 56.46 & 0.20 & \textbf{28.40} \\
$(0.83, 1.33]$ & 34.88 & 0.32 & \textbf{13.98} \\
\bottomrule
\end{tabular}
\end{table}
\setlength{\aboverulesep}{\oldaboverulesep}
\setlength{\belowrulesep}{\oldbelowrulesep}
\setlength{\extrarowheight}{\oldextrarowheight}
\setlength{\tabcolsep}{\oldtabcolsep}

\emph{D2D} reduces error across all initial overlap ranges. In cases of high initial overlap (e.g., the third and fourth quartiles), the post-D2D overlap drops below the initial overlap range, suggesting that \emph{D2D} actively reduces overlap. 

Because over-generation is particularly prevalent in high-count settings (e.g., 78\% of the initial images from above were over-generations), \emph{D2D} works to delete objects and reduces overlap.

\newpage
\section{\emph{D2D} vs. YOLO-COUNT~\cite{zeng2025yolocount}}\label{app:yolocount}
We include comparisons against a more recent work, YOLO-COUNT~\cite{zeng2025yolocount}. A key distinction between the two methods is that \emph{D2D} leverages detector logits out-of-the-box, without any additional detector or counter training, making it easily adaptable to newly developed detectors.

We evaluated YOLO-COUNT using their two pretrained checkpoints (on FSC147~\cite{ranjan2021fsc} and LVIS v1.0~\cite{gupta2019lvis}) on CoCoCount (seed = 0) and found that LVIS (47.5\%) performed stronger than FSC (40.5\%). We replace the strongest \emph{D2D} critic with YOLO-COUNT’s strongest checkpoint and compare performance on CoCoCount and D2D-Small/Multi/Large (averaged over four seeds). \emph{D2D} outperforms YOLO-COUNT across all four benchmarks on SDXL-Turbo.

\begin{table}[H]
  \caption{\textbf{\emph{D2D} vs. YOLO-COUNT on SDXL-Turbo.} \emph{D2D} outperforms YOLO-COUNT across all four benchmarks on SDXL-Turbo. Standard deviations indicate the significance of our findings. Avg. over four seeds.}
  \label{tab:yc_sdxlturbo}
  \centering
  \begin{tabular}{l c c c c}
    \toprule
\multicolumn{1}{c}{Method}   & CoCoCount~\cite{binyamin2025mic}   & D2D-Small       & D2D-Multi & D2D-Large \\
\midrule
YOLO-COUNT~\cite{zeng2025yolocount} & 44.38 $\pm$ 3.25   & 30.44 $\pm$ 2.69  & 5.69 $\pm$ 3.02 & 8.06 $\pm$ 0.90\\
\emph{D2D} & \textbf{55.62 $\pm$ 2.72}    & \textbf{43.69 $\pm$ 2.36 }     & \textbf{9.81 $\pm$ 0.97} & \textbf{9.94 $\pm$ 1.57}\\
    \bottomrule
  \end{tabular}
\end{table}

\section{\emph{D2D} vs. EnMMDiT~\cite{wei2026enhancing}}\label{app:enmmdit}

We additionally compare our method against EnMMDiT~\cite{wei2026enhancing}, a relevant work. The two methods address different aspects of generation: \emph{D2D} directly optimizes object counts using a detector-based differentiable critic, whereas EnMMDiT addresses subject mixing and neglect by leveraging a suite of losses that correct attention activation in earlier layers, text encoder inconsistency, and overlapping attentions of subjects.

We report numeracy for EnMMDiT (base model SD3~\cite{esser2024scaling}) and \emph{D2D} (base model SDXL-Turbo) using CoCoCount and a subset of 100 prompts from D2D-Multi. Standard deviations reported across four seeds.

\setlength{\aboverulesep}{1pt}
\setlength{\belowrulesep}{0pt}
\setlength{\extrarowheight}{.4ex}
\setlength{\tabcolsep}{6pt}
\begin{table}[h]
\centering
\small
\caption{\textbf{Although EnMMDiT builds on SD3 (a substantially stronger base model with nearly 2$\times$ numeracy over SDXL-Turbo), \emph{D2D} achieves comparable performance with EnMMDiT, as seen by the overlapping performance range.} On CoCoCount and 100-prompt subset of D2D-Multi. Standard deviations noted. Across four seeds.}
\label{tab:enmmdit}
\begin{tabular}{lcc}
\toprule
 \multicolumn{1}{c}{Method} & CoCoCount~\cite{binyamin2025mic} & D2D-Multi (subset) \\
\midrule
\rowcolor{gray!22} SD3~\cite{esser2024scaling}        & 54.38 $\pm$ 5.07 & 17.00 $\pm$ 3.83 \\
\:\:\: + EnMMDiT~\cite{wei2026enhancing}              & \textbf{58.62 $\pm$ 1.31} & \textbf{20.75 $\pm$ 4.99} \\
\midrule
\rowcolor{gray!22} SDXL-Turbo~\cite{sauer2025turbo} & 27.38 $\pm$ 2.69 & 3.50 $\pm$ 1.73 \\
\:\:\: + \emph{D2D}                  & \textbf{55.62 $\pm$ 2.72} & \textbf{15.25 $\pm$ 2.63} \\
\bottomrule
\end{tabular}
\end{table}
\setlength{\aboverulesep}{\oldaboverulesep}
\setlength{\belowrulesep}{\oldbelowrulesep}
\setlength{\extrarowheight}{\oldextrarowheight}
\setlength{\tabcolsep}{\oldtabcolsep}

\newpage
\section{CountGD in comparison with other counters}\label{app:countgd}
CountGD performs well in low and high-density settings.
\begin{figure}[H]
\centering
    \includegraphics[width=0.8\textwidth]{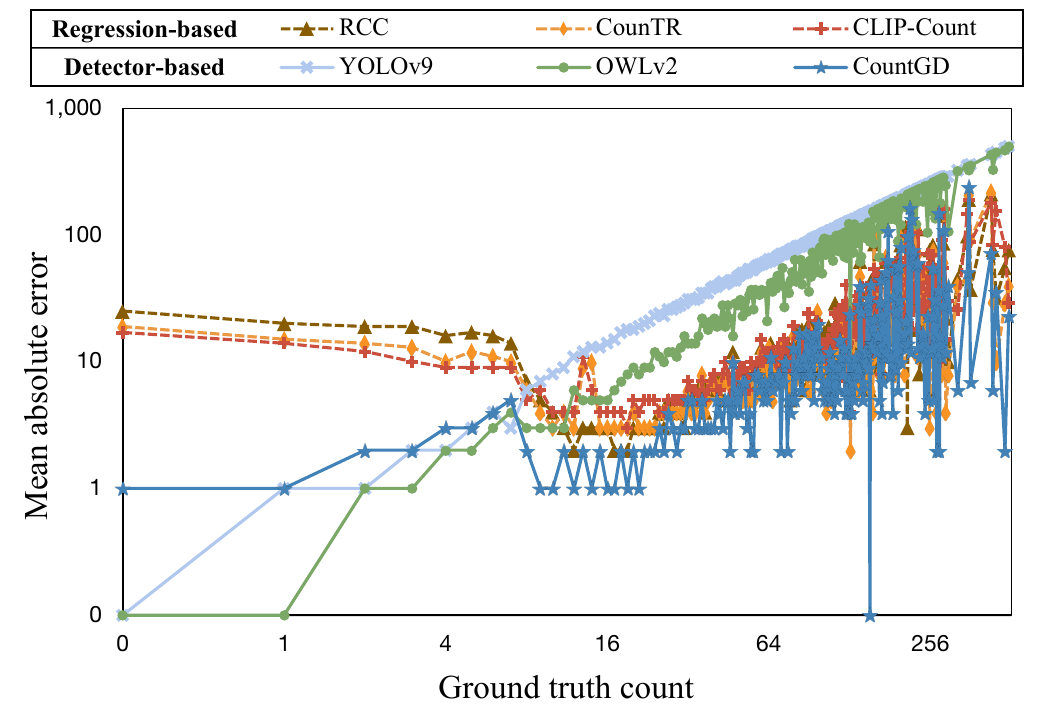}
    \caption{\textbf{CountGD} is a state-of-the-art counter, based on GroundingDINO.}
    \label{fig:countgd_eval}
    \centering
\end{figure}

\newpage
\section{Additional qualitative results}\label{app:add_qual}
Additional qualitative results on prompts from benchmarks CoCoCount~\cite{binyamin2025mic} and D2D-Small shown.

\begin{figure}[H]
  \begin{center}
    \includegraphics[width=\textwidth]{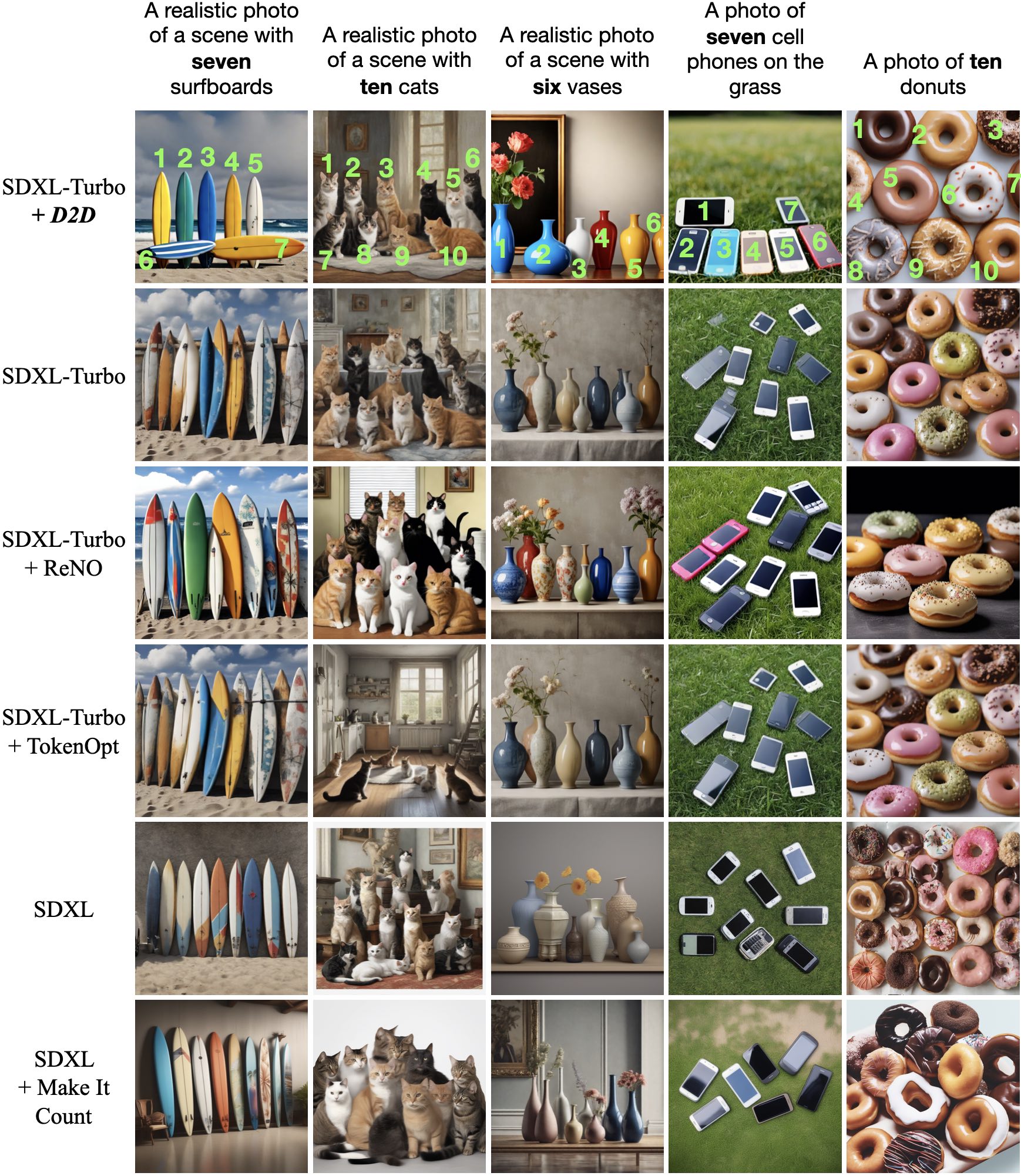}
  \end{center}
  \caption{\textbf{Qualitative examples} from count correction methods on base model SDXL-Turbo, along with SDXL.}
  \label{fig:add_sdxl}
\end{figure}
\hfill

\begin{figure}[H]
  \begin{center}
    \includegraphics[width=\textwidth]{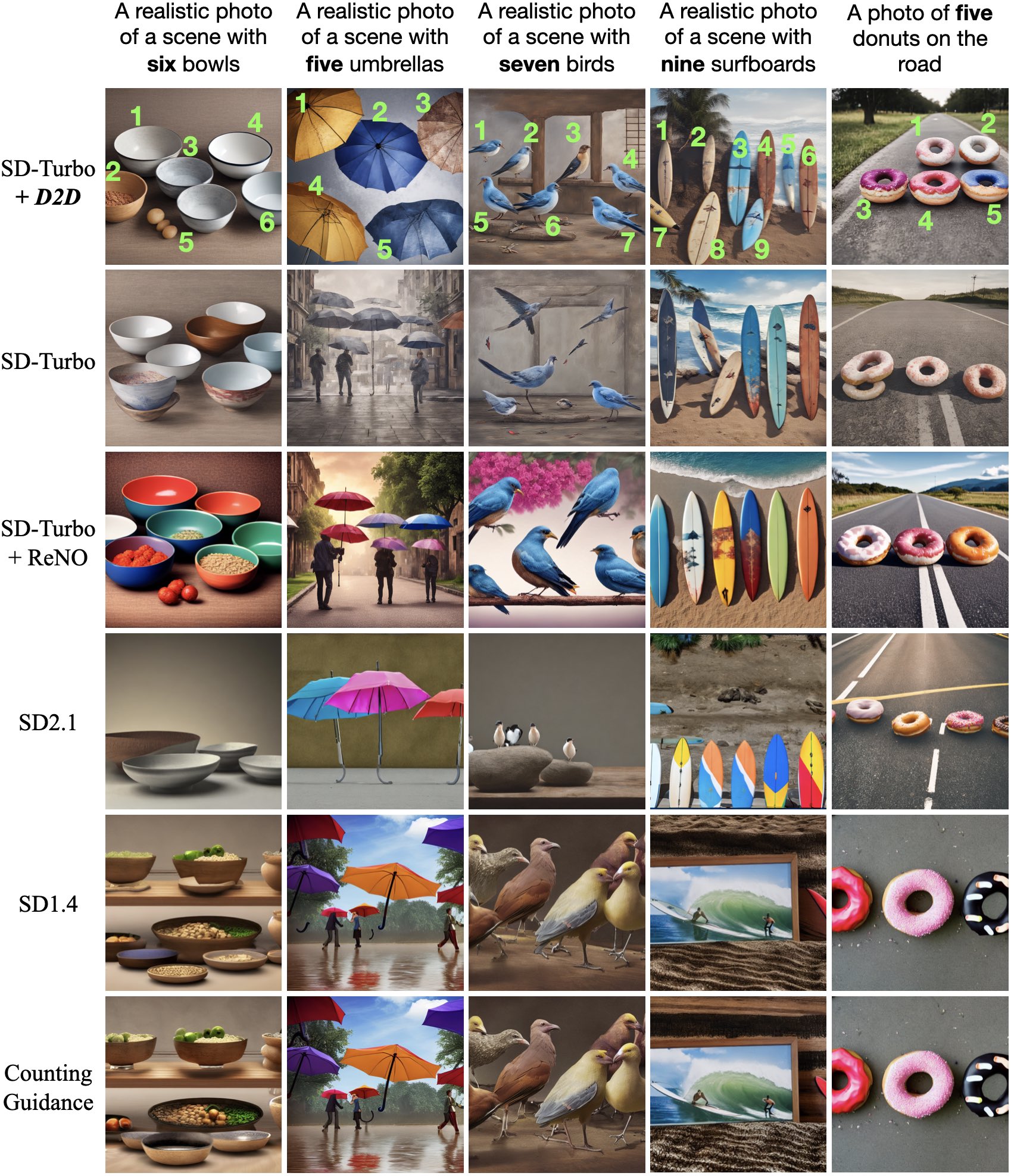}
  \end{center}
  \caption{\textbf{Qualitative examples} from count correction methods on base model SD-Turbo, along with SD1.4, SD2.1, and Counting Guidance~\cite{kang2025cg}.}
  \label{fig:add_sd}
\end{figure}

\begin{figure}[H]
  \begin{center}
    \includegraphics[width=\textwidth]{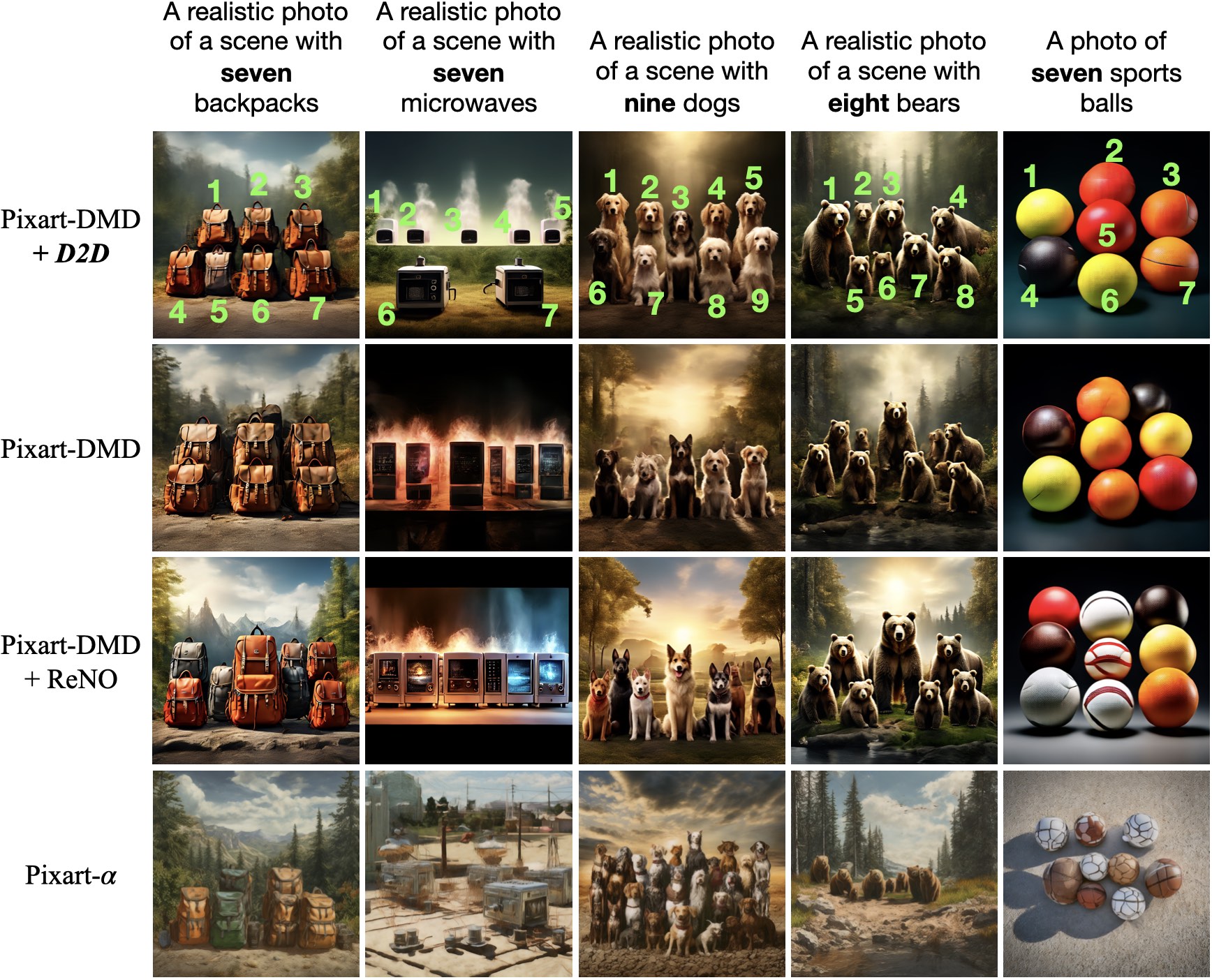}
  \end{center}
  \caption{\textbf{Qualitative examples} from count correction methods on base model Pixart-DMD, along with Pixart-$\alpha$.}
  \label{fig:add_pix}
\end{figure}

\subsection{\emph{D2D} on free-form prompts}\label{app:freeform}
\emph{D2D} as a method just requires the user to provide the prompt along with the desired (object, count) pair. While our quantitative evaluation focuses on prompts that follow structured formats, we provide additional qualitative figures on more free-form prompts.

\begin{figure}[H]
\centering
    \includegraphics[width=0.8\textwidth]{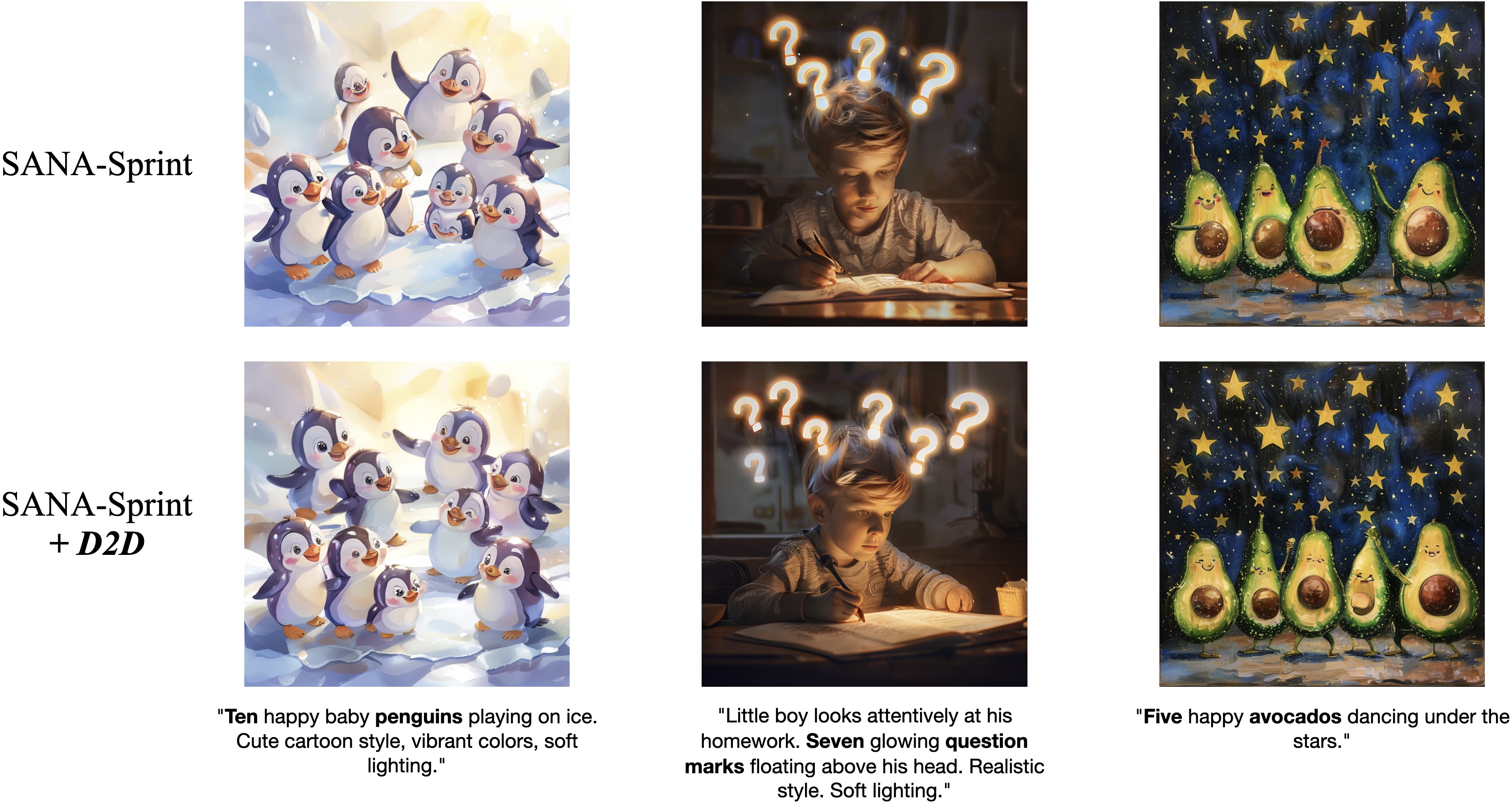}
    \caption{\textbf{Qualitative examples} of \emph{D2D} on more free-form prompts using base model SANA-Sprint~\cite{chen2025sana}.}
    \label{fig:freeform}
    \centering
\end{figure}

\newpage
\section{Additional ablations}\label{app:add_abl}

\subsection{\boldmath$w$ = 0.2 is optimal for numeracy and quality}\label{app:w_abl}
\begin{figure}[H]
    \centering
    \includegraphics[width=0.5\textwidth]{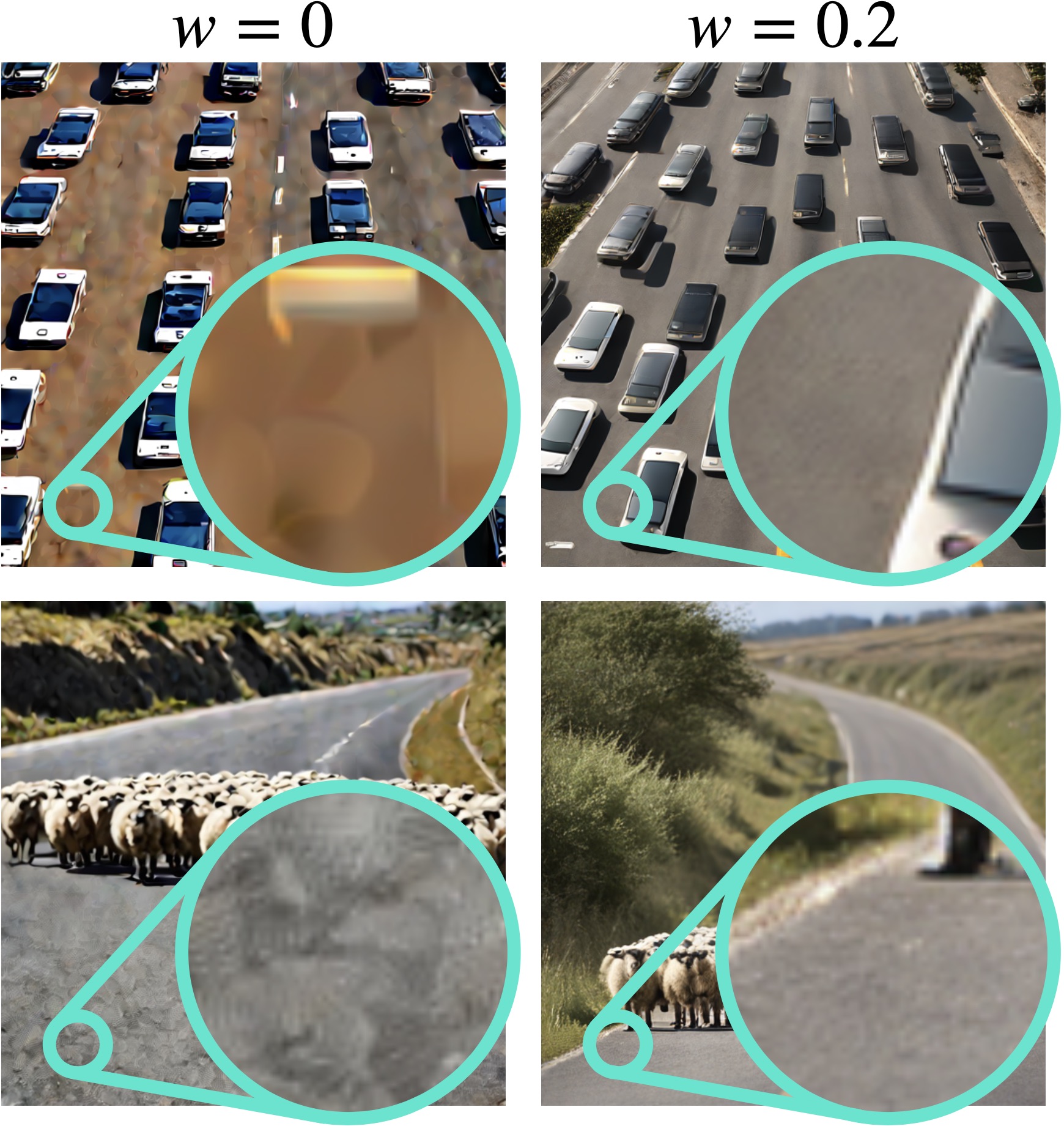}
    \caption{\textbf{$\bm{w}$ = 0 results in patchy visual artifacts, relative to $\bm{w}$ = 0.2.} So, we use $w$ = 0.2.}
    \label{fig:w_abl_qual}
    \centering
\end{figure}

$w$ is the weight used to determine how to mix the LMN output with the original noise. $w$ = 0 is numeracy-wise the best, followed by $w$ = 0.2, but results in patchy visual artifacts, relative to $w$ = 0.2 (Fig.~\ref{fig:w_abl_qual}). Hence, the hyperparameter value we use in our main experiments is $w$ = 0.2.
    \setlength{\aboverulesep}{2pt}
    \setlength{\belowrulesep}{0pt}
    \setlength{\extrarowheight}{.6ex}
    \setlength{\tabcolsep}{3pt}
    \begin{table}[H]
    
      \caption{\textbf{In terms of just numeracy, $\bm{w}$ = 0 is optimal, followed by $\bm{w}$ = 0.2.} Base model SDXL-Turbo, on CoCoCount~\cite{binyamin2025mic}, seed = 0.}
      \label{tab:abl_w}
      \centering
      \begin{tabular}{c p{1em} @{} c c c c}
        \toprule
        && \multicolumn{4}{c}{$\bm{w}$} \\ 
        \cmidrule{3-6} 
         && 0.0 & 0.2 & 0.5 & 0.8 \\
        \midrule
        Numeracy && \textbf{62.50} & 55.50 & 37.50 & 48.00 \\
        \bottomrule
      \end{tabular}
    \end{table}
    \setlength{\aboverulesep}{\oldaboverulesep}
    \setlength{\belowrulesep}{\oldbelowrulesep}
    \setlength{\extrarowheight}{\oldextrarowheight}
    \setlength{\tabcolsep}{\oldtabcolsep}

\newpage

\subsection{LMN architecture}\label{app:lmn_arc}
For the choice of 3-layer MLP, we perform additional ablation studies and present the results in Tab.~\ref{tab:lmn_arc}. Between MLP architectures, the 3-layer, 100-width and 200-width variants perform the best, but the former is more parameter-efficient. We also try using a 6-layer CNN (with adjusted kernels and channels to approximate the parameter count of the 3-layer, 100-width MLP), which yielded a slight 1\% boost in numeracy. Our results demonstrate that even a small and simple MLP works quite well as an LMN architecture.

\begin{table}[H]
  \caption{\textbf{Even a small and simple 3-layer, 100-width MLP works effectively for boosting numeracy.} On CoCoCount~\cite{binyamin2025mic}. On SDXL-Turbo, seed=0.}
  \label{tab:lmn_arc}
  \centering
  
   \resizebox{0.6\textwidth}{!}{
\begin{tabular}{lc}
\toprule
\multicolumn{1}{c}{LMN Architecture} & Numeracy \\ \midrule
3-layers, width 100 & 55.5 \\ \midrule
2-layers, width 100 &  52.5 \\ \midrule
4-layers, width 100 & 52.0  \\ \midrule
3-layers, width 50 &  54.5 \\ \midrule
3-layer, width 200 & 55.5  \\ \midrule
CNN (approximate parameter-count match)  & 56.5 \\
\bottomrule
\end{tabular}
  }
\end{table}

\subsection{Pre-inference alignment of \boldmath$M_\phi$}\label{app:preinf_align}

We conduct experiments without the pre-inference alignment stage and report below (Tab.~\ref{tab:preinfalign_exp}). We find that without pre-inference alignment, it generally takes an additional 20-30 steps during inference time to align the network to Gaussian outputs, which effectively reduces the number of steps available to perform numeracy optimization, because we cap the max number of steps.

\begin{table}[H]
  \caption{\textbf{Pre-inference alignment frees up more inference-time steps to be dedicated to numeracy optimization,} yielding higher numeracy. On CoCoCount~\cite{binyamin2025mic}, using SDXL-Turbo, seed=0.}
  \label{tab:preinfalign_exp}
  \centering
  
   \resizebox{0.4\textwidth}{!}{
        \begin{tabular}{ccc}
        \toprule
        Method & Numeracy \\ \midrule
        w/o pre-inference alignment & 47.0 \\
        w/ pre-inference alignment & \textbf{55.5} \\
        \bottomrule
      \end{tabular}
  }
\end{table}

\subsection{Inference-time calibration of \boldmath$M_\phi$}\label{app:lmd_infalign}
\begin{figure}[H]
    \hspace{4em}
     \begin{subfigure}[b]{0.3\textwidth}
         \centering
         \includegraphics[scale=0.15]{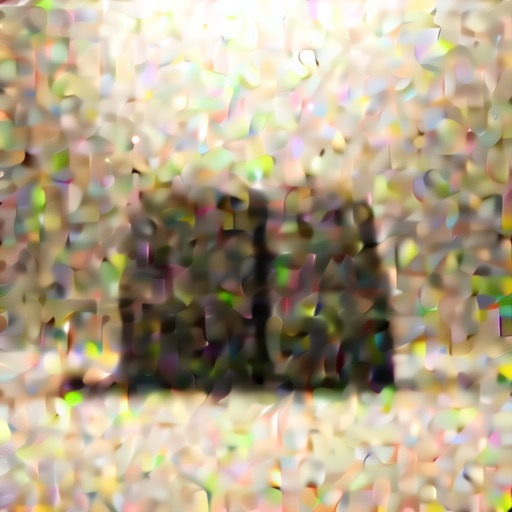}
         \caption{Without alignment.}
         \label{fig:woalign}
     \end{subfigure}
     \hfill
     \begin{subfigure}[b]{0.3\textwidth}
         \centering
         \includegraphics[scale=0.15]{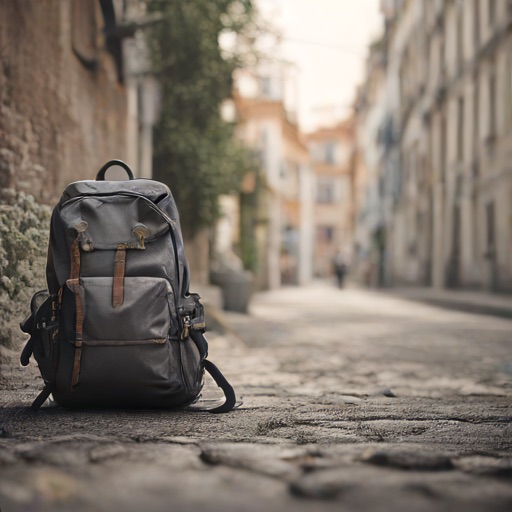}
         \caption{With alignment.}
         \label{fig:walign}
     \end{subfigure}
     \hspace{4em}
     \caption{\textbf{Inference-time alignment of $\bm{M_\phi}$ is necessary.} Prompt: ``A realistic photo of a scene with one backpack.''}
    \label{fig:alignment}
\end{figure}

At inference-time, given a new $\mathbf{x}_T$, we allow a few iterations to calibrate $M_\phi$'s output to the Gaussian distribution. Fig.~\ref{fig:alignment} shows alignment is crucial to maintaining high image quality.

\newpage
\subsection{Regularization}\label{app:reg}
Our regularization term (Eq.~\ref{eq:our_reg}) is a variant of the one ReNO~\cite{eyring2024reno} uses (Eq.~\ref{eq:reno_reg}) that penalizes larger deviations from Gaussian more and smaller deviations less, to allow enough flexibility in the vicinity of the initial distribution to accommodate updates specified by the count critic. We scale $\mathcal{L}_{\text{reg}}'$ by coefficient $a$ and shift by constant $c$, which we take to the power of 10.

\begin{figure}[H]
\centering
    \includegraphics[width=0.5\textwidth]{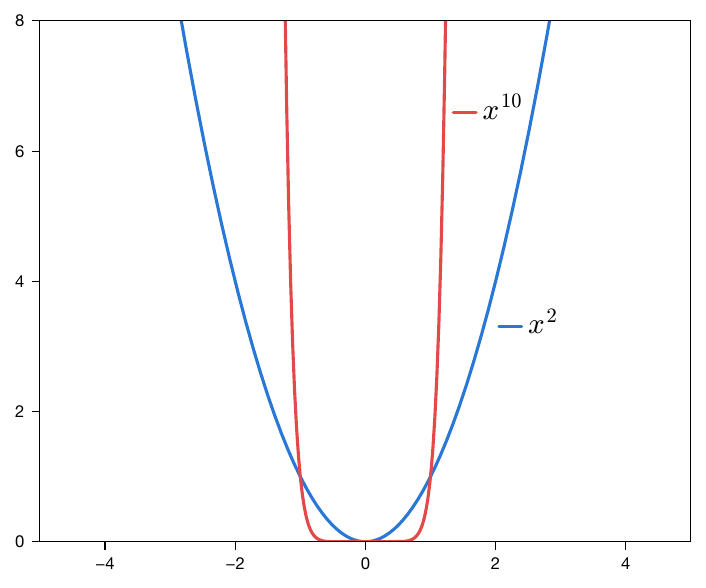}
    \caption{\textbf{Exponentiating by the power of 10 sharpens the gradients far away from the minimum while flattening the gradient closer to the minimum}, effectively allowing the optimization process to focus more on numeracy optimization when the latent is already well-regularized.}
    \label{fig:sharper_reg}
    \centering
\end{figure}

\begin{equation}\label{eq:reno_reg}
    \mathcal{L}_{\text{reg}}' = \frac{||\mathbf{x}_T'||^2}{2} - (d - 1) \cdot \text{log}(||\mathbf{x}_T'||),
\end{equation}
\begin{equation}\label{eq:our_reg}
    \mathcal{L}_{\text{reg}} = (a\mathcal{L}_{\text{reg}}' + c)^{10}.
\end{equation}

We set coefficient $a$ to 0.03. Shift constant $c$ is set to 2139 to offset the optimal value of $a\mathcal{L}_{\text{reg}}'$, which we find to be -2139 in practice.

Our ablation analysis confirms that this variant of the regularization leads to overall numeracy improvements across both CoCoCount and D2D-Small (Tab.~\ref{tab:reg}), indicating our regularization objective allows for effective numeracy correction, even while enforcing a heavier penalty on non-Gaussian initial noises.

\begin{table}[H]
  \caption{\textbf{Our variant of regularization is beneficial to numeracy optimization.} $\mathcal{L}_{\text{reg}}$, which penalizes larger deviations from Gaussian more and smaller deviations less, yields higher numeracy on CoCoCount and D2D-Small benchmarks. On SDXL-Turbo. Avg. over four seeds.}
  \label{tab:reg}
  \centering
  
        \begin{tabular}{lcc}
        \toprule
        \multicolumn{1}{c}{Method} & CoCoCount~\cite{binyamin2025mic} & D2D-Small \\ \midrule
        \emph{D2D} w/ $\mathcal{L}_{\text{reg}}'$ & 53.88 & 42.44 \\ \midrule
        \emph{D2D} w/ $\mathcal{L}_{\text{reg}}$ & \textbf{55.62} & \textbf{43.69} \\
        \bottomrule
      \end{tabular}
\end{table}

\begin{table}[H]
  \caption{\textbf{Our variant of regularization does not significantly impact diversity}, as measured by Vendi score~\cite{friedman2023vendi}, on CoCoCount~\cite{binyamin2025mic} and D2D-Small. On SDXL-Turbo. Across four seeds.}
  \label{tab:reg_div_metrics}
  \centering
  
\begin{tabular}{lc}
\toprule
Method & Vendi \\ \midrule
 \emph{D2D} w/ $\mathcal{L}_{\text{reg}}'$ & 1.82 \\ \midrule
\emph{D2D} w/ $\mathcal{L}_{\text{reg}}$ & 1.82 \\
\bottomrule
\end{tabular}
\end{table}

\newpage
\section{\emph{D2D} scaling formulation}\label{app:d2d_scalingform}

\begin{table}[H]
  \caption{\textbf{Our \emph{D2D} loss formulation is effective.} We compare our formulation against several alternatives. Ours yields the highest numeracy with fastest inference. On CoCoCount~\cite{binyamin2025mic}. On SDXL-Turbo, seed=0.}
  \label{tab:scaling_form}
  \centering
  
   \resizebox{0.7\textwidth}{!}{
\begin{tabular}{lcc}
\toprule
\multicolumn{1}{c}{\textbf{Scaling formulation}} & \textbf{Numeracy} & \textbf{Inference time (s)} \\ \midrule
$\sigma (\beta(z-\tau)) \cdot (z-\tau)$ (ours) & 55.50 & 17.10 \\ \midrule
$e^{z-\tau}$ & 52.00 & 56.28\\ \midrule
$e ^ {\sigma(\beta(z - \tau))}$  & 31.50 & 32.24\\ \midrule
$z-\tau$  & 50.50    &   53.23             \\ \midrule
$\ln(1+e^{z-\tau})$    &   55.00    &   20.09             \\ \midrule
 $(z-\tau) \cdot e^{z-\tau}$   &   49.00    &   72.72             \\ \midrule
 $(z-\tau) \cdot \ln(1+e^{z-\tau})$    &   54.00    &   22.31             \\ \midrule
$\sigma(\beta(z-\tau)) \cdot e^{z-\tau}$  &   50.50   &   63.96             \\ \midrule
$\sigma(\beta(z-\tau)) \cdot \ln(1+e^{z-\tau})$    &   55.00    &   18.65    \\
\bottomrule
\end{tabular}
  }
\end{table}

\begin{figure}[H]
\centering
    \includegraphics[width=0.7\textwidth]{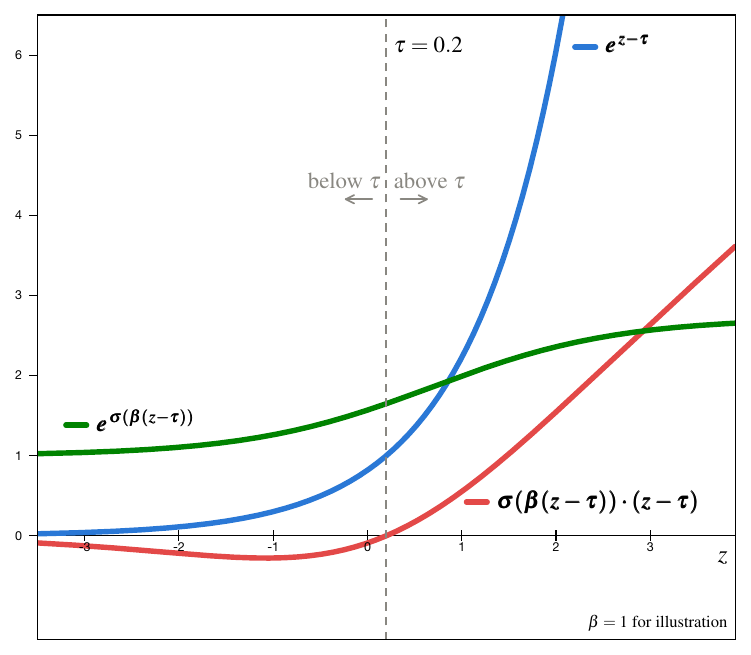}
    \caption{\textbf{Design of \emph{D2D} loss.} We want to design a fully-smooth loss that will effectively update each bbox's logit in order to bring the total count closer to the requested count. To illustrate, if the total generated count $>$ requested count, then the overall objective is to bring the logits down. However, practically speaking, on the level of individual logits, this means focusing on decreasing the logits that are above the $\tau$ and allowing some flexibility for logits that are already below $\tau$ so that the optimization can focus on the $>\tau$ logits. In other words, we want a strong gradient signal above $\tau$ and little signal below. This plots the curves of the first three scaling formulations from the table above, illustrating how they behave above and below $\tau$. Our formulation (in \textcolor{red}{red}) has a strong signal above and weak signal below $\tau$. $e^{z-\tau}$ still maintains a significant gradient below $\tau$. $e^{\sigma(\beta(z-\tau))}$ still shows plateauing above $\tau$.}
    \label{fig:scaling_form}
    \centering
\end{figure}

\newpage
\section{Sample numeracy optimization trajectory}\label{app:opt_traj}
\begin{figure}[H]
\centering
    \includegraphics[width=0.8\textwidth]{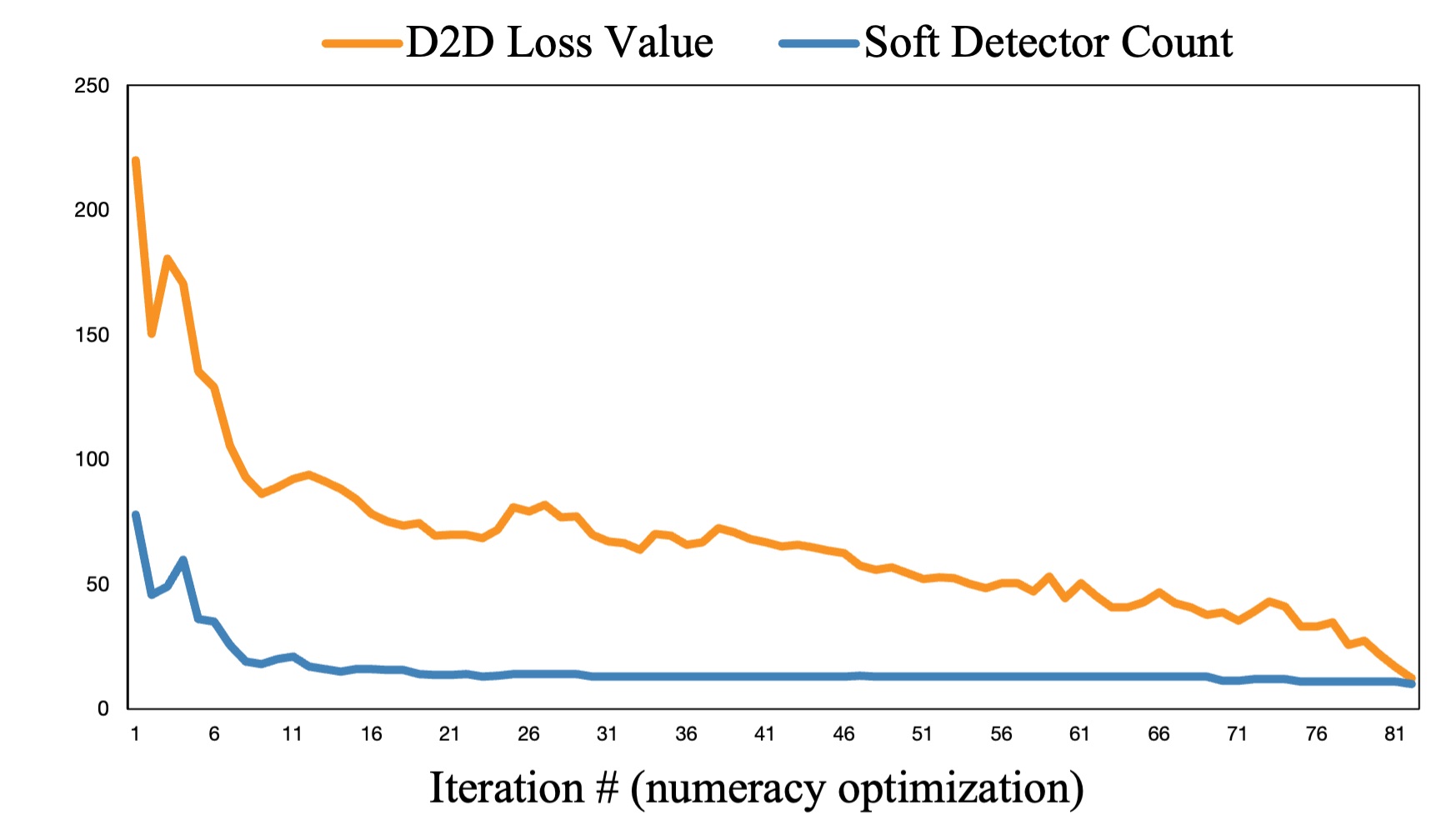}
    \caption{\textbf{Sample trajectory of $\bm{\mathcal{L}_{\text{D2D}}}$} and the soft detector count $f$ over numeracy optimization steps for the prompt ``A photo of ten sheep on the grass'' from CoCoCount~\cite{binyamin2025mic}. Though the two curves do not have the same raw value, their coarse trends do correlate.}
    \label{fig:opt_traj}
    \centering
\end{figure}

\newpage
\section{Selective \emph{D2D} activation}\label{app:selective_d2d}

\emph{D2D} focuses on numerical prompts, but not all prompts are numeric in nature. Furthermore, prompts may lack concrete numerical targets, as in the prompt ``A few cups.'' From the perspective of user intention, using words, like ``few'' or ``many,'' that lack clear numerical grounding may indicate lack of strict numerical target in mind. So, it is reasonable to allow some flexibility in mapping those words to numbers. At test-time, we can leverage an LLM to determine whether a prompt is a numerical one and, if so, to extract the (object, requested count) pairs. Below is a sample context for this task.

\vspace{0.5in}
\begin{tcolorbox}[colback=gray!5,colframe=gray!50,
                  title=In-context prompting for \emph{D2D} selective activation]
\small

You are an expert prompt analyzer. Given a text-to-image prompt, you extract two things:\\

1. Boolean indicator of whether the prompt specifies a count of objects. Examples include ``Six dogs,'' ``Some cats.'' It is not a countable prompt if all the objects are uncountable nouns. For the prompt to be a counting prompt, at least one object must be countable.\\
2. For each **countable** object and requested count, extract this information (object in singular form, object in form seen in the prompt, requested count in number form). So, for prompts with multiple objects and corresponding requested counts, you would extract a list like this [(object in singular form, object in form seen in the prompt, requested count), (object in singular form, object in form seen in the prompt, requested count), ...].\\

\#\#\#\#\#\#\#\#\#\#\#\#\#\#\#\#\#\#\#\#\#\#\#\#\#\\
Examples:\\

Prompt: A photo of seven dogs.\\
Think: Prompt requests seven dogs, and dog is countable, so it is a counting prompt. Seven is 7 in number form. The object form seen in the prompt is dogs. The corresponding singular form is dog.\\
Response: True, [(``dog'', ``dogs'', 7)]\\

Prompt: A photo of seven dogs and two cats.\\
Think: Prompt requests seven dogs and two cats. In this case, both dog and cat are countable, so it is a counting prompt. Seven is 7 in number form, and two is 2. The object form seen in the prompt is dogs and cats. The corresponding singular forms are dog and cat.\\
Response: True, [(``dog'', ``dogs'', 7), (``cat'', ``cats'', 2)]\\

Prompt: A photo of some cups.\\
Think: Prompt requests some cups; ``some'' indicates a small number, so it is countable. Let's just choose a reasonable number between 1 and 5, feel free to randomize within this range. Let's choose 3 this time. The object form seen in the prompt is cups. The corresponding singular form is cup.\\
Response: True, [(``cup'', ``cups'', 3)]
\end{tcolorbox}

\newpage
\begin{tcolorbox}[colback=gray!5,colframe=gray!50,
                  title=Formatted transcript of selected LLM responses]
\small

\textbf{Prompt:} A realistic photo of a scene with nine pineapples.\\
\textbf{ChatGPT:} Think: The prompt specifies ``nine pineapples.'' Pineapple is a countable noun, so this is a counting prompt. The number ``nine'' corresponds to 9. The object form seen in the prompt is ``pineapples,'' and the singular form is ``pineapple.''\\

Response:\\
True, [(``pineapple'', ``pineapples'', 9)]\\

\textbf{Prompt:} A photo of some water.\\
\textbf{ChatGPT:} Think: The prompt specifies ``some water.'' Water is an uncountable noun, so it cannot be treated as a countable object. There is no countable object requested here.\\

Response:\\
False, []\\

\textbf{Prompt:} A few apples.\\
\textbf{ChatGPT:} Think: The prompt specifies ``a few apples.'' Apple is a countable noun, so this is a counting prompt. ``A few'' typically corresponds to 3. The object form seen in the prompt is ``apples,'' and the singular form is ``apple.''\\

Response:\\
True, [(``apple'', ``apples'', 3)]\\

\textbf{Prompt:} A magnificent landscape with lots of sunshine, animals everywhere, and plenty of greenery.\\
\textbf{ChatGPT:} Think: The prompt mentions ``animals,'' but it does not specify a number or quantity for any countable object. All quantities are vague (“lots,” “plenty”), so there is no explicit count requested.\\

Response:\\
False, []\\

\textbf{Prompt: } A bright outdoor farmers’ market captured in a natural documentary style: show 4 wooden tables arranged in two rows, each table displaying 11 neatly stacked pumpkins. Behind the tables stand 5 vendors, each with 3 small hanging lanterns attached to their stall. On the left, include 3 crates, each filled with 6 fresh apples, all clearly visible. In front of the crates, show 2 children crouched down examining 9 colorful marbles scattered on the ground. On the right side of the scene, place 2 bicycles, each with 2 woven baskets, and inside each basket include 4 lemons. Add 5 small fabric flags strung across the walkway above. Ensure all objects are visible, not overlapping, and easy to count. \textit{(prompt generated separately by ChatGPT).}\\
\textbf{ChatGPT:} True, [
(``table'', ``tables'', 4),
(``pumpkin'', ``pumpkins'', 11),
(``vendor'', ``vendors'', 5),
(``lantern'', ``lanterns'', 3),
(``crate'', ``crates'', 3),
(``apple'', ``apples'', 6),
(``child'', ``children'', 2),
(``marble'', ``marbles'', 9),
(``bicycle'', ``bicycles'', 2),
(``basket'', ``baskets'', 2),
(``lemon'', ``lemons'', 4),
(``flag'', ``flags'', 5)
]
\end{tcolorbox}

\newpage
\section{Extension to multi-step models}\label{app:multistepdm}

\begin{figure}[H]
\centering
    \includegraphics[width=0.6\textwidth]{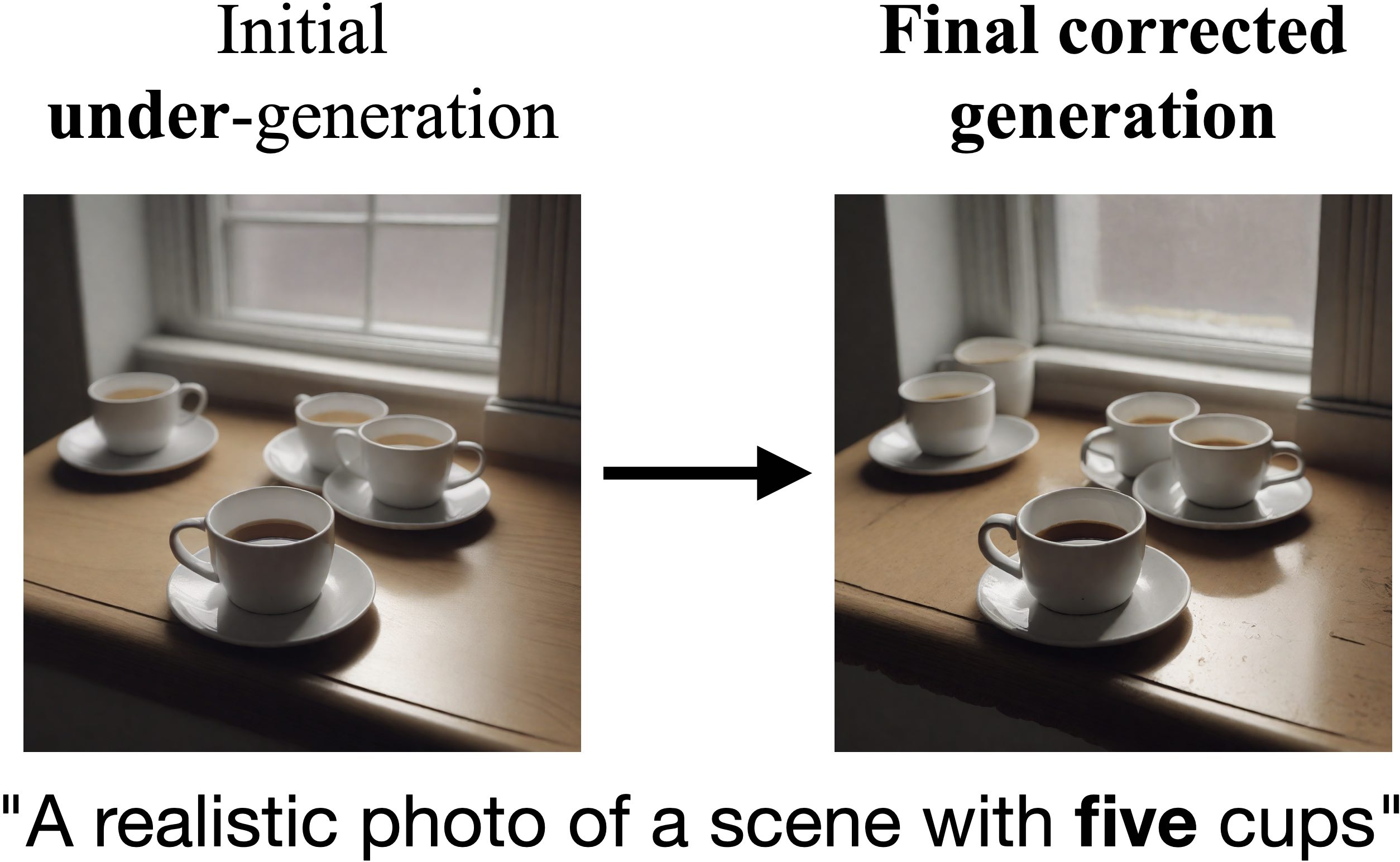}
    \caption{\textbf{Example of numeracy correction using DNO~\cite{tang2025dno} with our \emph{D2D} critic.} This took 12 optimization steps (approximately 7 minutes).}
    \label{fig:multi_apply}
    \centering
\end{figure}

We show that the \emph{D2D} critic can be extended to multi-step diffusion models as well. We demonstrate this by applying \emph{D2D} to the multi-step initial noise framework proposed in Direct Noise Optimization (DNO)~\cite{tang2025dno} (see Fig.~\ref{fig:multi_apply}). However, there is a significant tradeoff in latency, as convergence took approximately 7 minutes for the prompt ``A realistic photo of a scene with five cups.''

\section{Diversity analysis of generated images}\label{app:div_gen}

\begin{table}[H]
  \caption{\textbf{\emph{D2D} does not significantly impact diversity compared to corresponding base models.} This aligns with prior work~\cite{wang2024silent} which found that selecting higher-quality initial noises does not significantly impact diversity. We compute Vendi scores across images generated from different random seeds for the same prompt, and average the scores across prompts, using Inception-based embeddings. Standard deviations noted. On CoCoCount~\cite{binyamin2025mic} and D2D-Small, using base models SDXL-Turbo, SD-Turbo, and Pixart-DMD. Across four seeds.}
  \label{tab:div_metrics}
  \centering
  
\begin{tabular}{lc}
\toprule
\multicolumn{1}{c}{Method} & Vendi~\cite{friedman2023vendi} \\ \midrule
SDXL-Turbo~\cite{sauer2025turbo} & 1.76 $\pm$ 0.23\\ 
SDXL-Turbo~\cite{sauer2025turbo} + \emph{D2D} & 1.82 $\pm$ 0.31 \\\midrule
SD-Turbo~\cite{sauer2025turbo} & 1.87 $\pm$ 0.25\\
SD-Turbo~\cite{sauer2025turbo} + \emph{D2D} & 1.98 $\pm$ 0.33 \\\midrule
Pixart-DMD~\cite{chen2025pixartdmd} & 1.76 $\pm$ 0.26\\ 
Pixart-DMD~\cite{chen2025pixartdmd} + \emph{D2D} & 1.82 $\pm$ 0.32 \\
\bottomrule
\end{tabular}
\end{table}

\section{Broader impacts}\label{app:broader}
As a text-to-image pipeline, our method has many practical downstream uses, so it is necessary to exercise caution in deployment. Our method offers the advantage of stronger numeracy, which may be desirable in applications where users need to generate specific counts of objects. Our model may inherit biases of pre-trained diffusion base models and detectors. We suggest using strict NSFW filters and building more robust detectors.

\section{Use of Large Language Models (LLMs)}\label{app:llm}
We used LLMs to help with word choice for clarity and debugging.

\newpage
\section*{NeurIPS Paper Checklist}

\begin{enumerate}

\item {\bf Claims}
    \item[] Question: Do the main claims made in the abstract and introduction accurately reflect the paper's contributions and scope?
    \item[] Answer: \answerYes{} 
    \item[] Justification: Yes, the methods and experiments sections justify our claims, with additional details in the appendix.

\item {\bf Limitations}
    \item[] Question: Does the paper discuss the limitations of the work performed by the authors?
    \item[] Answer: \answerYes{} 
    \item[] Justification: Yes, we discuss limitations in our experiments section.

\item {\bf Theory assumptions and proofs}
    \item[] Question: For each theoretical result, does the paper provide the full set of assumptions and a complete (and correct) proof?
    \item[] Answer: \answerNA{} 
    \item[] Justification: Our work does not present new theoretical results.

    \item {\bf Experimental result reproducibility}
    \item[] Question: Does the paper fully disclose all the information needed to reproduce the main experimental results of the paper to the extent that it affects the main claims and/or conclusions of the paper (regardless of whether the code and data are provided or not)?
    \item[] Answer: \answerYes{} 
    \item[] Justification: The paper and appendix provide fine-grained implementation details. We will release our code and benchmarks.

\item {\bf Open access to data and code}
    \item[] Question: Does the paper provide open access to the data and code, with sufficient instructions to faithfully reproduce the main experimental results, as described in supplemental material?
    \item[] Answer: \answerYes{} 
    \item[] Justification: Yes, we will release our code.

\item {\bf Experimental setting/details}
    \item[] Question: Does the paper specify all the training and test details (e.g., data splits, hyperparameters, how they were chosen, type of optimizer) necessary to understand the results?
    \item[] Answer: \answerYes{} 
    \item[] Justification: The experimental setting is detailed in the experiments section, with more specifics included in the appendix.

\item {\bf Experiment statistical significance}
    \item[] Question: Does the paper report error bars suitably and correctly defined or other appropriate information about the statistical significance of the experiments?
    \item[] Answer: \answerYes{} 
    \item[] Justification: For different experiments, we report the number of seeds we averaged results over (often four seeds, with one seed for some ablations).

\item {\bf Experiments compute resources}
    \item[] Question: For each experiment, does the paper provide sufficient information on the computer resources (type of compute workers, memory, time of execution) needed to reproduce the experiments?
    \item[] Answer: \answerYes{} 
    \item[] Justification: We provide details on the compute setting (GPU type and hence, the corresponding GPU VRAM, and inference cost) in our experiments section.
    
\item {\bf Code of ethics}
    \item[] Question: Does the research conducted in the paper conform, in every respect, with the NeurIPS Code of Ethics \url{https://neurips.cc/public/EthicsGuidelines}?
    \item[] Answer: \answerYes{} 
    \item[] Justification:  Yes, our work conforms with the NeurIPS Code of Ethics.

\item {\bf Broader impacts}
    \item[] Question: Does the paper discuss both potential positive societal impacts and negative societal impacts of the work performed?
    \item[] Answer: \answerYes{} 
    \item[] Justification: Broader impacts are discussed in the appendix.
    
\item {\bf Safeguards}
    \item[] Question: Does the paper describe safeguards that have been put in place for responsible release of data or models that have a high risk for misuse (e.g., pre-trained language models, image generators, or scraped datasets)?
    \item[] Answer: \answerYes{} 
    \item[] Justification: We discuss safeguards in the appendix.

\item {\bf Licenses for existing assets}
    \item[] Question: Are the creators or original owners of assets (e.g., code, data, models), used in the paper, properly credited and are the license and terms of use explicitly mentioned and properly respected?
    \item[] Answer: \answerYes{} 
    \item[] Justification: The related works, models, and datasets are explicitly cited.

\item {\bf New assets}
    \item[] Question: Are new assets introduced in the paper well documented and is the documentation provided alongside the assets?
    \item[] Answer: \answerYes{} 
    \item[] Justification: Aside from our code, which we will release with appropriate licenses, there are no new assets.

\item {\bf Crowdsourcing and research with human subjects}
    \item[] Question: For crowdsourcing experiments and research with human subjects, does the paper include the full text of instructions given to participants and screenshots, if applicable, as well as details about compensation (if any)? 
    \item[] Answer: \answerYes{} 
    \item[] Justification: We conducted a questionnaire-based human evaluation of generated images. The participant instructions and additional details of the study protocol are provided in the appendix.

\item {\bf Institutional review board (IRB) approvals or equivalent for research with human subjects}
    \item[] Question: Does the paper describe potential risks incurred by study participants, whether such risks were disclosed to the subjects, and whether Institutional Review Board (IRB) approvals (or an equivalent approval/review based on the requirements of your country or institution) were obtained?
    \item[] Answer: \answerYes{} 
    \item[] Justification: The study consisted of a non-biomedical questionnaire evaluating generated images and was conducted in accordance with applicable local regulations and institutional ethics requirements.

\item {\bf Declaration of LLM usage}
    \item[] Question: Does the paper describe the usage of LLMs if it is an important, original, or non-standard component of the core methods in this research? Note that if the LLM is used only for writing, editing, or formatting purposes and does \emph{not} impact the core methodology, scientific rigor, or originality of the research, declaration is not required.
    \item[] Answer: \answerYes{} 
    \item[] Justification: Our core method does not involve LLMs.

\end{enumerate}

\end{document}